\documentclass[10pt,twocolumn,letterpaper]{article}

\usepackage{cvpr}              

\usepackage[dvipsnames]{xcolor}

\usepackage{amsmath}
\usepackage{amssymb}
\usepackage{bm}
\usepackage{multirow}
\usepackage[ruled,vlined]{algorithm2e}

\usepackage{booktabs}
\usepackage{xspace}
\usepackage{url}

\usepackage{graphicx}
\usepackage{lipsum}
\usepackage{microtype}
\usepackage{nicefrac}
\usepackage{verbatim}
\usepackage{bbding}
\usepackage{subcaption}
\usepackage{soul}
\usepackage[export]{adjustbox}

\definecolor{cvprblue}{rgb}{0.21,0.49,0.74}
\usepackage[pagebackref,breaklinks,colorlinks,citecolor=cvprblue]{hyperref}

\title{Optimisation-Based Multi-Modal Semantic Image Editing}

\author{Bowen Li$^{2 \dag}$ \quad Yongxin Yang$^{1}$ \quad Steven McDonagh$^{3\dag}$ \quad Shifeng Zhang$^{1}$ \\ 
Petru-Daniel Tudosiu$^{1}$ \quad Sarah Parisot$^{1}$\\
$^{1}$Huawei Noah's Ark Lab \quad $^{2}$ University of Oxford \quad $^{3}$ University of Edinburgh\\
}

\begin{document}
\maketitle
\let\thefootnote\relax\footnotetext{$\dag$ Work done in part at Huawei Noah's Ark Lab}

\begin{abstract}
Image editing affords increased control over the aesthetics and content of generated images. Pre-existing works focus predominantly on text-based instructions to achieve desired image modifications, which limit edit precision and accuracy. In this work, we propose an inference-time editing optimisation, designed to extend beyond textual edits to accommodate multiple editing instruction types (\eg spatial layout-based; pose, scribbles, edge maps). We propose to disentangle the editing task into two competing subtasks: successful local image modifications and global content consistency preservation, where subtasks are guided through two dedicated loss functions. By allowing to adjust the influence of each loss function, we build a flexible editing solution that can be adjusted to user preferences. We evaluate our method using text, pose and scribble edit conditions, and highlight our ability to achieve complex edits, through both qualitative and quantitative experiments. 
\end{abstract} 

\section{Introduction}
Large scale text-to-image generative diffusion models have revolutionised image generation capabilities. Recent foundation models such as DALL-E~\citep{ramesh2022hierarchical}, Stable Diffusion~\citep{rombach2022high} and Imagen~\citep{saharia2022photorealistic} have achieved impressive results in terms of composition and image quality.  
While these models successfully leverage the power of text conditioning, precise controllability of image outputs remain limited and achieving desired outcomes often requires time-consuming and cumbersome prompt engineering~\citep{sonkar2022visual}. A natural way of increasing control over image structure and simplifying this process is to introduce additional conditioning in the form of image layout descriptions. Recent works have introduced additional trainable modules that, when combined with a pre-trained foundation model, allow to condition generated images according to specific layout instructions. Such conditioning has been realised through object bounding boxes~\citep{li2023gligen}, poses, edge maps, scribbles and segmentation maps~\citep{zhang2023adding, mou2023t2i}. 
The increased controllability provided by these modules have further exacerbated a key missing component: editability of generated images. Small modifications to layout or text instructions can lead to large undesirable image modifications, in particular for looser constraints (\eg pose). As an alternative to iterative prompt updates, with no guarantees of maintaining a consistent image appearance, editing can afford local image modifications whilst also consistently preserving desired aesthetics. 

In light of these potential benefits, text-driven image editing has been heavily explored recently.  
Editing solutions typically leverage a pre-trained foundational model and introduce a novel editing mechanism. Methods can be separated in three categories: 1) training-based methods that fine-tune the base models \citep{brooks2023instructpix2pix,zhang2023sine,kawar2023imagic}, 2) training-free approaches that keep the generative model frozen~\citep{hertz2022prompt,couairon2022diffedit, tumanyan2023plug} and 3) inference-time optimisation~\citep{ma2023directed, parmar2023zero,epstein2023diffusion,dong2023prompt}. Training-based approaches achieve higher quality edits, at the cost of expensive dataset construction, training and reduced generalisability. Training-free editing typically relies on attention manipulation or inversion mechanisms. These methods are faster and more practical, but also parameter sensitive and more restricted in editing ability. Alternatively, one can leverage ideas from both lines of research by considering inference-time optimisation, \ie optimise intermediate image features directly instead of model weights. This strategy increases control over the edit process, while maintaining the flexibility of training free methods. Inference-time optimisation has been leveraged to guide cross attention mechanisms~\citep{ma2023directed, parmar2023zero}, move and modify instances through features and attention updates~\citep{epstein2023diffusion} and automatically tune input prompts~\citep{dong2023prompt}; increasing control over the structure of generated images. 

Nonetheless, the aforementioned methods mainly rely on text-driven instructions to update image content. As discussed above, this can limit the precision and accuracy of required edits, as well as introduce high sensitivity and brittleness to the considered text instructions.
In this work, we propose a novel editing method that relies on inference-time optimisation of latent image features that is intrinsically designed to work with a variety of conditioning signals, including text, pose or scribble instructions. We disentangle the editing task into preservation and modification tasks via two optimisation losses. The former is realised by imposing consistency between latent features of the input and edited image, while the latter leverages an intermediate output (a \emph{guidance image}) with relaxed preservation constraints to guide the editing process in target modification areas. Our method can make use of layout control modules, effectively allowing editing using a wide range of editing queries. By disentangling preservation and modification tasks, we provide a flexible solution where the influence of each subtask can be adjusted to user preferences.
We evaluate our method using text, pose and scribble conditions.  
Further, we highlight an ability to achieve complex edits, derived from multiple concurrent conditions, and evaluate resulting performance using both qualitative and quantitative experiments. 

\noindent To summarise, our main contributions are as follows: 
\begin{itemize}
    \item A novel image editing method for frozen diffusion models that goes beyond text editing, additionally capable of handling image layout conditions.
    \item A disentangled inference-time optimisation-based editing strategy that separates background preservation from foreground editing, affording flexibility to increase focus on one subtask and increased robustness to poor quality edit masks.
    \item We adapt state of the art editing method Diffedit to leverage layout conditions, and provide empirical evidence that a special configuration of our approach yields equivalent results with increased robustness to edit mask quality. 
\end{itemize}

\label{sec:intro}

\section{Related Work}\label{sec:rel}
We review closely related image editing work, with a focus on text-guided editing, under text-to-image diffusion models. We identify three main methodologies: fine-tuning based, training-free and inference-time optimisation methods.

\paragraph{Fine-tuning based image editing} 
enables a pre-trained diffusion model to edit images by fine-tuning model weights.
InstructPix2Pix~\citep{brooks2023instructpix2pix} leverages a large language model to build an extensive training dataset comprising image, edit-instruction pairs that are used to fine-tune a diffusion model to achieve editing based on written instructions. 
Imagic~\citep{kawar2023imagic} fine-tunes the textual-embedding and the generative model to ensure a capable reconstruction ability on each considered image, then interpolates between the original and editing prompt in order to obtain the edited image. Lastly, SINE~\citep{zhang2023sine} fine-tunes a pre-trained models on each image edit target, then combines predictions from the pre-trained and fine-tuned models, conditioned on the edit and original image prompt, respectively. 
These techniques generally require building a large training dataset, or per image fine-tuning at inference time. Both constraints can be costly, and notably make extensions to additional types of edit conditions (beyond text) more complex and expensive. 

\paragraph{Training-free image editing} achieves image modification using a frozen pre-trained model. These approaches typically involve manipulation of the denoising diffusion process to introduce editing functionality. 
Prompt-to-prompt~\citep{hertz2022prompt} enables image editing by manipulating cross-attention maps between text and image features and exploiting their differences when considering the original and edit prompts. Similarly, Plug-and-play~\citep{tumanyan2023plug} leverages attention mechanisms, injecting self attention features from the original image into the generation process of the edited image. \citet{ravi2023preditor} propose an editing method, built on top of the \mbox{DALL-E 2} model~\citep{ramesh2022hierarchical}, leveraging CLIP~\citep{radford2021learning}
embeddings. 

An alternative popular strategy involves inverting the input image to a noise vector, and using this noise vector as input (instead of random noise) to the edit-conditioned generation process~\citep{meng2021sdedit}. DiffEdit~\citep{couairon2022diffedit} extends the method with a deterministic inversion strategy and automated edit mask estimation.  EDICT~\citep{wallace2023edict} further proposes an improved inversion strategy inspired from affine coupling layers. While benefiting from higher flexibility and speed, these methods typically have reduced ability to perform large or complex modifications due to their entirely training-free nature. 

\begin{figure*}[t]
    \centering
    \includegraphics[trim={0 3.4cm 0 3.6cm},clip,width=0.9\linewidth]{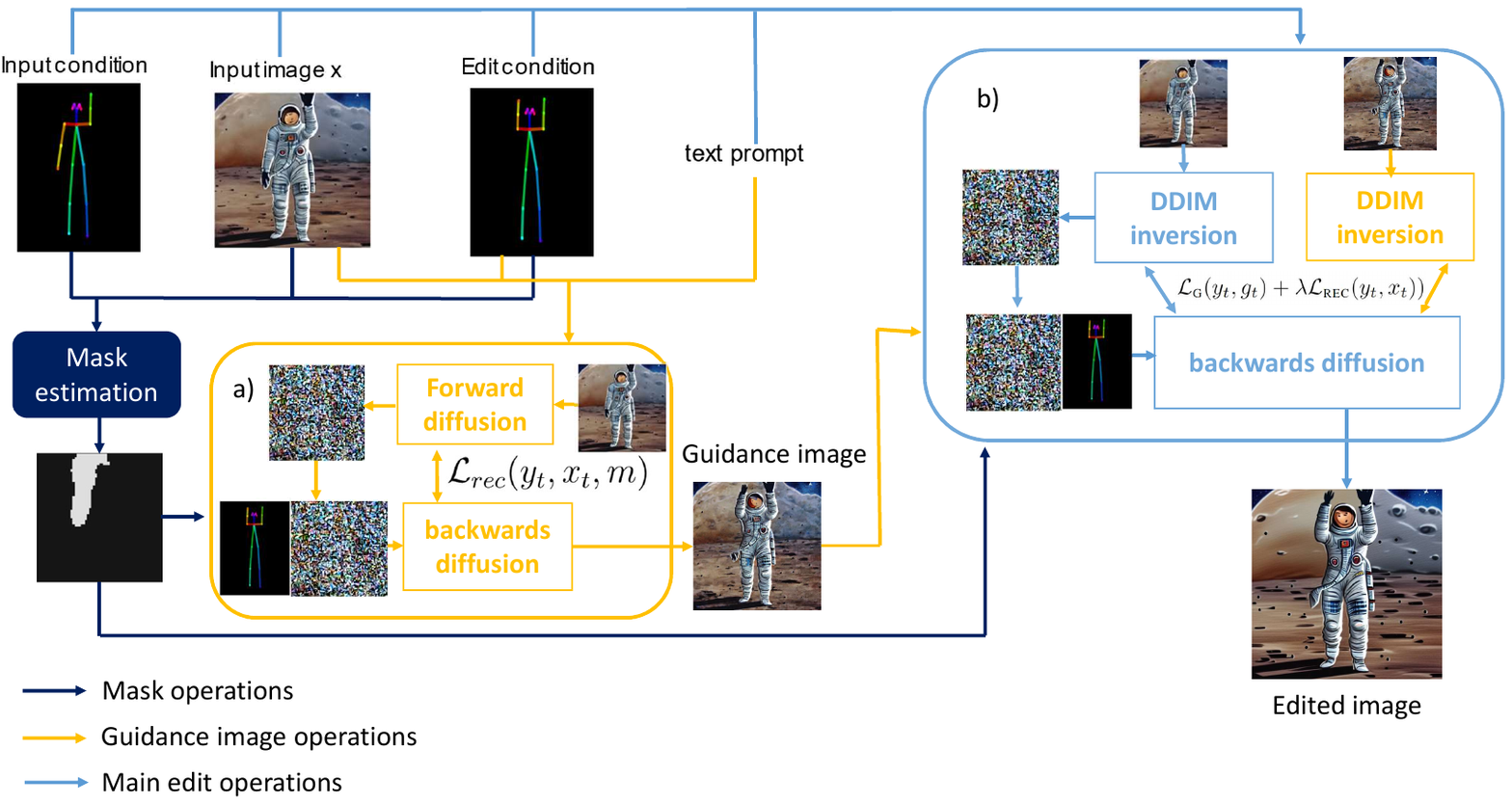}
    \caption{Multi-step overview of our proposed framework, a) guidance image generation block using preservation loss only, b) final image generation block using both preservation and guidance losses.}
    \label{fig:archi}
\end{figure*}

\paragraph{Inference-time optimisation (ITO)} is an intermediate approach that leverages frozen pre-trained models, but introduces an optimisation component at inference-time, typically to update image features. 
Initial works leveraged CLIP text-image similarity losses to guide the editing process using GANs~\citep{patashnik2021styleclip} or diffusion models~\citep{avrahami2022blended, kim2021diffusionclip}. 
\citet{mokady2023null} introduce an extension to prompt-to-prompt for natural image editing using a pivotal inversion technique and null-text optimisation. 
\citet{parmar2023zero} aim to preserve content, unrelated to the edit-text, 
by imposing consistency between the text-image cross-attention maps before and after edit. The method requires computing a textual embedding direction in advance using a thousand sentences, which is expensive and makes extension to additional edit conditions challenging, and is limited by the resolution of the cross-attention map for structure guidance.

These pre-existing approaches have focused on text-driven editing methods, requiring complex strategies to achieve precise local modifications, notably by leveraging attention mechanisms.
\citet{epstein2023diffusion} introduce an optimisation procedure to guide the positioning, size and appearance of objects in an image using precise edit instructions (\eg object desired pixel location) and dedicated loss functions. These loss functions drive the update of intermediate image features and attention maps. This approach requires expensive manipulation of attention layers, and explicit estimation of local positioning of objects at all levels of the diffusion UNet. It is heavily dependent on the accuracy of internal attention maps, leading to leakage when instances are not strongly salient. 
 By developing a method capable of leveraging explicit layout constraints, we are able to provide precise edit instructions with much simpler optimisation constraints and remove dependency on internal attention map representations. Furthermore, our disentangling strategy between preservation and modification affords additional controllability and flexibility, allowing to adjust the influence of each subtask easily.

\section{Method}\label{sec:method}
Given an input image $I$ and a semantic edit condition $C$, our objective is to locally modify $I$ according to $C$ while accurately preserving image content not directly related to the desired edit. We assume access to a text-based image caption $S$, hand-crafted or generated by an image captioning model, and a pre-trained text-to-image diffusion model capable of leveraging image-space layout conditions (\eg pose or scribble). The latter model can be instantiated using \eg ControlNet modules~\citep{zhang2023adding}. In direct contrast to pre-existing work, that restrict edit condition inputs to text form, we alternatively assume $C$ can be provided as either text or image-space layouts. 

An overview of our method is provided in Fig.~\ref{fig:archi}. We propose a semantic image editing method by leveraging the concept of inference time optimisation. Our strategy is uniquely designed to accommodate multiple types of edit instruction including text, pose, scribble and more complex image layouts. We build on the concept of image-to-image editing where an input image is progressively converted into a noise vector, then denoised across multiple timesteps using predictions from a pre-trained latent diffusion model, according to edit condition $C$. 
We introduce an optimisation procedure over latent image features within this encoding-decoding process. We propose to decompose the edit task into two competing subtasks, using distinct content preservation and local image modification losses. 
Our preservation loss imposes consistency between the input and edited images, thus driving the preservation of original details. The image modification loss
provides appearance guidance for the expected image edit area, using an intermediate edited image obtained with reduced content preservation constraints. We additionally enhance our preservation loss with a binary mask that defines a local edit area, thus enabling both content preservation and locally accurate editing. 

We provide an overview of diffusion models and image-to-image editing~\cite{meng2021sdedit} in~\cref{sec:method:prelim}, followed by description of our core method components: preservation loss, guidance loss, mask, and image guidance generation in~\cref{sec:method:optim_edit}.  

\subsection{Preliminaries}
\label{sec:method:prelim}

Denoising diffusion models learn to invert a multi-step diffusion process, where an input image $x_0$ is progressively converted into a Gaussian noise vector $x_T$, by iteratively adding noise $\epsilon \sim \mathcal{N}(0,I)$ across multiple timesteps, according to a specific variance schedule. Given $x_T$ and a text prompt, a model $\epsilon_{\theta}$ (\eg a UNet~\citep{ronneberger2015u}) is trained to reconstruct the input image by predicting the noise vector to remove at a given timestep $t$. At inference time, $\epsilon_{\theta}$ iteratively estimates noise vectors across $T$ timesteps to generate a new image from random noise. Using the DDIM deterministic sampling procedure~\citep{song2020denoising}, we can estimate the updated intermediate denoised image at timestep $t{-1}$ as:
\begin{equation}
\small
\label{eq:backwards}
    x_{t-1} = \sqrt{\frac{\alpha_{t-1}}{\alpha_t}}x_t +  \sqrt{\alpha_{t-1}} \left( \sqrt{\frac{1}{\alpha_{t-1}}-1} - \sqrt{\frac{1}{\alpha_{t}}-1} \right) \epsilon_{\theta}(x_t,t,\mathcal{C}),
\end{equation}
where $\mathcal{C}$ is the embedding of the text prompt condition, and $\alpha_t$ is a measure of the noise level, dependent on the variance scheduler.

The forward-backward diffusion procedure can be viewed as an image encoding-decoding process. The SDEdit algorithm~\citep{meng2021sdedit}, commonly referred to as image-to-image, leverages this concept for the purpose of image editing. The key idea is to stop the forward diffusion process at an intermediate timestep $t_E < T$, and reconstruct the image, from $t_E$, using a \emph{different} text condition. This strategy typically results in preservation of some of the original image attributes yet often leads to unwanted global modifications~\citep{couairon2022diffedit}. 
Leveraging the trained diffusion model $\epsilon_{\theta}$, in conjunction with a deterministic reverse-DDIM process that encodes the input image~\citep{couairon2022diffedit}, can substantially improve image fidelity, yet sometimes at the cost of edit flexibility. Concretely, we carry out the same process as the backwards diffusion (decoding process), in the $x_0 \rightarrow x_{t_E}$ direction, using the following inverse update: 
\begin{equation}
\small
\label{eq:forward}
    x_{t+1} = \sqrt{\frac{\alpha_{t+1}}{\alpha_t}}x_t +  \sqrt{\alpha_{t+1}}\left( \sqrt{\frac{1}{\alpha_{t+1}}-1} - \sqrt{\frac{1}{\alpha_{t}}-1} \right) \epsilon_{\theta}(x_t,t).
\end{equation}
This process is commonly referred to as DDIM inversion; a naive way to invert a natural image (\ie estimate the noise vector that will generate this exact image).

We highlight that in this work, we consider diffusion models that operate in a latent space~\citep{rombach2022high}, where \mbox{$x_0 = f_{\phi} (I)$} is a latent representation of input image $I$, obtained using a pre-trained Variational AutoEncoder (VAE) model. 
\subsection{Optimisation-Based Image Editing}
\label{sec:method:optim_edit}

\noindent\textbf{Inference-Time Optimisation: Preservation Loss.}
Our method builds on the process, outlined in~\cref{sec:method:prelim}, to perform image editing. We encode our input latent image $x_0$ with DDIM inversion, up to an intermediate encoding level $t_E$, obtaining intermediate noised image latent $x_{t_E}$. Using this noised latent as input, we decode our image by firstly defining an edit condition in order to obtain latent edited image $y_0$. This edit condition can be a modified text prompt, or a modified layout input (\eg modified pose, scribble or edge map). We note that multiple edits can be considered \emph{conjointly} (\eg modified pose \emph{and} text prompt), without explicit methodology modification. 

To enforce appearance and structure consistency between the input and edited images, we introduce our latent feature optimisation process. At timestep $t$ of the backwards diffusion process, we update  the latent image features $y_t$ to increase similarity with $x_t$. To this end, we introduce a reconstruction task that pushes $y_t$ towards the direction of $x_t$. Formally, we update $y_t$ as:
\begin{equation}
\label{eq:term:recon}
    y_t' = y_t - \gamma \nabla_{y_t} \|y_t - x_t\|_2^2,
\end{equation}
where $\gamma$ is the learning rate. This update is repeated for $k$ gradient updates, pushing $y_t$ towards $x_t$ whilst limiting the number of gradient updates to avoid the trivial solution $y_t'=x_t$. 
We then proceed with the standard diffusion process, using $y_t'$ as input to the diffusion model at the next timestep.
Providing a binary mask $m$ to identify image regions that our edit condition will impact explicitly provides appropriate preservation of related original image content. Integration of mask $m$ into our reconstruction loss also leads to implicitly unconstrained flexibility in image regions that \emph{are} valid edit regions:
\begin{equation}
\label{eq:loss:recon}
   \mathcal{L}_{\textsc{rec}}(y_t, x_t, m)=\|m{\odot}y_t - m{\odot}x_t\|_2^2,
\end{equation}
\ie the reconstruction loss is only computed within the masked area. State of the art approaches typically invoke $m$, after computing $y_{t-1}$ using Eq.~\ref{eq:backwards}, by updating latents as \mbox{$y_{t-1}^m = m{\odot}y_{t-1} + (1-m){\odot}x_{t-1}$}. We find that by introducing the mask constraint in an optimisation framework, our strategy is more robust to mask quality (\eg underestimation of the edit area) and reduces the risks of introducing artefacts, whilst preserving non-edit regions. 

\noindent\textbf{Inference-Time Optimisation: Guidance Loss.}
While our preservation loss (\cref{eq:loss:recon}) focuses on faithfully maintaining image content, our complementary guidance loss seeks to improve controllability, enable precise local modification, and steer edited areas towards a desired appearance. We assume availability of an image $G$, with latent representation $g_0$, that accurately depicts the desired appearance of the target edit area. Our solution to obtain $g_0$ will be discussed later in this section. 
We first encode $g_0$ to obtain $g_{t_E}$, in similar fashion to the encoding process of $x_0$. At timestep $t$, we update $y_t$ towards increasing the cosine similarity with the guidance image features: 
\begin{equation}
    y_t' = y_t - \gamma \nabla_{y_t} \mathcal{L}_{\textsc{g}}(y_t, g_t) \qquad \text{where} \qquad \mathcal{L}_{\textsc{g}} = 1 - \frac{y_t\cdot g_t}{\| y_t\| \| g_t\|}.
\end{equation}
As larger differences are expected between $y$ and $g$ (\cf between $y$ and $x$), we make use of a more conservative cosine similarity loss for this task, in comparison with the previous MSE reconstruction loss. 
By combining guidance and preservation losses, our complete disentangled ITO process can update intermediate image features as:  
\begin{equation}
    y_t' = y_t - \gamma \nabla_{y_t} ((1-\lambda) \mathcal{L}_{\textsc{g}}(y_t, g_t) + \lambda \mathcal{L}_{\textsc{rec}}(y_t, x_t)),
\end{equation}
where $\lambda$ is a hyperparameter balancing the influence of the editing subtasks. As such, adjusting $\lambda$ allows the user to balance between content preservation and edit instruction accuracy. 
We carry out feature updates for the first $t_u$ steps of the backwards diffusion process.

\noindent\textbf{Guidance Image Generation.} The guidance image plays the crucial role of providing an accurate depiction of the expected appearance of the edited region. This is information that we do not have available beforehand, and need to generate according to an intermediate step. We rely on the observation that encoding our input image using random noise (as in~\citep{meng2021sdedit}) facilitates image modifications at the expense of reduced content preservation; while using DDIM inversion (as in \cite{couairon2022diffedit}) only affords more conservative changes. In contrast with our final edited output,  
which has to accurately preserve image background, our focus here is on accurately editing the regions of interest.  
Based on this observation, we propose to generate our guidance image using our ITO process with $\lambda=1$ (preservation only), with one key difference: here, our input image is encoded using random noise, instead of DDIM inversion to afford larger modifications of the input image.

\noindent\textbf{Edit Mask Generation.}
We estimate the binary edit mask by leveraging the mask generation approach proposed in~\citet{couairon2022diffedit}. The idea proposes to measure the differences between noise estimates using (1) the original image text prompt and (2) the edit prompt as conditioning, effectively inferring which image regions are affected most by the differing conditions. In contrast with~\citet{couairon2022diffedit}, which only considers text conditions, we additionally estimate masks from layout conditions. After encoding our input $x_0$ using random noise with seed $s$, we obtain $x_E(s)$ and carry out the backward diffusion \mbox{$x_E(s) \rightarrow y_0(\mathcal{C},s)$} for two different conditions: the original condition $\mathcal{C}_{\textsc{o}}$ (\eg original prompt or pose), and the edit condition $\mathcal{C}_{\textsc{edit}}$. In settings where only a layout condition is used, both images share the same text prompt and the layout input is integrated in the diffusion process using conditioning modules such as ControlNet~\citep{zhang2023adding} or T2IAdapter~\citep{mou2023t2i}. 
We estimate our mask by comparing noise estimates at the last timestep:
\begin{equation}
    m(x_0,\mathcal{C}_{\textsc{edit}}) = \frac{1}{n}\sum_{i \in [1:n]} \vert \epsilon_{\theta}(x_1, 1, \mathcal{C}_{\textsc{edit}},s_i) - \epsilon_{\theta}(x_1, 1, \mathcal{C}_{\textsc{o}}, s_i) \vert , 
\end{equation}
where $m$ is then converted to a binary mask using threshold $\tau$. The mask is averaged over $n$ runs with different seeds to increase the stability and accuracy of noise estimates.

\noindent\textbf{Task Difficulty Driven Modular Process.}
Disentangling preservation from modification provides a modular edit procedure, where a user can adapt constraints according to preferences and task difficulty; by adjusting loss-weighting parameter $\lambda$. 
Our complete edit procedure (\cref{fig:archi}) comprises three steps: mask estimation, guidance image generation, and image editing. The role of the guidance image is to ensure layout constraints are respected. This is typically useful for difficult pose modifications, where the ControlNet conditioned image-to-image setting fails to achieve the desired modification. For simple edit tasks (\eg small modifications such as a $90$ degrees arm movement, uniform backgrounds), it is possible to set $\lambda=1$, effectively focusing solely on the preservation task. This skips the guidance image generation step (depicted in yellow, \cref{fig:archi}), simplifying the edit process. We demonstrate empirically how this setting yields equivalent results to DiffEdit~\citep{couairon2022diffedit}, with reduced mask artefacts.

\section{Experiments}\label{sec:exp}

In this section, we describe our experimental setup, followed by our experiments under various conditions.

\begin{figure*}[t]
    \centering
    \begin{adjustbox}{minipage=\linewidth,scale=0.8}
    \begin{subfigure}[t]{.169\linewidth}
    \centering\includegraphics[width=\linewidth]{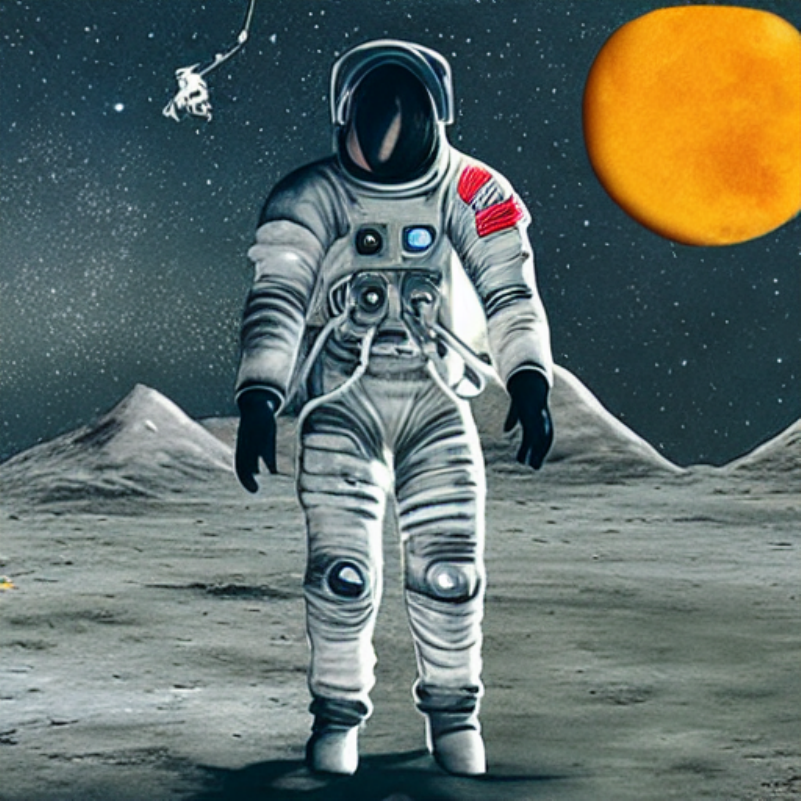}
    \end{subfigure}
      \begin{subfigure}[t]{.123\linewidth}
    \centering\includegraphics[width=\linewidth]{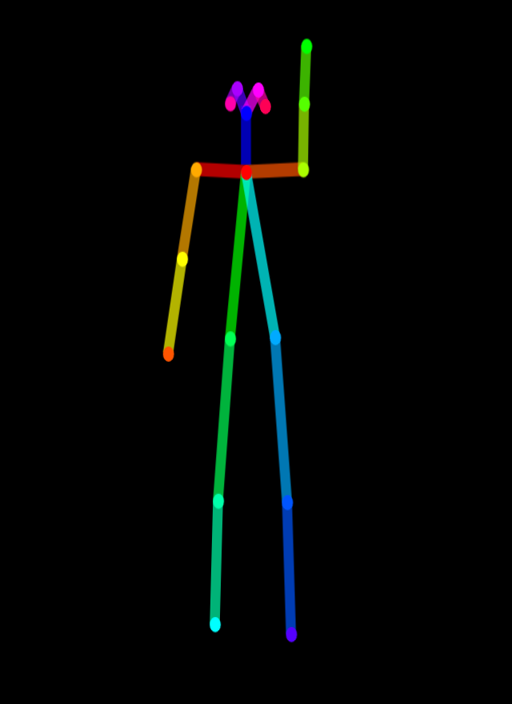}
    \end{subfigure}
  \begin{subfigure}[t]{.169\linewidth}
    \centering\includegraphics[width=\linewidth]{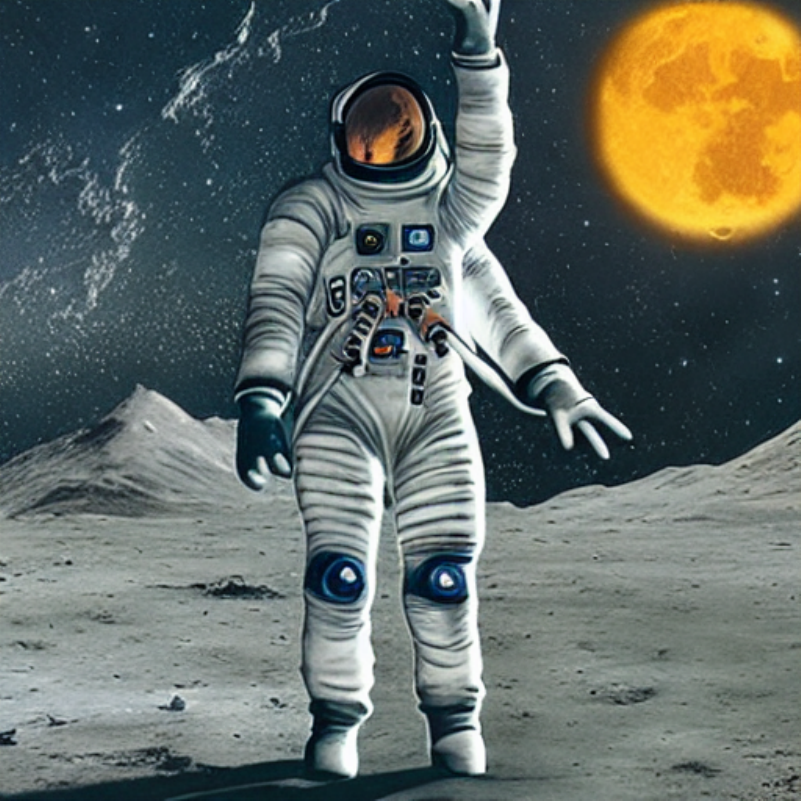}
  \end{subfigure}
  \begin{subfigure}[t]{.169\linewidth}
    \centering\includegraphics[width=\linewidth]{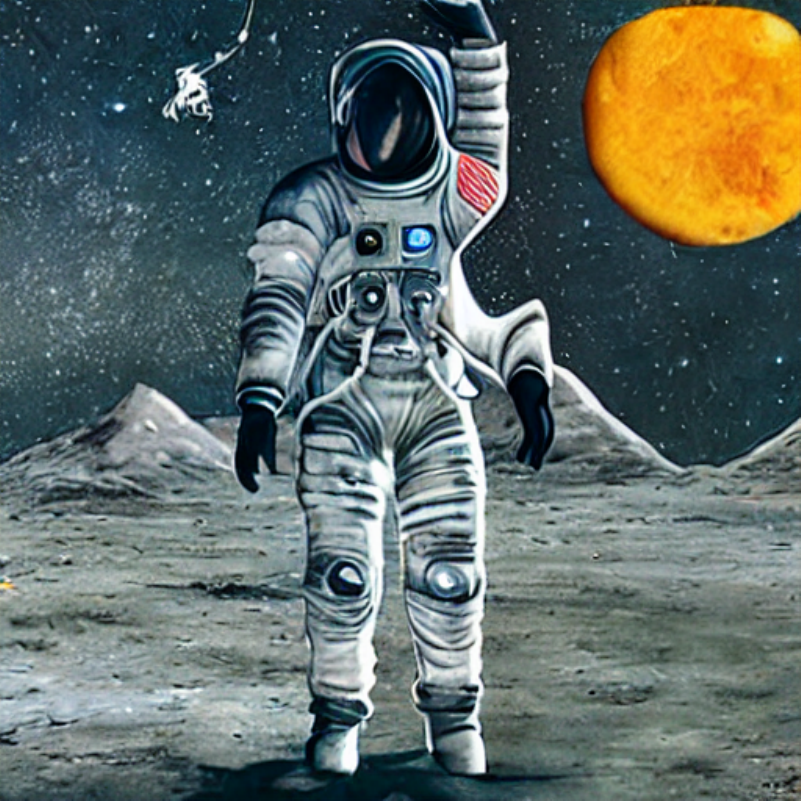}
  \end{subfigure}
  \begin{subfigure}[t]{.169\linewidth}
    \centering\includegraphics[width=\linewidth]{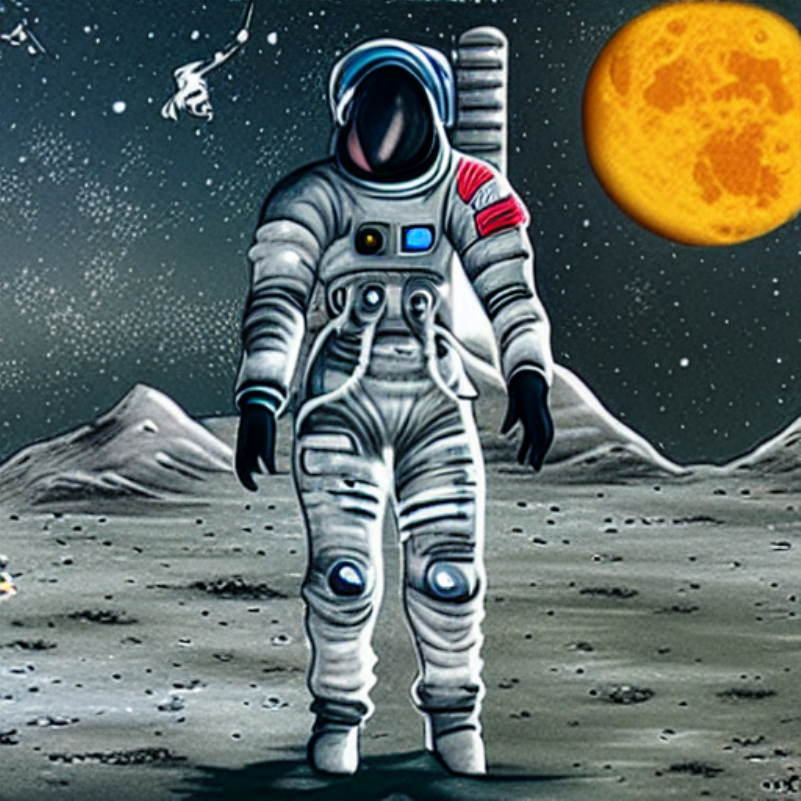}
  \end{subfigure}
  \begin{subfigure}[t]{.169\linewidth}
    \centering\includegraphics[width=\linewidth]{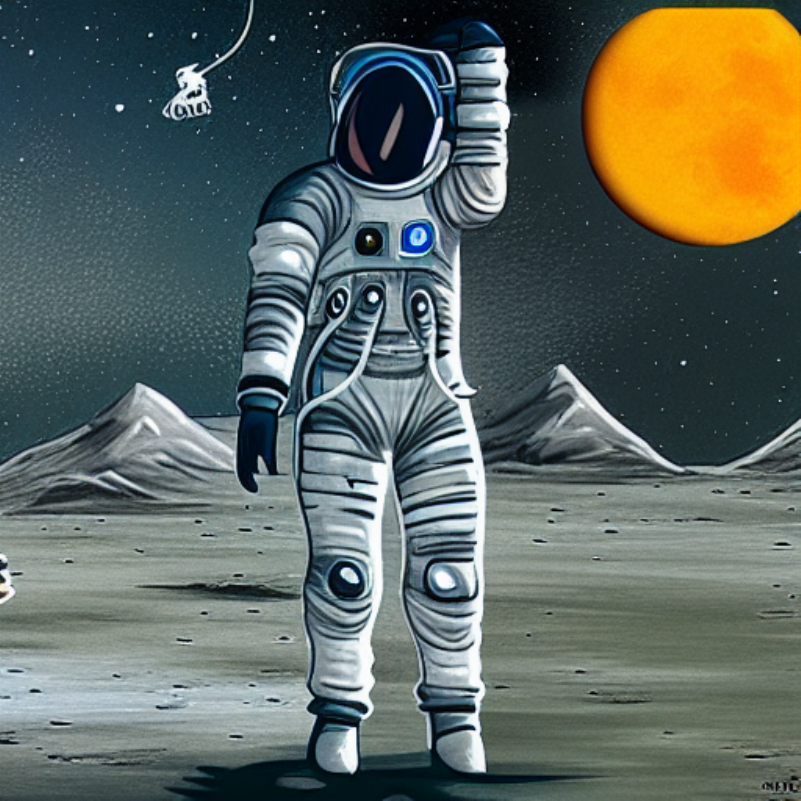}
  \end{subfigure}
  

\centering
    \begin{subfigure}[t]{.169\linewidth}
    \centering\includegraphics[width=\linewidth]{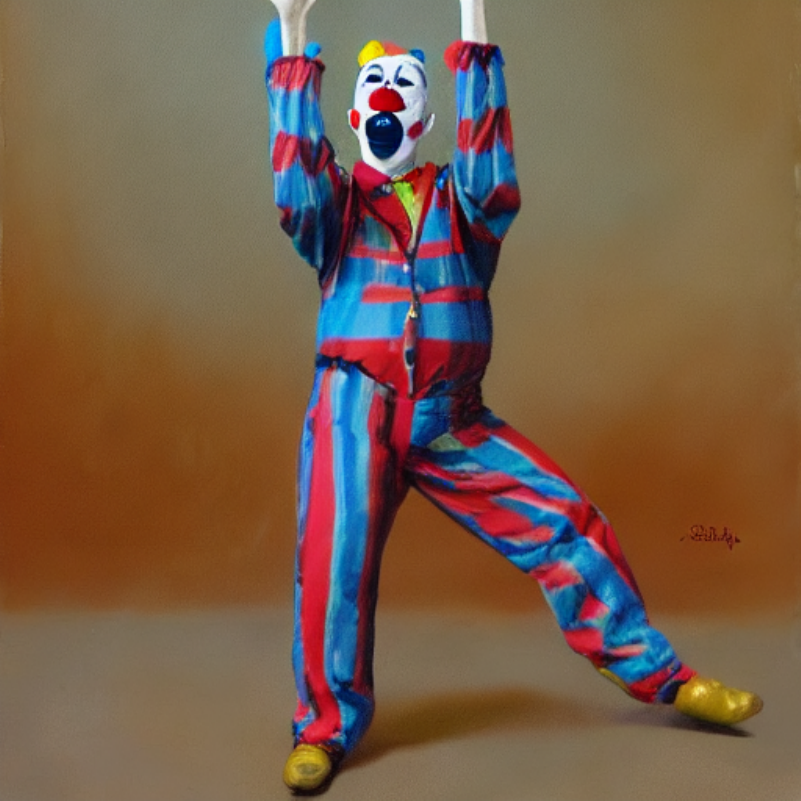}
    \end{subfigure}
      \begin{subfigure}[t]{.123\linewidth}
    \centering\includegraphics[width=\linewidth]{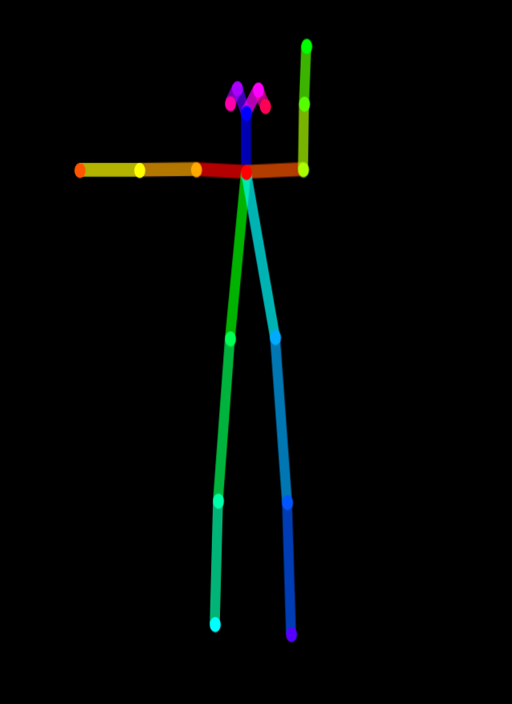}
    \end{subfigure}
  \begin{subfigure}[t]{.169\linewidth}
    \centering\includegraphics[width=\linewidth]{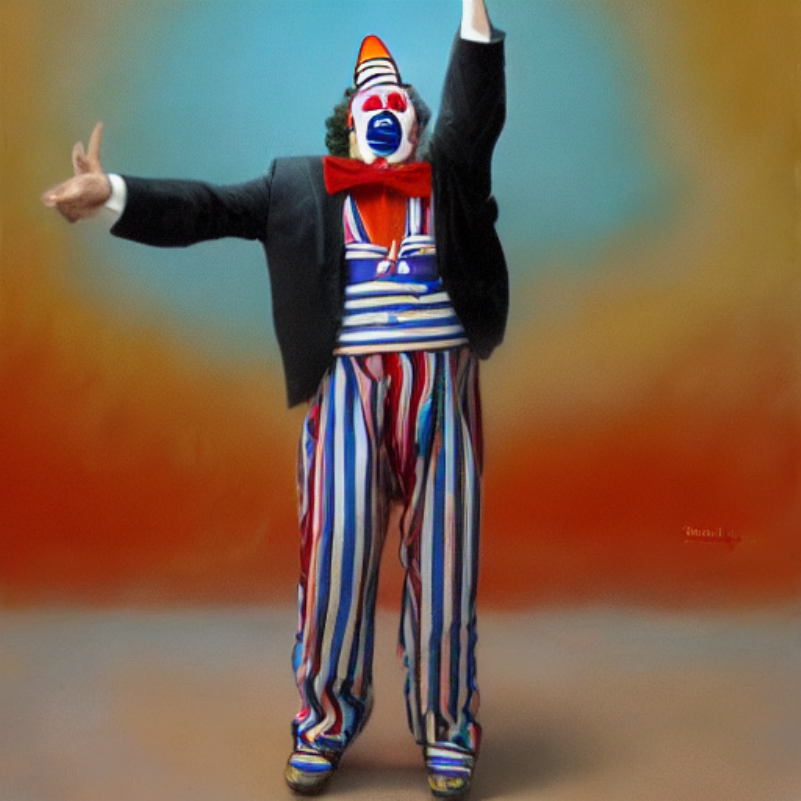}
  \end{subfigure}
  \begin{subfigure}[t]{.169\linewidth}
    \centering\includegraphics[width=\linewidth]{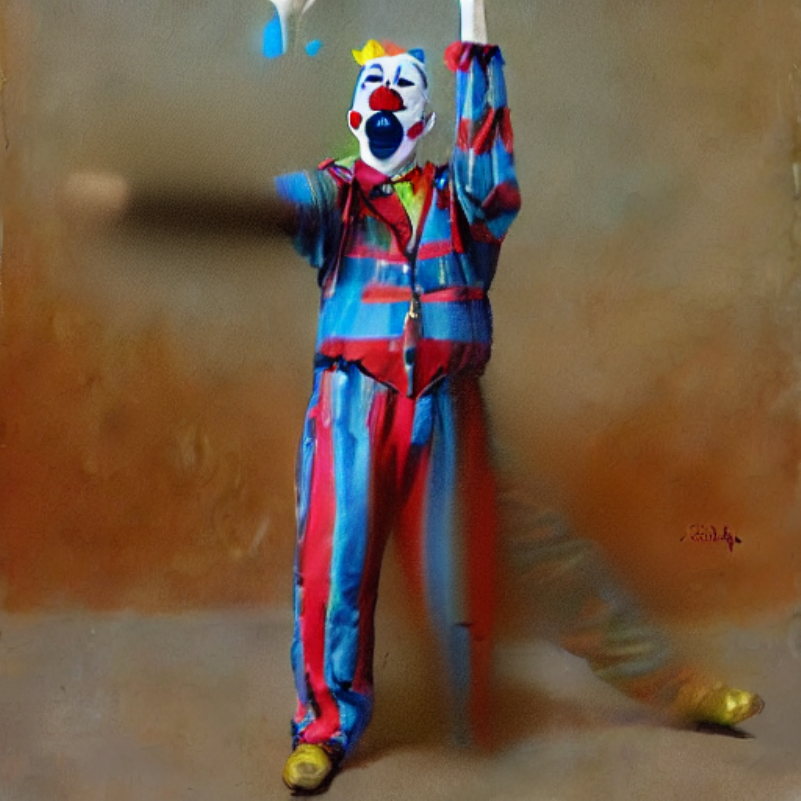}
  \end{subfigure}
  \begin{subfigure}[t]{.169\linewidth}
    \centering\includegraphics[width=\linewidth]{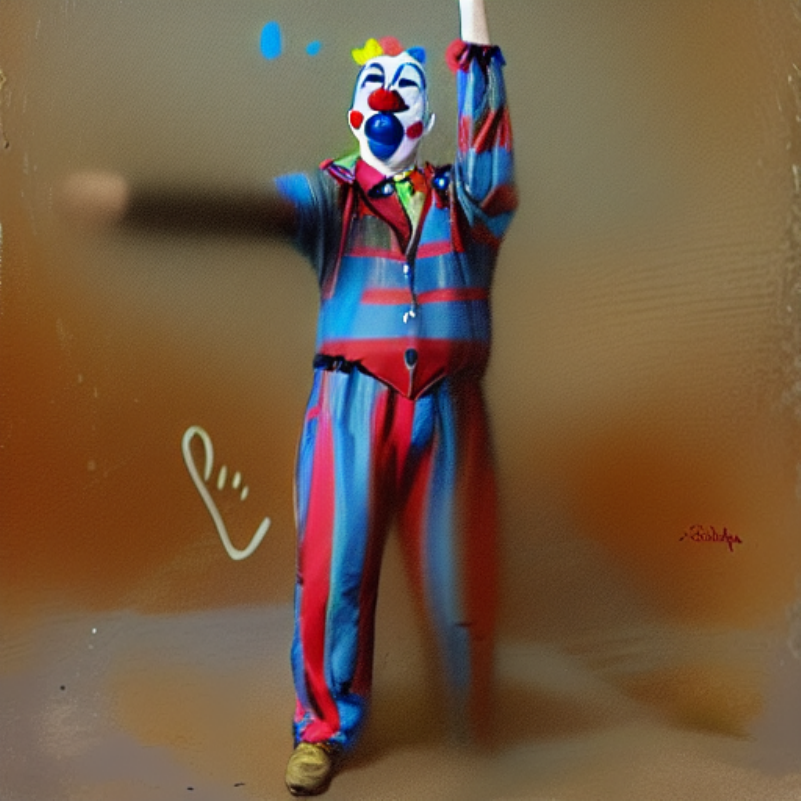}
  \end{subfigure}
  \begin{subfigure}[t]{.169\linewidth}
    \centering\includegraphics[width=\linewidth]{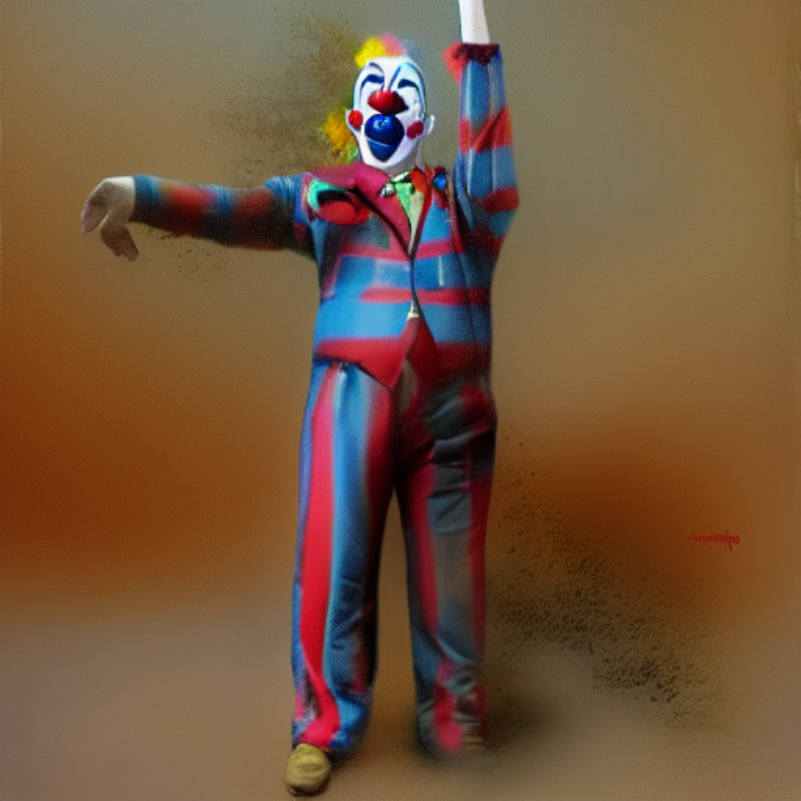}
  \end{subfigure}


\centering
    \begin{subfigure}[t]{.169\linewidth}
    \centering\includegraphics[width=\linewidth]{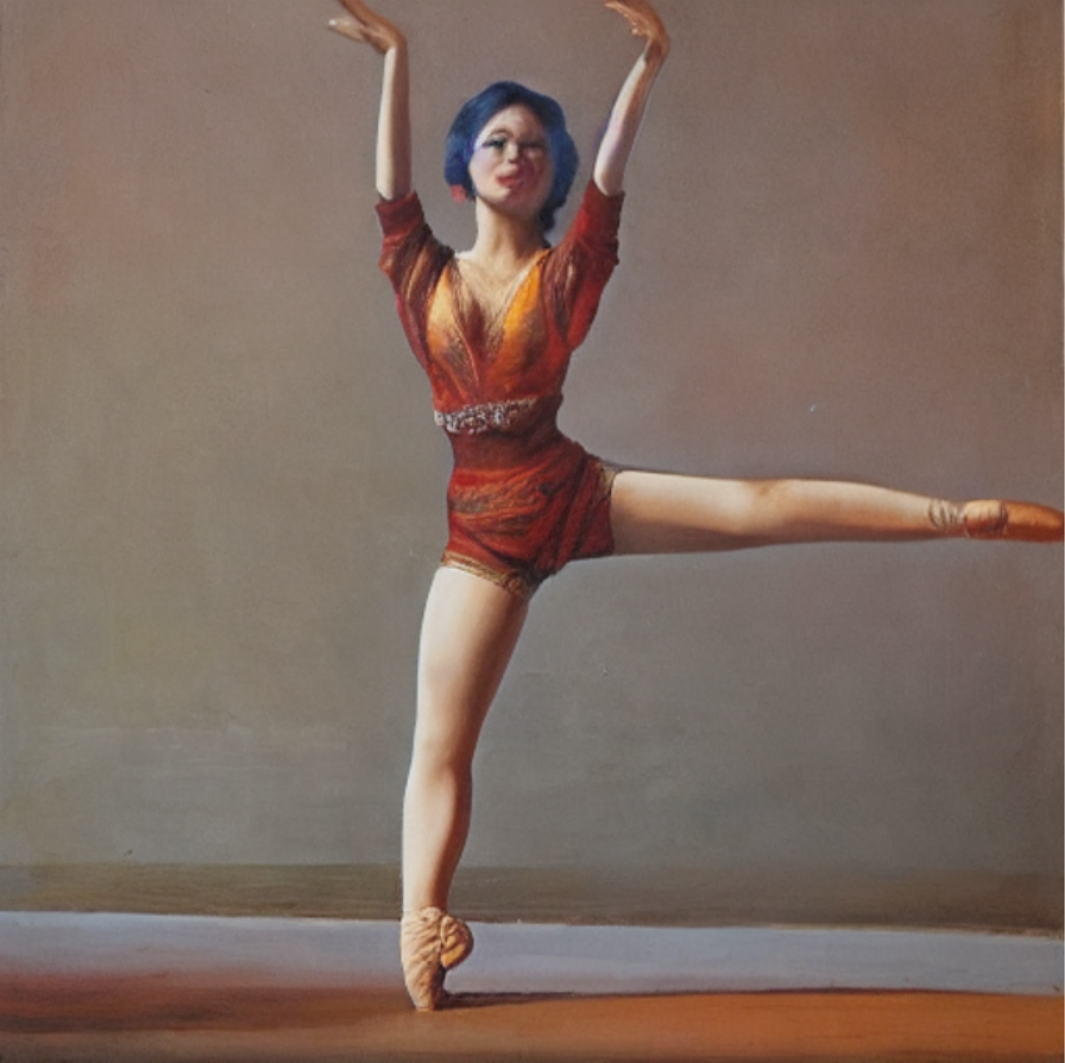}
    \end{subfigure}
      \begin{subfigure}[t]{.123\linewidth}
    \centering\includegraphics[width=\linewidth]{imgs/results/pose2.png}
    \end{subfigure}
  \begin{subfigure}[t]{.169\linewidth}
    \centering\includegraphics[width=\linewidth]{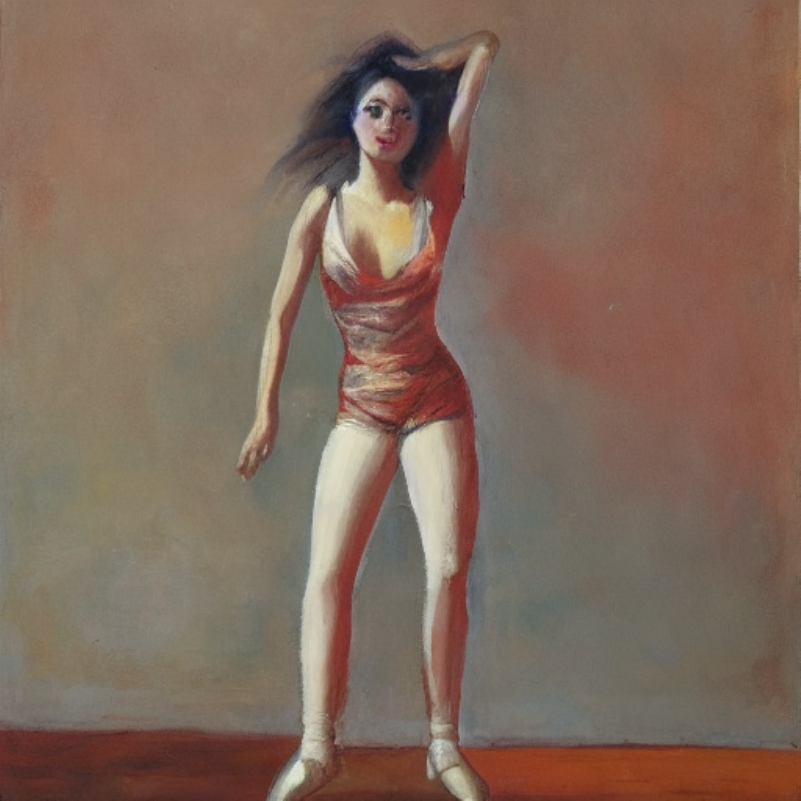}
  \end{subfigure}
  \begin{subfigure}[t]{.169\linewidth}
    \centering\includegraphics[width=\linewidth]{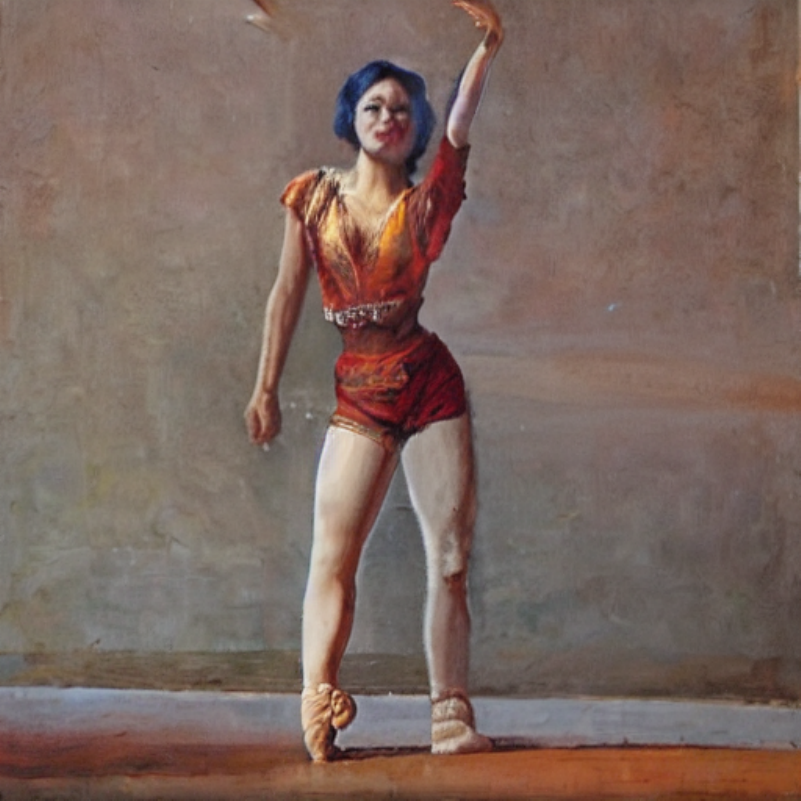}
  \end{subfigure}
  \begin{subfigure}[t]{.169\linewidth}
    \centering\includegraphics[width=\linewidth]{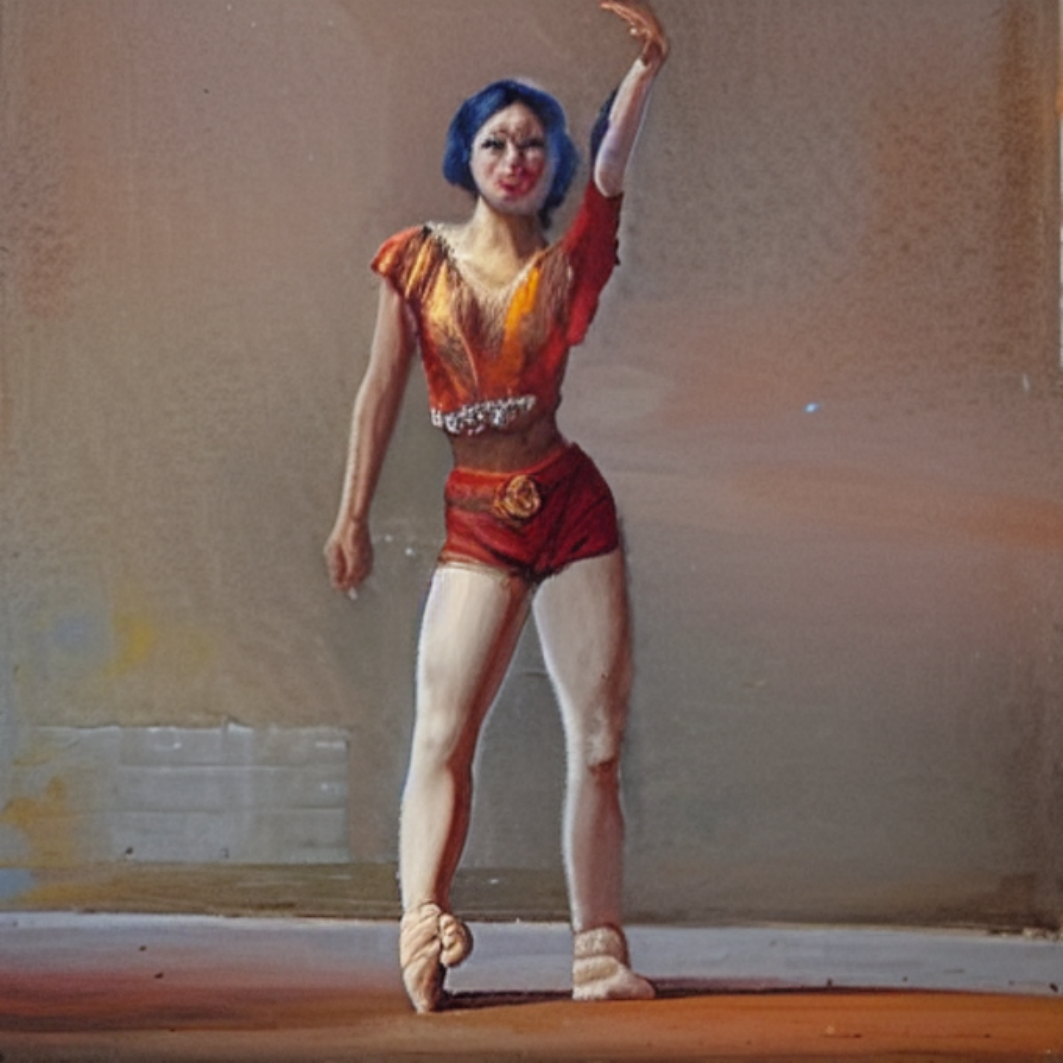}
  \end{subfigure}
  \begin{subfigure}[t]{.169\linewidth}
    \centering\includegraphics[width=\linewidth]{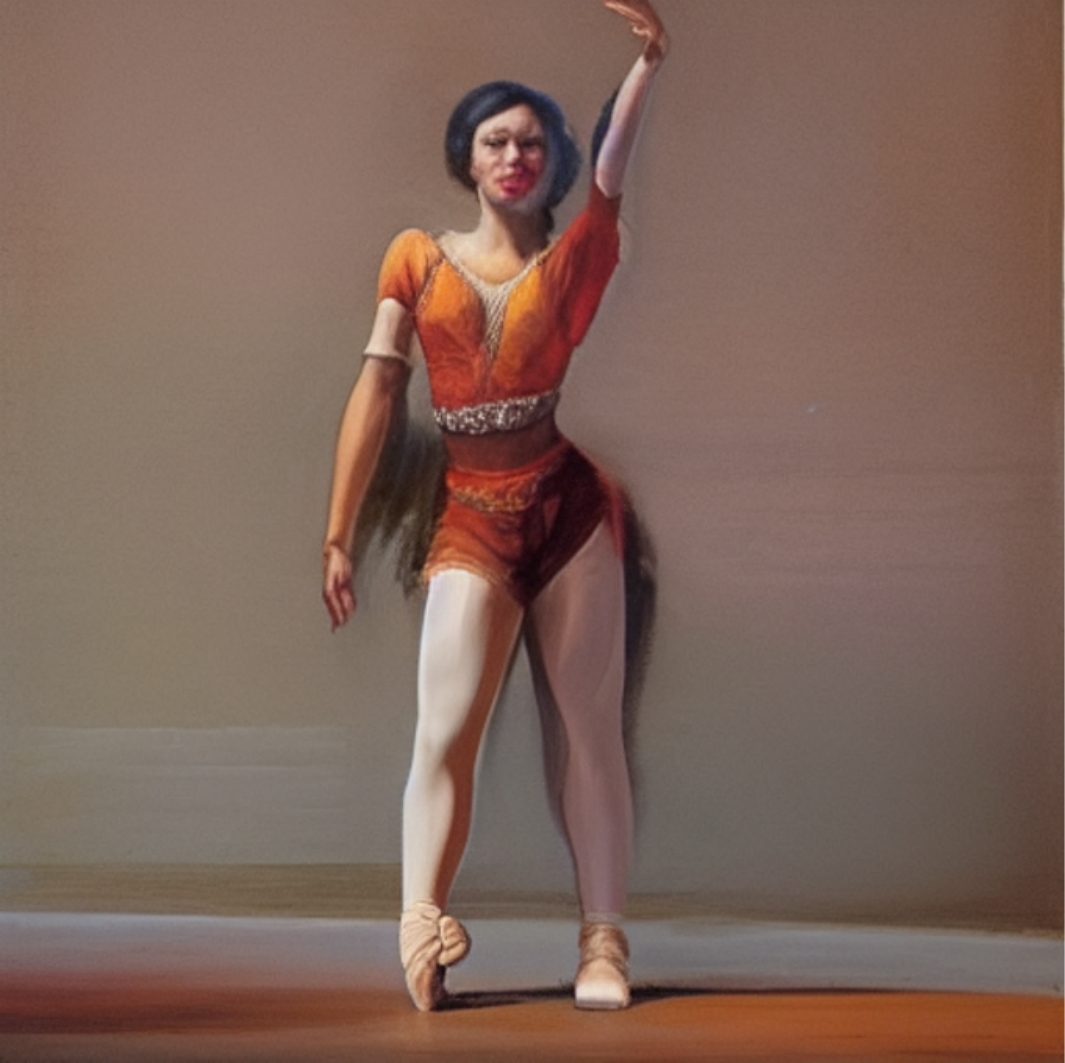}
  \end{subfigure}


    \begin{subfigure}[t]{.169\linewidth}
    \centering\includegraphics[width=\linewidth]{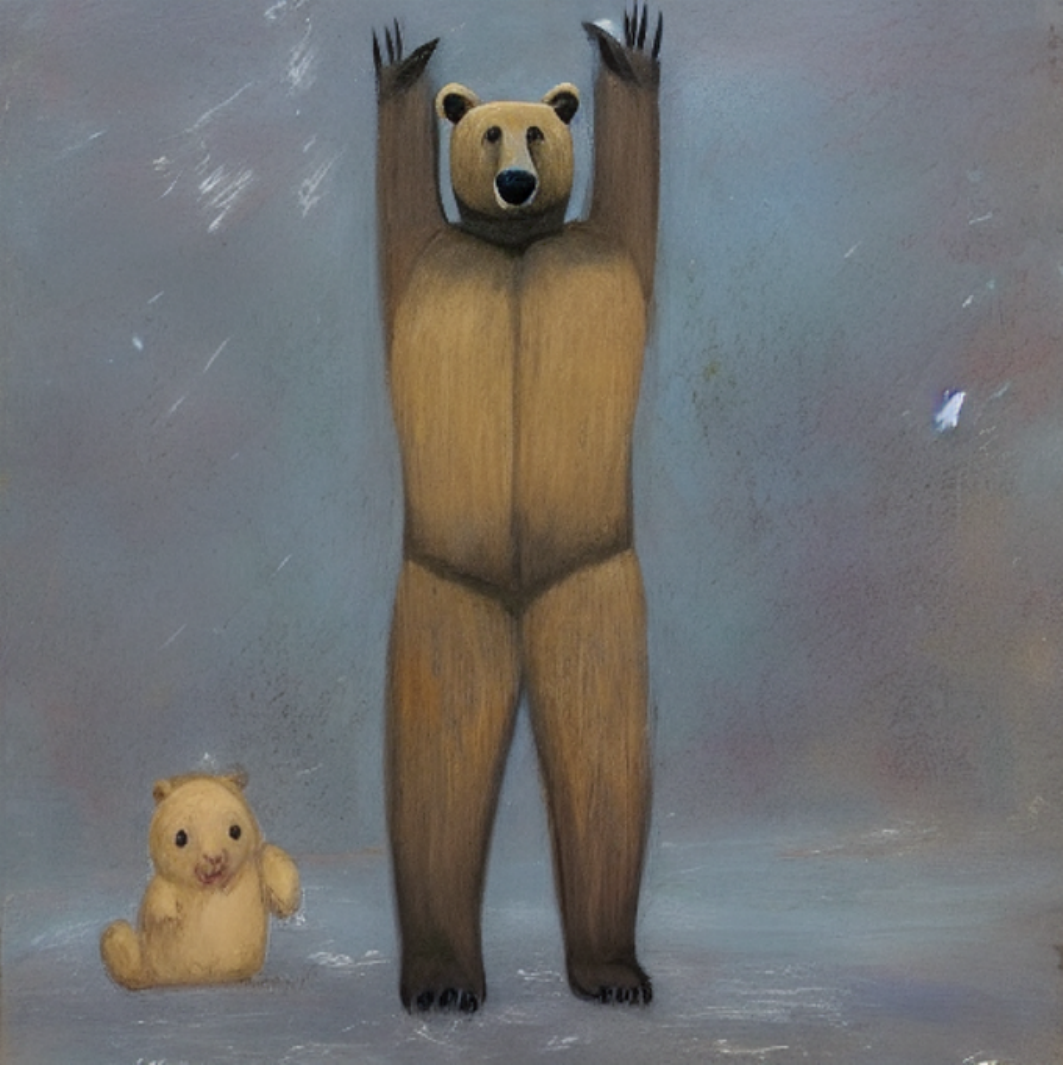}
    \end{subfigure}
      \begin{subfigure}[t]{.123\linewidth}
    \centering\includegraphics[width=\linewidth]{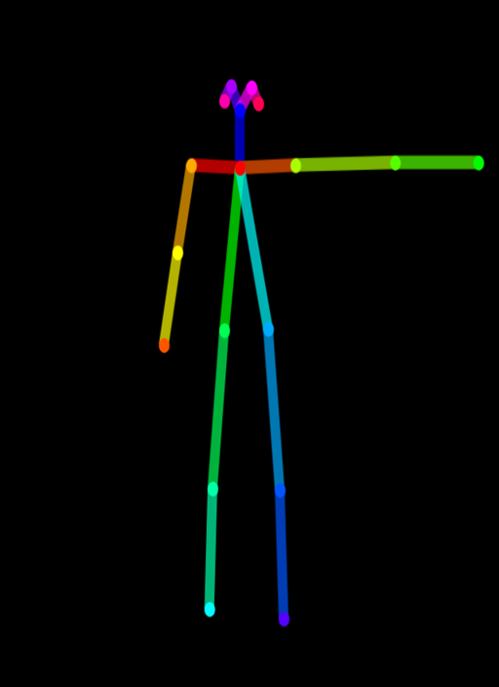}
    \end{subfigure}
  \begin{subfigure}[t]{.169\linewidth}
    \centering\includegraphics[width=\linewidth]{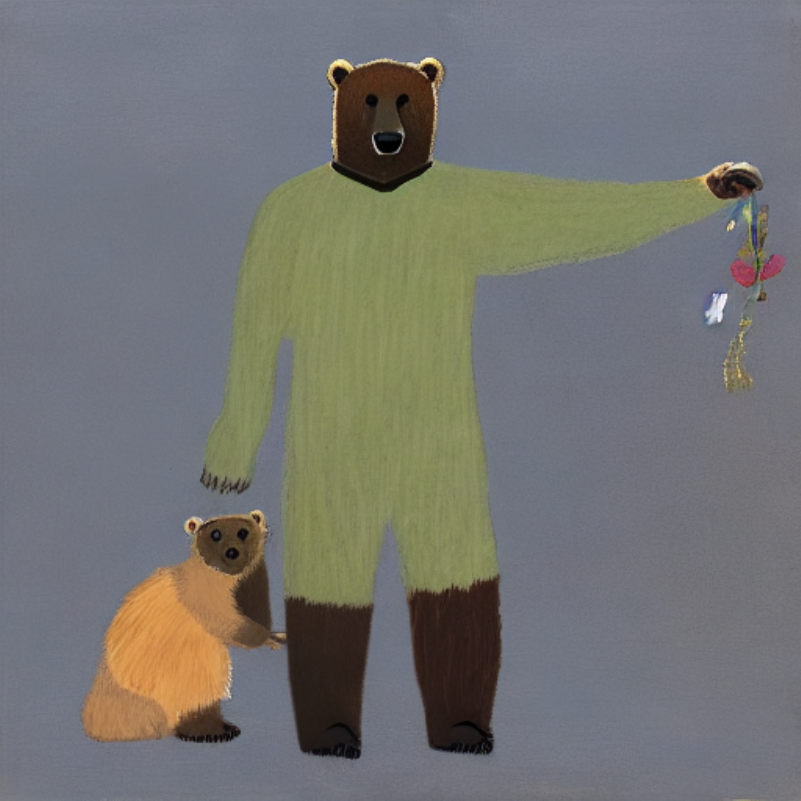}
  \end{subfigure}
  \begin{subfigure}[t]{.169\linewidth}
    \centering\includegraphics[width=\linewidth]{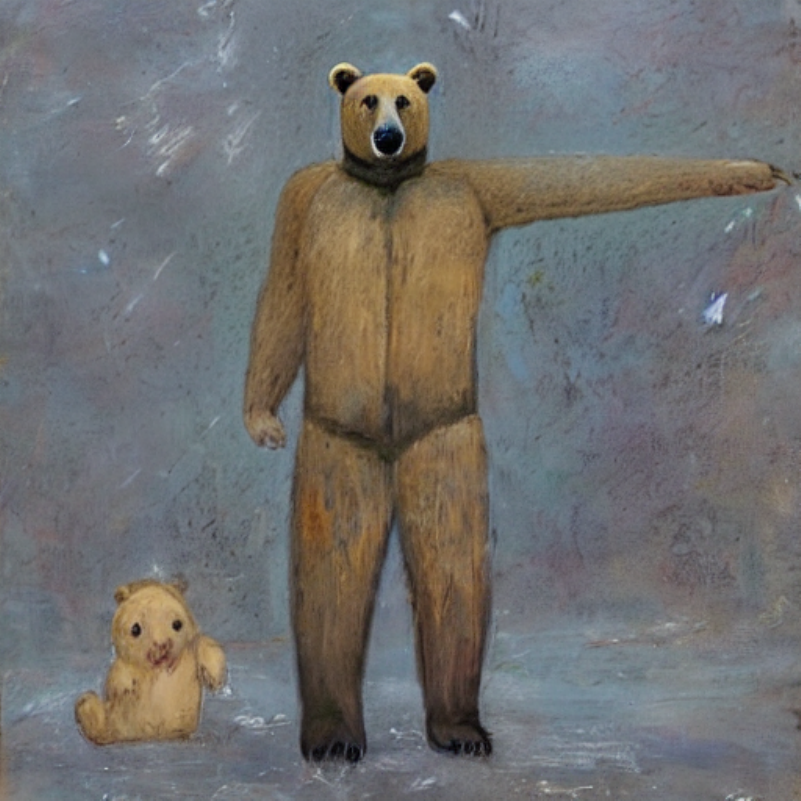}
  \end{subfigure}
  \begin{subfigure}[t]{.169\linewidth}
    \centering\includegraphics[width=\linewidth]{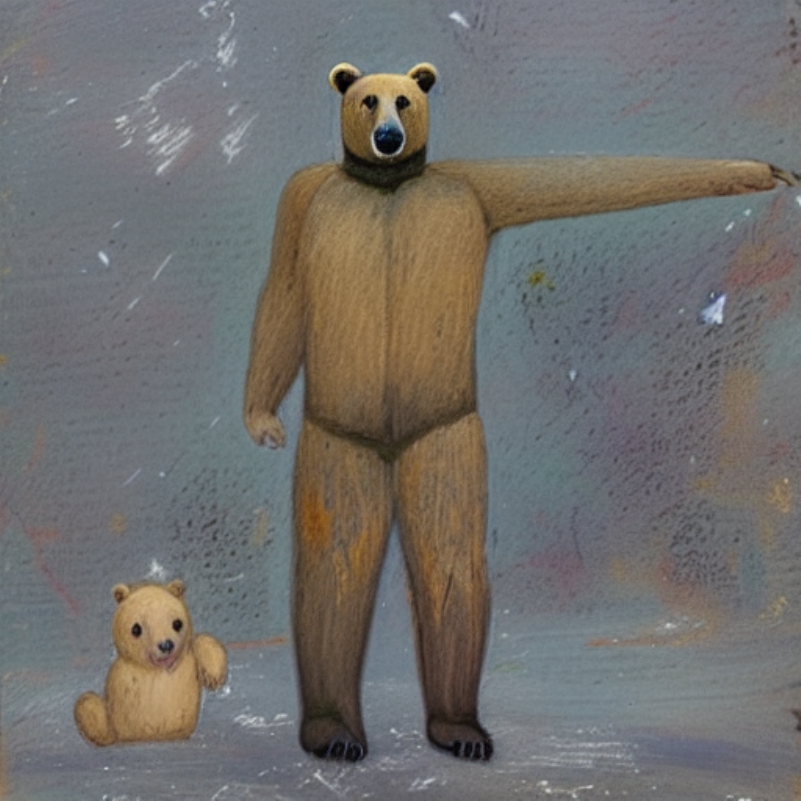}
  \end{subfigure}
  \begin{subfigure}[t]{.169\linewidth}
    \centering\includegraphics[width=\linewidth]{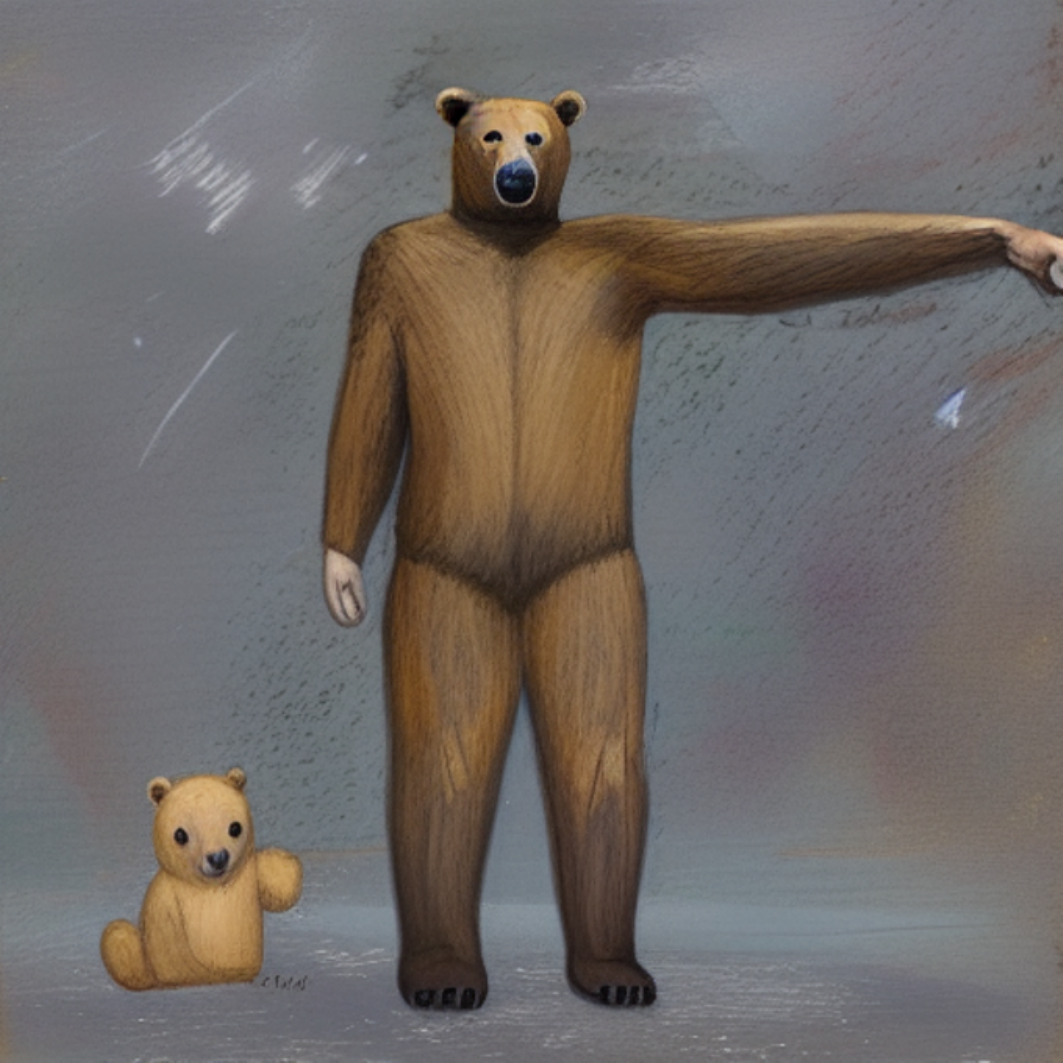}
  \end{subfigure}


    \begin{subfigure}[t]{.169\linewidth}
    \centering\includegraphics[width=\linewidth]{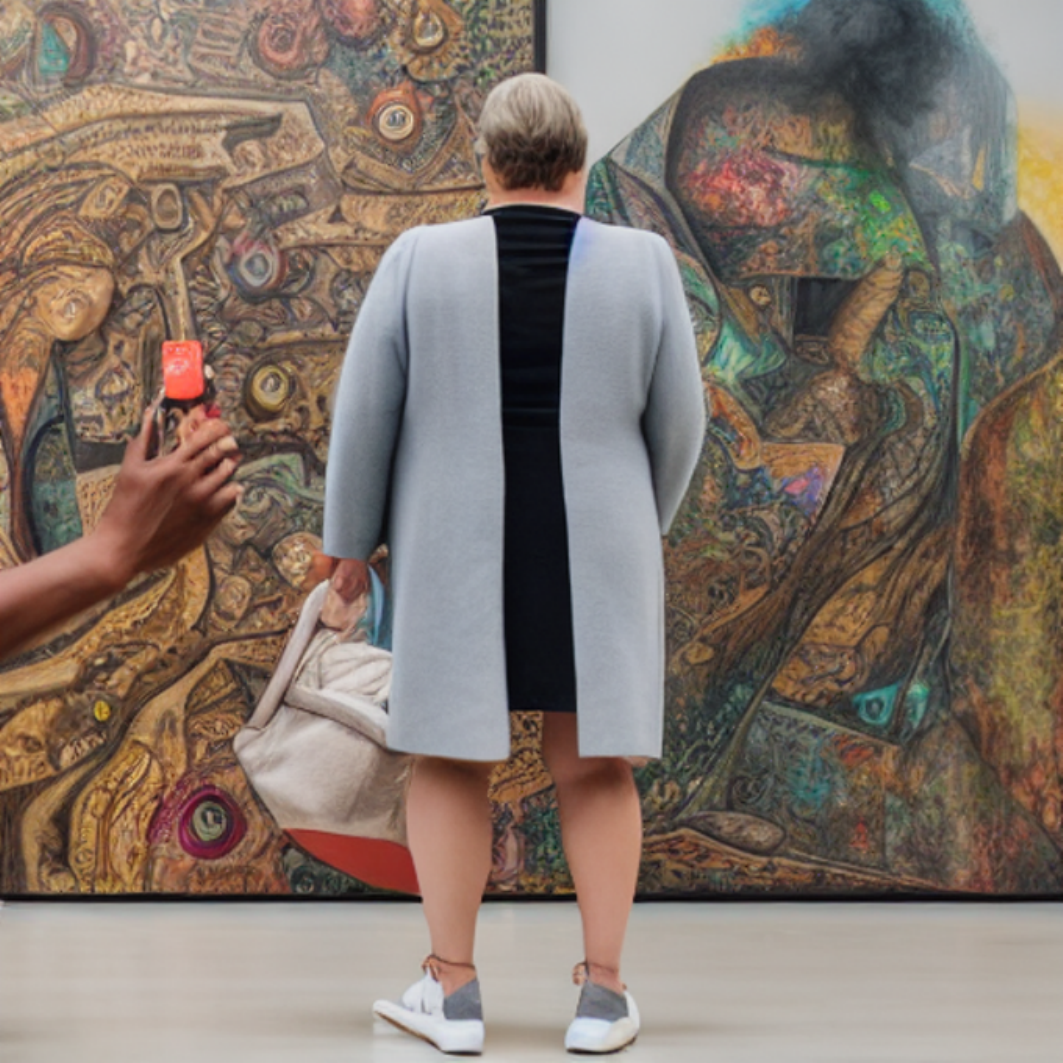}
    \end{subfigure}
      \begin{subfigure}[t]{.123\linewidth}
    \centering\includegraphics[width=\linewidth]{imgs/results/pose3.png}
    \end{subfigure}
  \begin{subfigure}[t]{.169\linewidth}
    \centering\includegraphics[width=\linewidth]{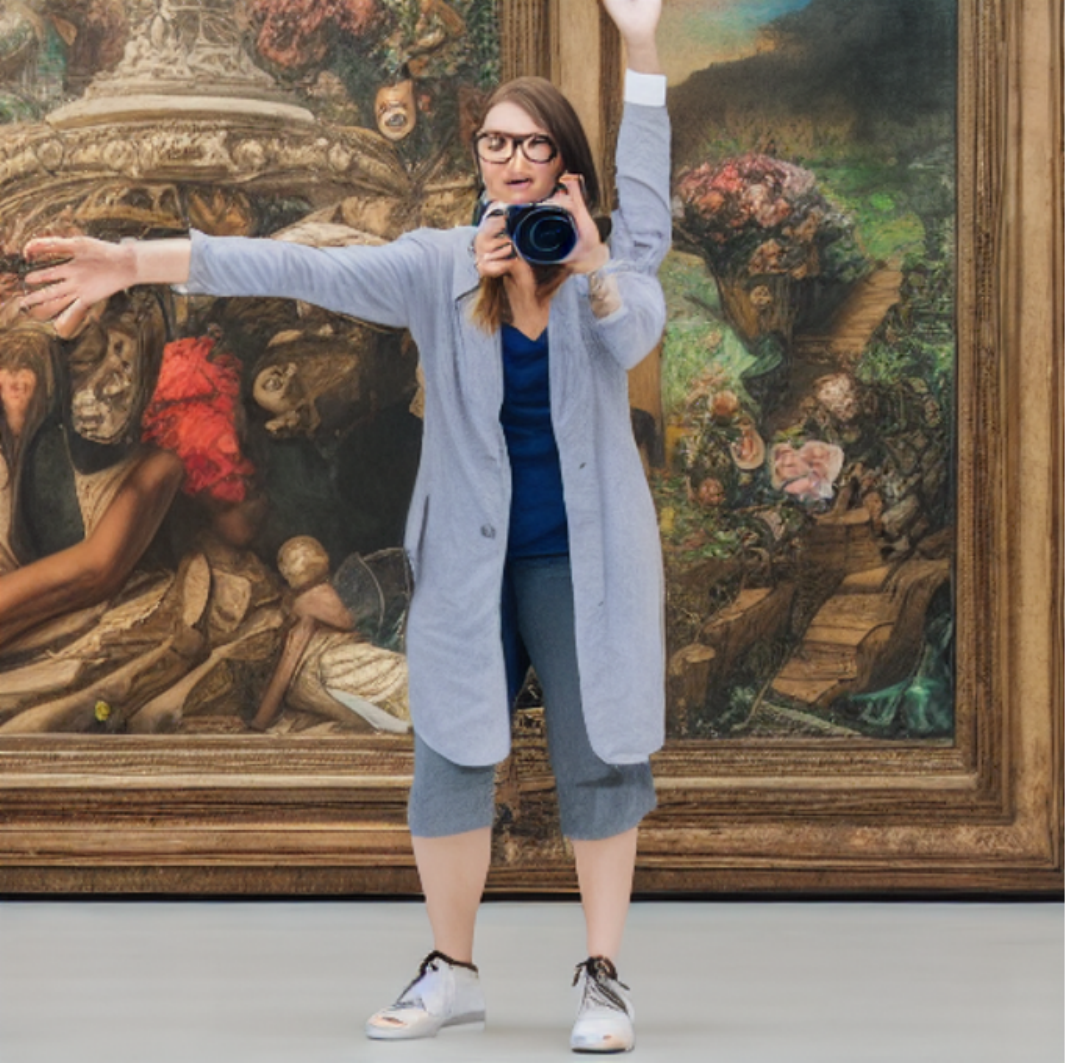}
  \end{subfigure}
  \begin{subfigure}[t]{.169\linewidth}
    \centering\includegraphics[width=\linewidth]{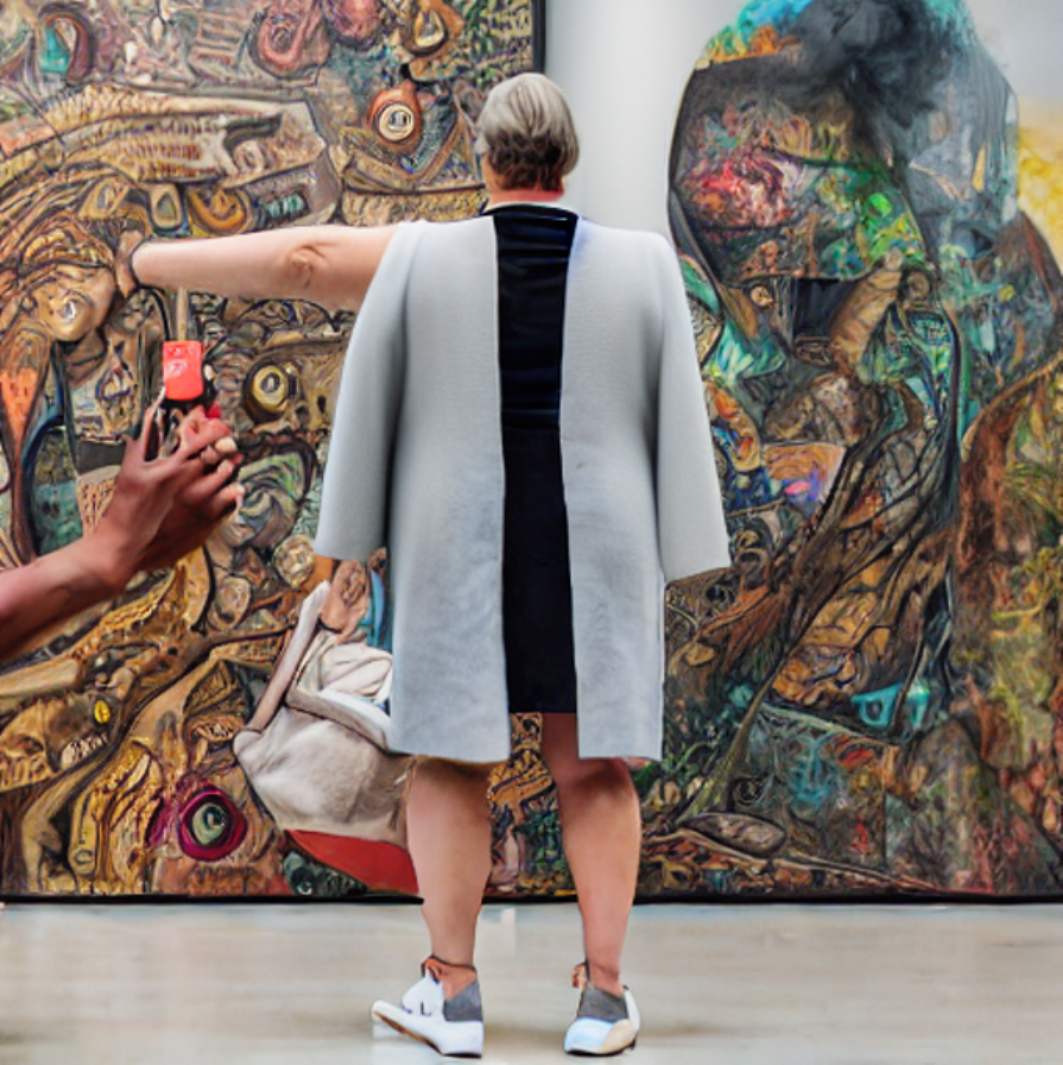}
  \end{subfigure}
  \begin{subfigure}[t]{.169\linewidth}
    \centering\includegraphics[width=\linewidth]{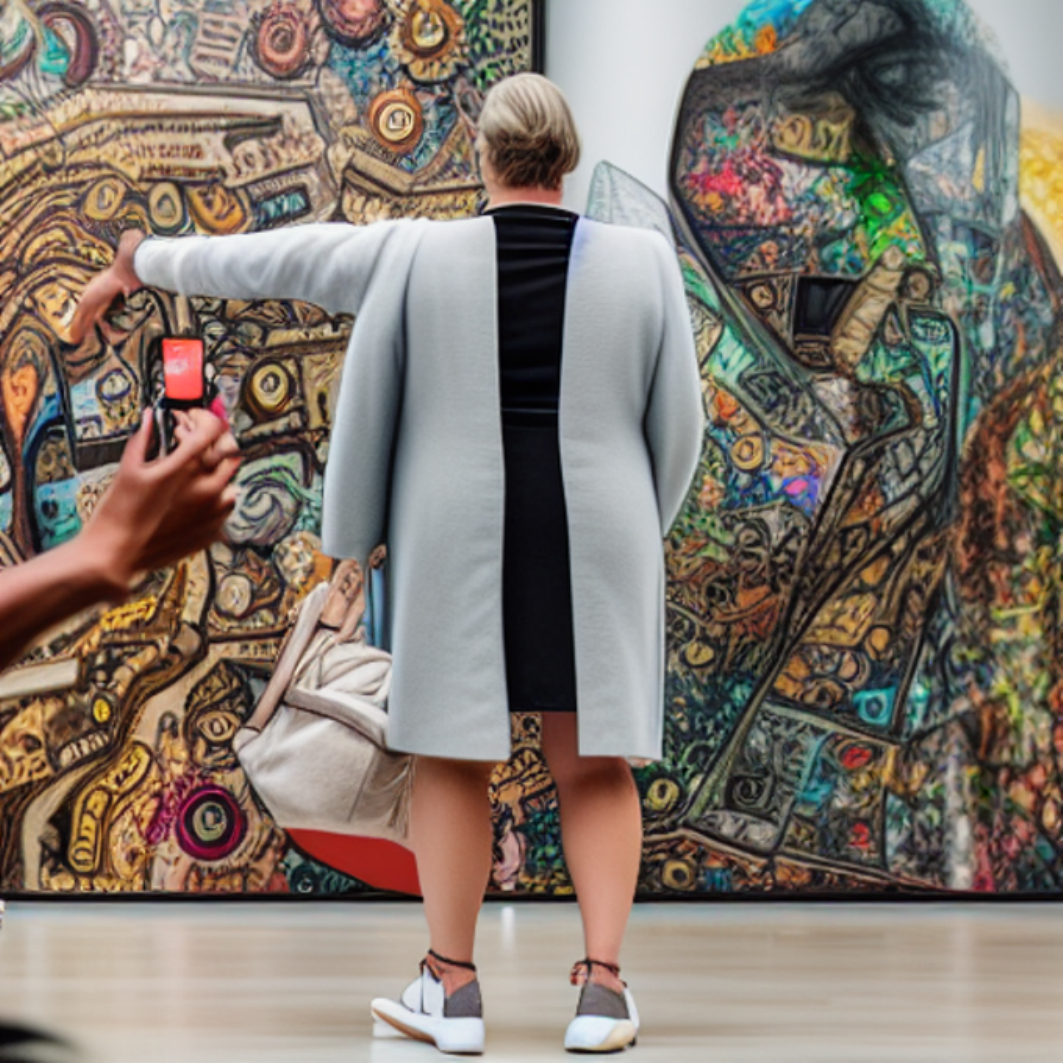}
  \end{subfigure}
  \begin{subfigure}[t]{.169\linewidth}
    \centering\includegraphics[width=\linewidth]{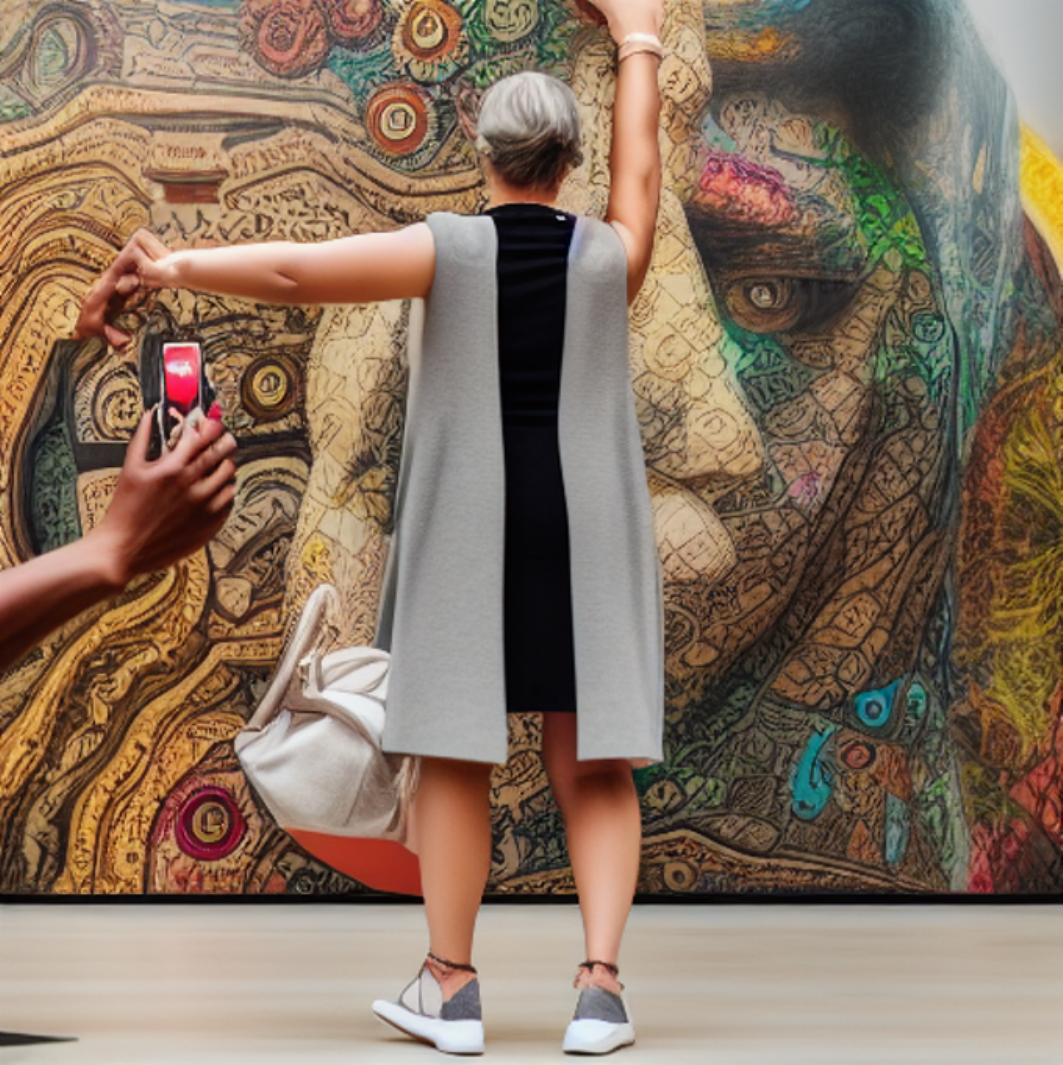}
  \end{subfigure}


  \begin{subfigure}[t]{.169\linewidth}
    \centering\includegraphics[width=\linewidth]{imgs/results/astronaut_pose0.pdf}
    \caption{Original image}
    \end{subfigure}
      \begin{subfigure}[t]{.123\linewidth}
    \centering\includegraphics[width=\linewidth]{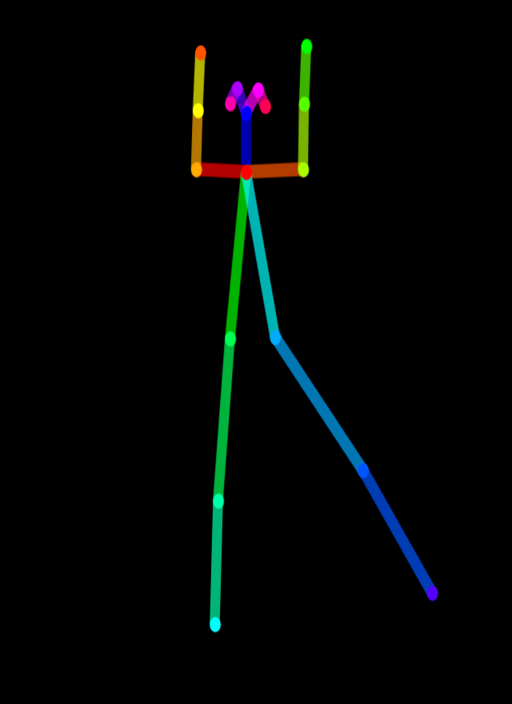}
    \caption{Target pose}
    \end{subfigure}
  \begin{subfigure}[t]{.169\linewidth}
    \centering\includegraphics[width=\linewidth]{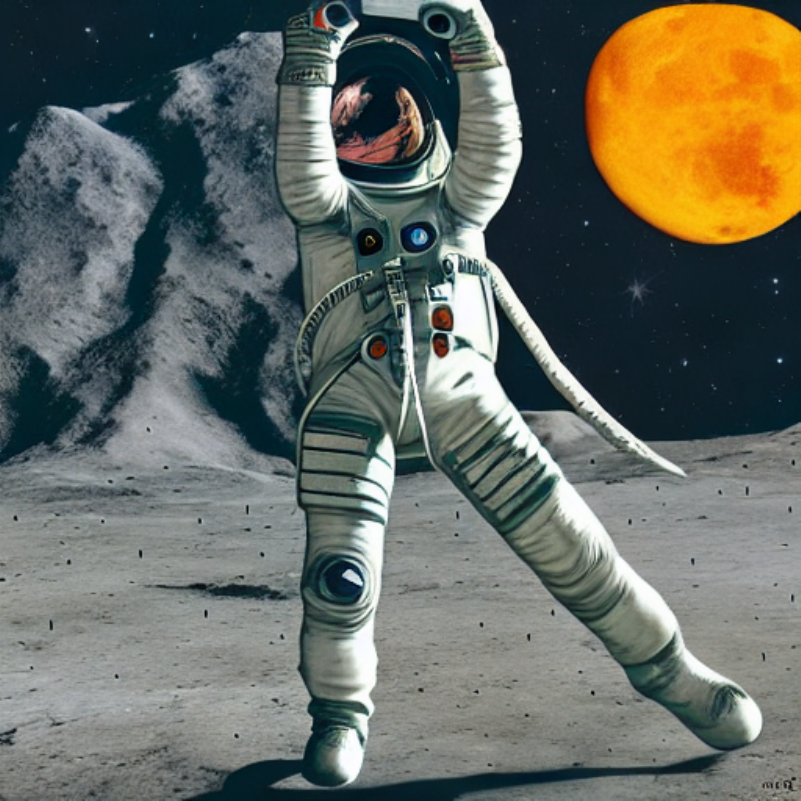}
    \caption{ControlNet}
  \end{subfigure}
  \begin{subfigure}[t]{.169\linewidth}
    \centering\includegraphics[width=\linewidth]{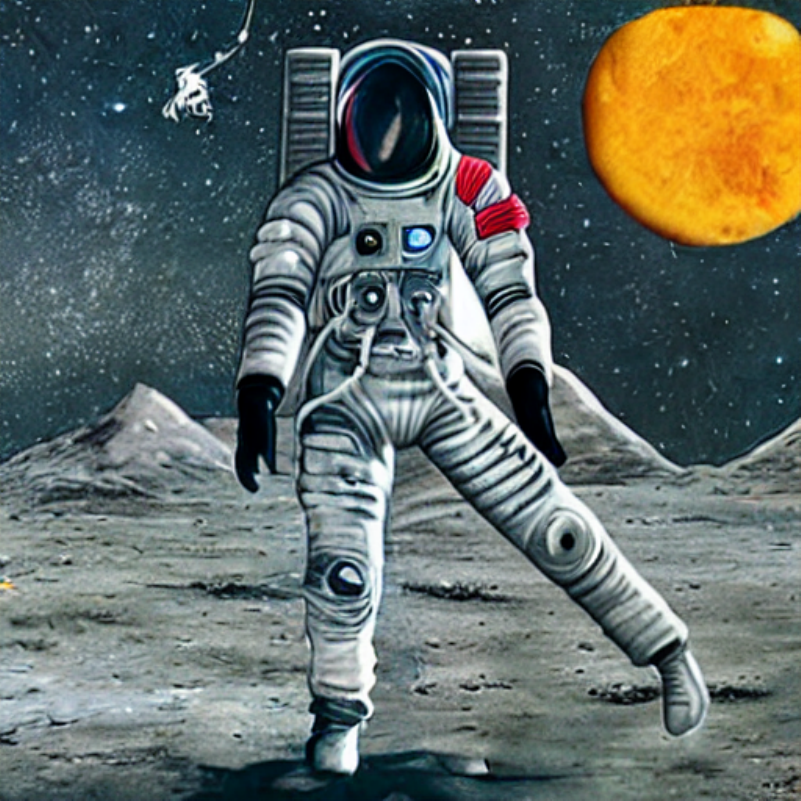}
    \caption{DiffEdit + our mask}
  \end{subfigure}
  \begin{subfigure}[t]{.169\linewidth}
    \centering\includegraphics[width=\linewidth]{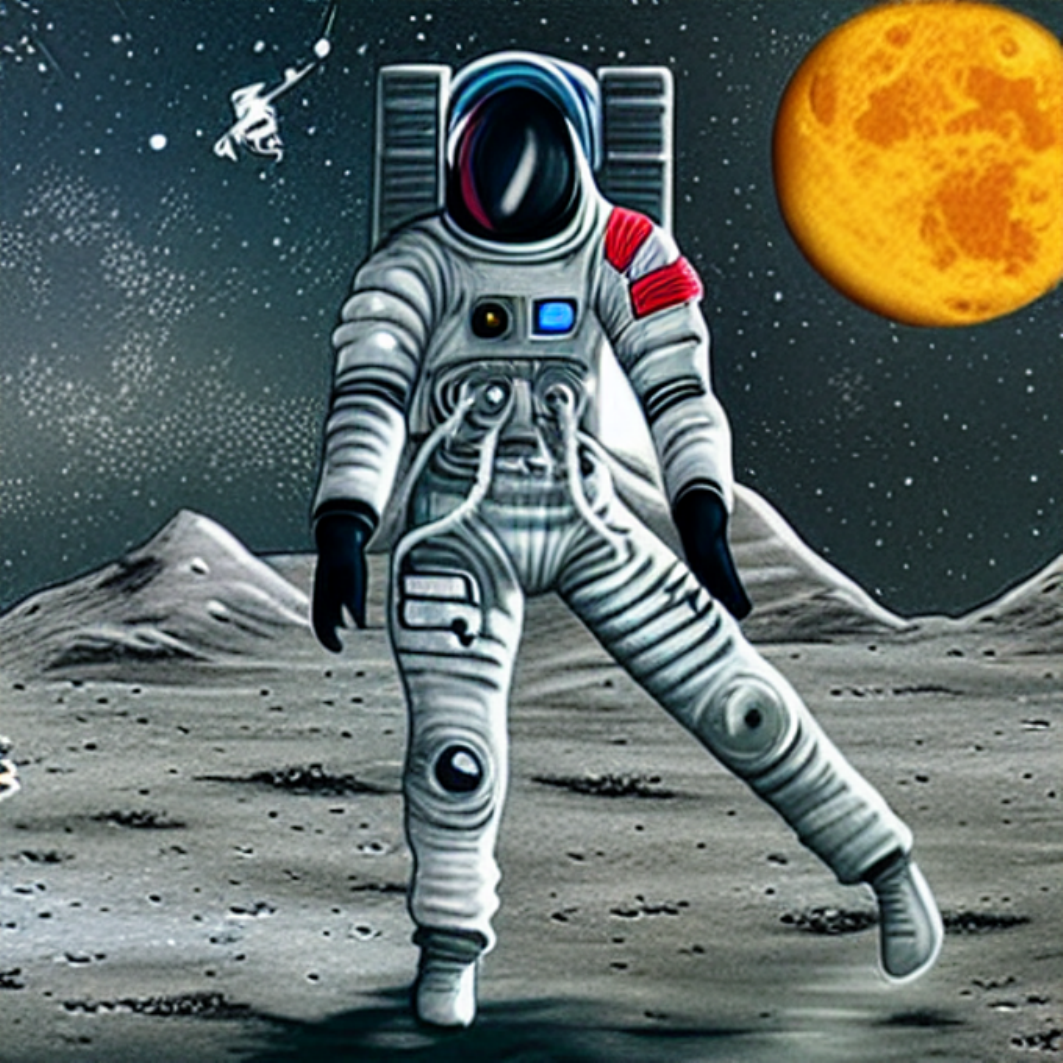}
    \caption{Ours ($\lambda=1$)}
  \end{subfigure}
  \begin{subfigure}[t]{.169\linewidth}
    \centering\includegraphics[width=\linewidth]{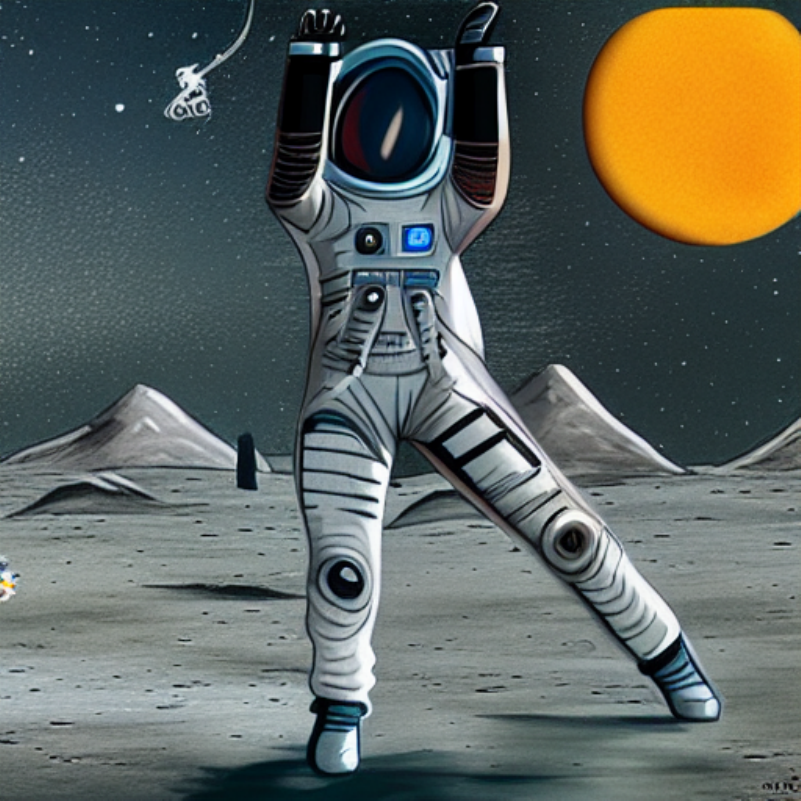}
    \caption{Ours}
  \end{subfigure}
\end{adjustbox}
\caption{ Visual results for pose based editing. }
\label{fig:results:pose}

\end{figure*}

\begin{table}[t]
\begin{adjustbox}{minipage=0.45\textwidth,scale=0.8}
\centering
\begin{tabular}{l|l|l|l}
\toprule
Metric                                                                   & DiffEdit    & Prompt-to-prompt & Ours  \\\midrule
L1 distance \textdownarrow                                               & \bf0.124    & 0.128            & 0.136 \\
CSFID   \textdownarrow                                                   & 67.7        & 74.2    & \bf 57.2  \\
\begin{tabular}{@{}l@{}}Classification accuracy\\ after edit (new category) \textuparrow \end{tabular} 
         & 37.8        & 34.2   & \bf 60.1 \\
   \begin{tabular}{@{}l@{}}Classification accuracy\\ after edit (original category) \textdownarrow\end{tabular}  & 45.9        &  45.5  & \bf20.0 \\
Pickscore \textuparrow                                                   & 0.31 &  0.27  & \bf  0.41 \\
\bottomrule
\end{tabular}
\end{adjustbox}
\caption{Quantitative comparison of text-guided image editing on the ImageNet dataset. }
\label{tab:results}
\end{table}


\begin{figure*}[ht]
    \centering
    \begin{adjustbox}{minipage=\linewidth,scale=0.835}
    
    \begin{subfigure}[t]{.162\linewidth}
    \centering\includegraphics[width=\linewidth]{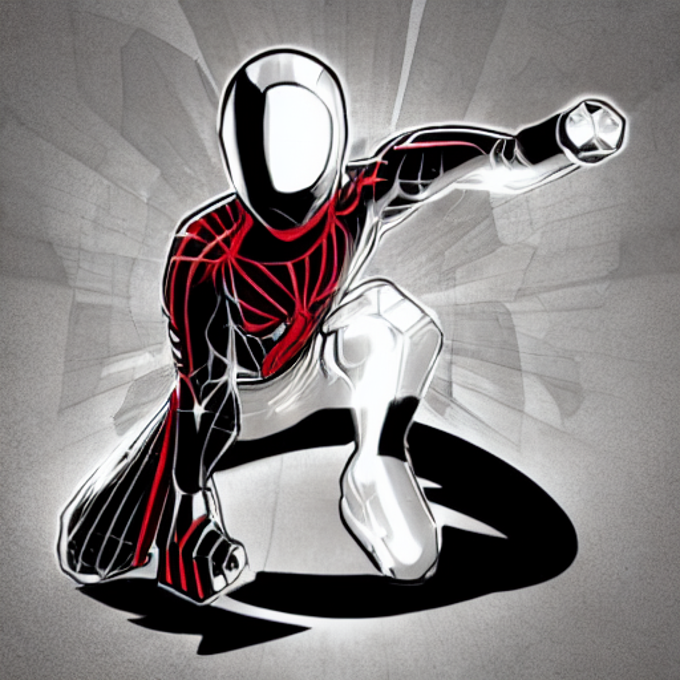}
    \end{subfigure}
      \begin{subfigure}[t]{.152\linewidth}
    \centering\includegraphics[width=\linewidth]{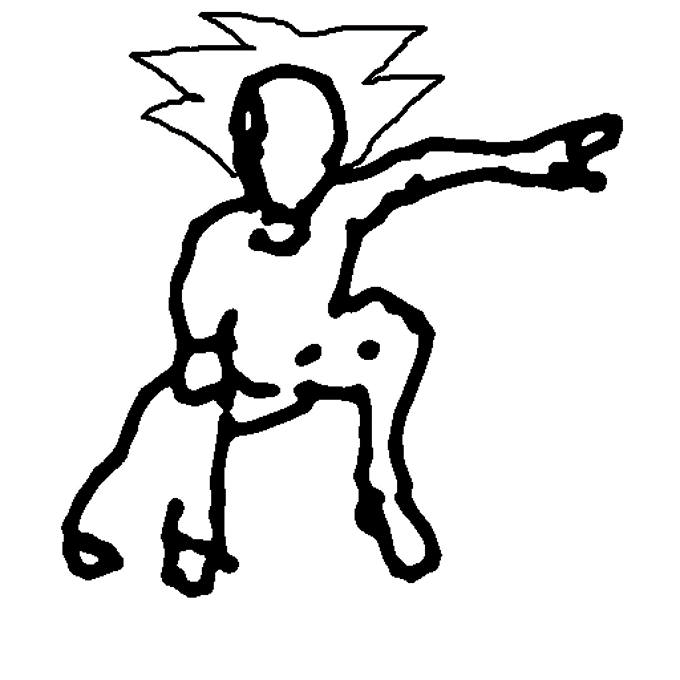}
    \end{subfigure}
  \begin{subfigure}[t]{.162\linewidth}
    \centering\includegraphics[width=\linewidth]{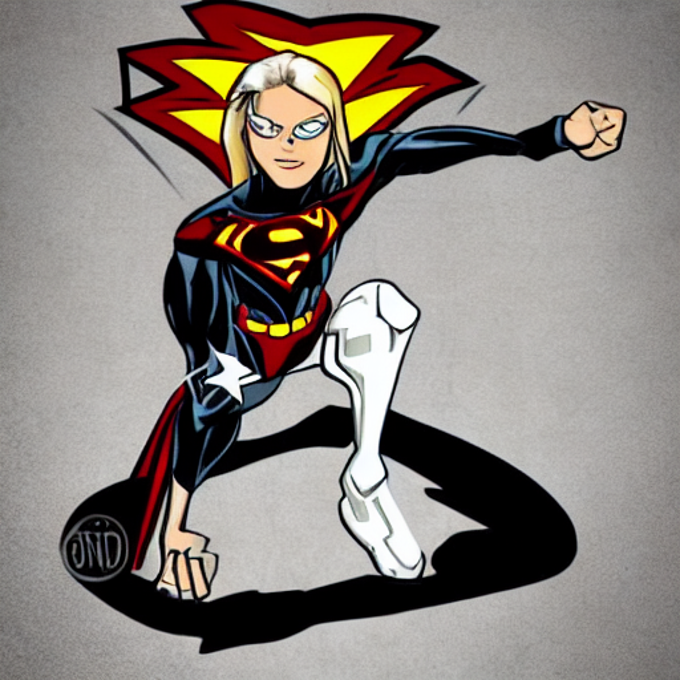}
  \end{subfigure}
  \begin{subfigure}[t]{.162\linewidth}
    \centering\includegraphics[width=\linewidth]{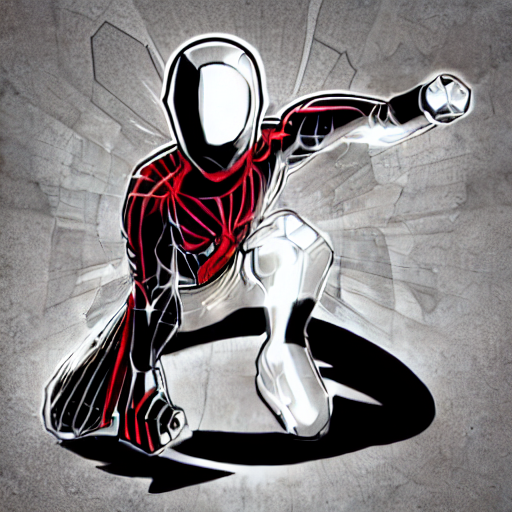}
  \end{subfigure}
  \begin{subfigure}[t]{.162\linewidth}
    \centering\includegraphics[width=\linewidth]{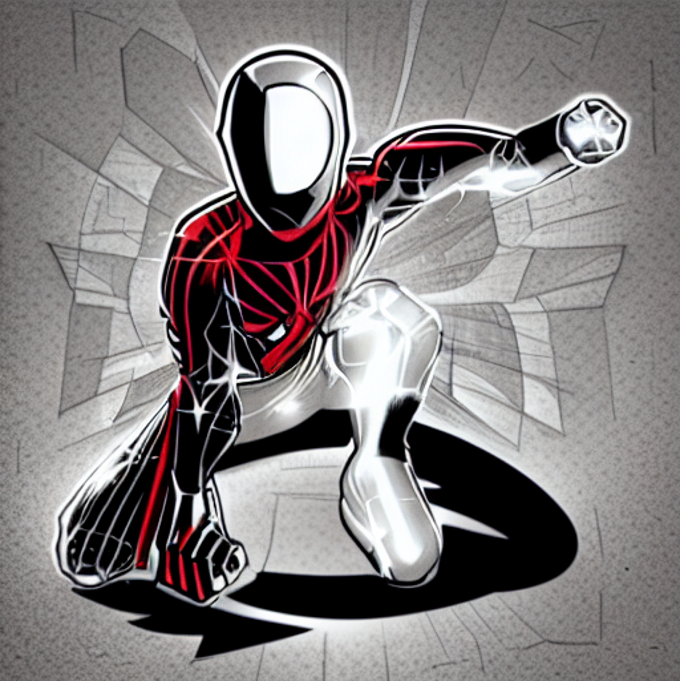}
  \end{subfigure}
  \begin{subfigure}[t]{.162\linewidth}
    \centering\includegraphics[width=\linewidth]{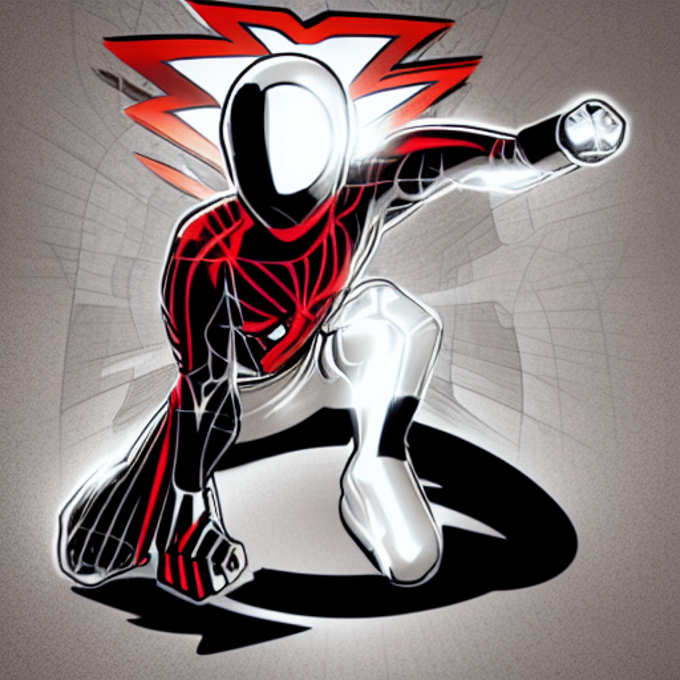}
  \end{subfigure}
  

\centering
    \begin{subfigure}[t]{.162\linewidth}
    \centering\includegraphics[width=\linewidth]{imgs/results/superhero.png}
    \end{subfigure}
      \begin{subfigure}[t]{.152\linewidth}
    \centering\includegraphics[width=\linewidth]{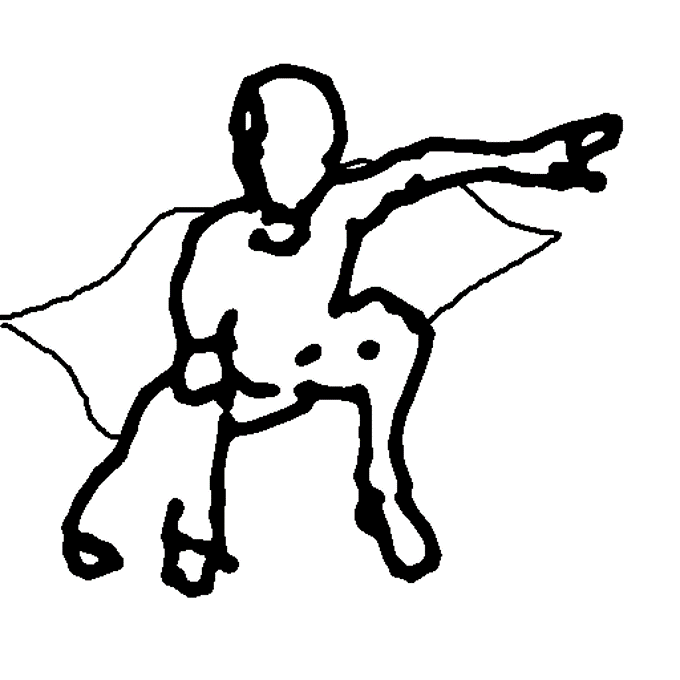}
    \end{subfigure}
  \begin{subfigure}[t]{.162\linewidth}
    \centering\includegraphics[width=\linewidth]{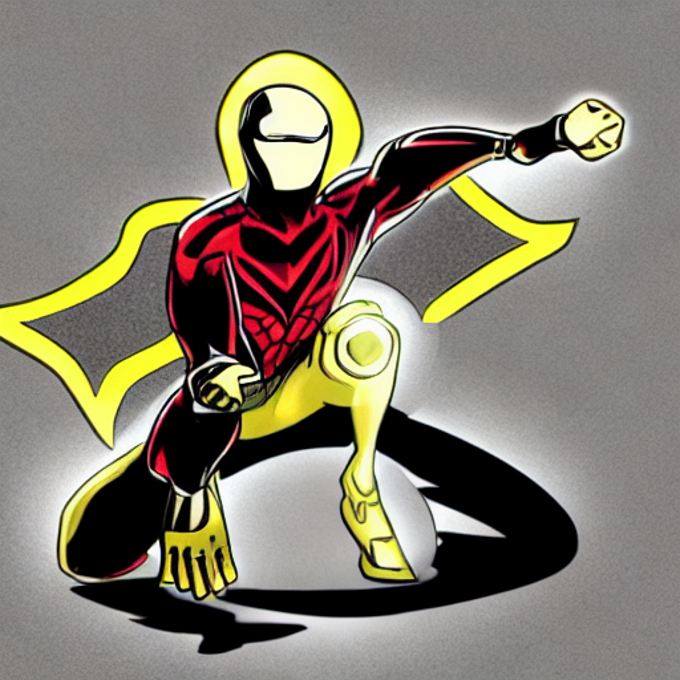}
  \end{subfigure}
  \begin{subfigure}[t]{.162\linewidth}
    \centering\includegraphics[width=\linewidth]{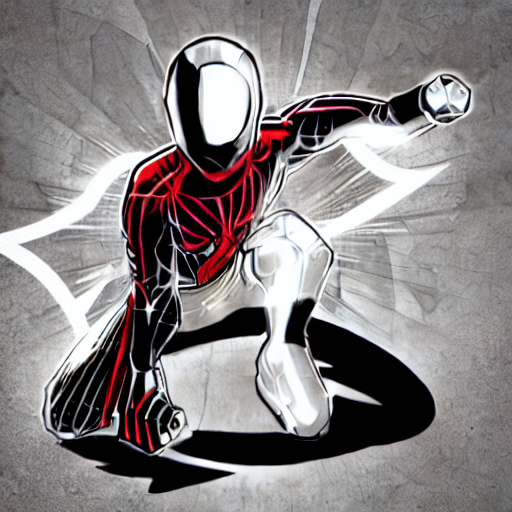}
  \end{subfigure}
  \begin{subfigure}[t]{.162\linewidth}
    \centering\includegraphics[width=\linewidth]{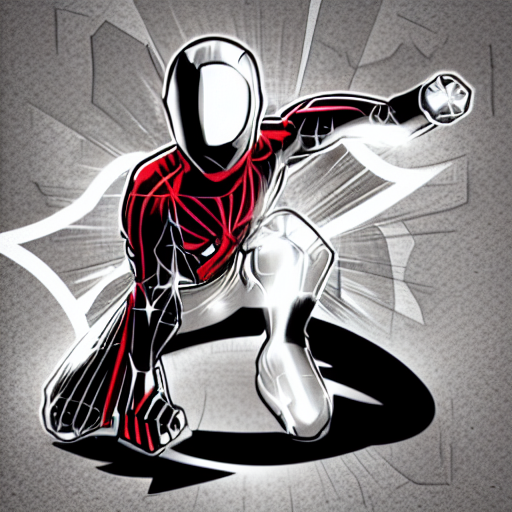}
  \end{subfigure}
  \begin{subfigure}[t]{.162\linewidth}
    \centering\includegraphics[width=\linewidth]{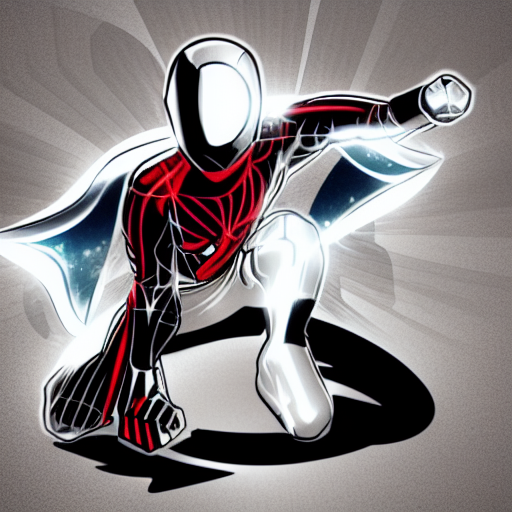}
  \end{subfigure}


\centering
    \begin{subfigure}[t]{.162\linewidth}
    \centering\includegraphics[width=\linewidth]{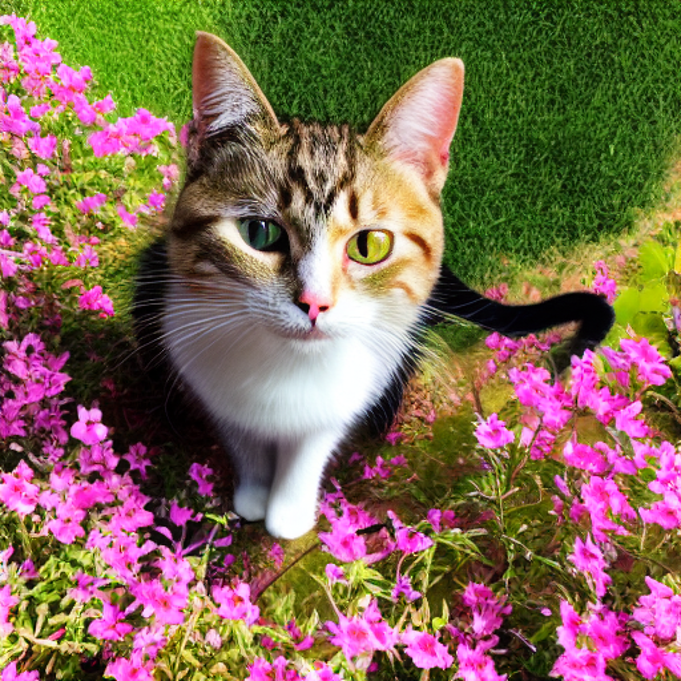}
    \end{subfigure}
      \begin{subfigure}[t]{.152\linewidth}
    \centering\includegraphics[width=\linewidth]{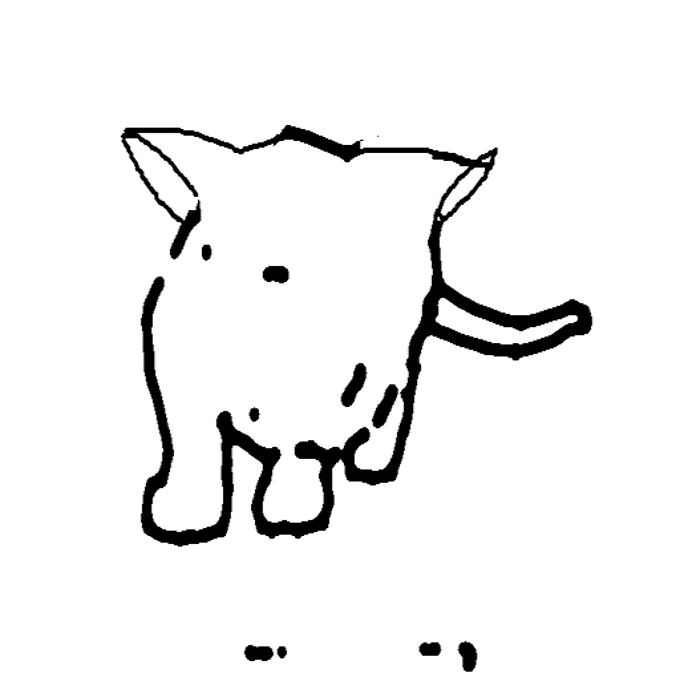}
    \end{subfigure}
  \begin{subfigure}[t]{.162\linewidth}
    \centering\includegraphics[width=\linewidth]{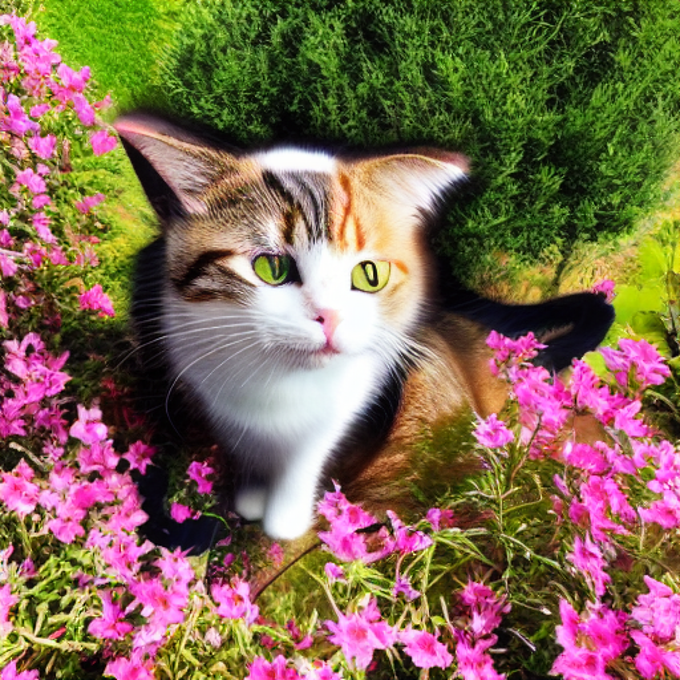}
  \end{subfigure}
  \begin{subfigure}[t]{.162\linewidth}
    \centering\includegraphics[width=\linewidth]{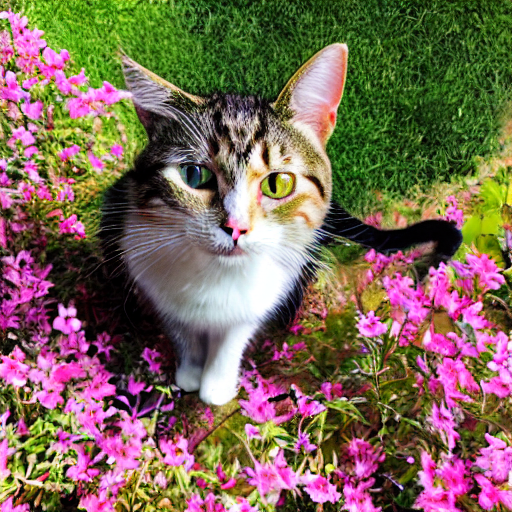}
  \end{subfigure}
  \begin{subfigure}[t]{.162\linewidth}
    \centering\includegraphics[width=\linewidth]{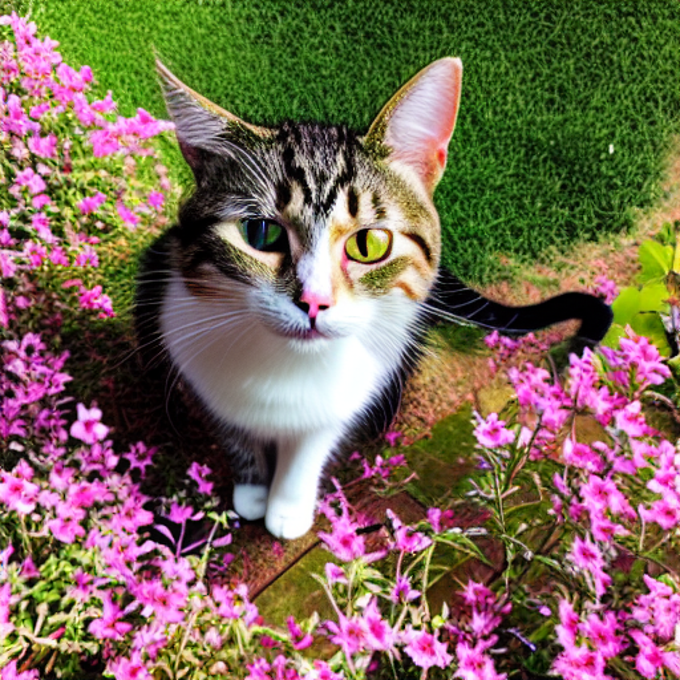}
  \end{subfigure}
  \begin{subfigure}[t]{.162\linewidth}
    \centering\includegraphics[width=\linewidth]{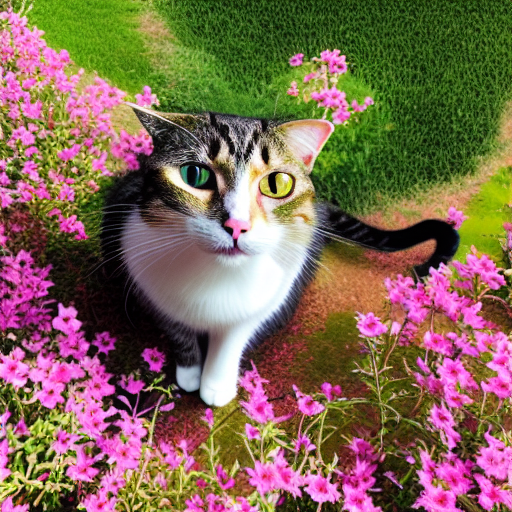}
  \end{subfigure}


\centering
    \begin{subfigure}[t]{.162\linewidth}
    \centering\includegraphics[width=\linewidth]{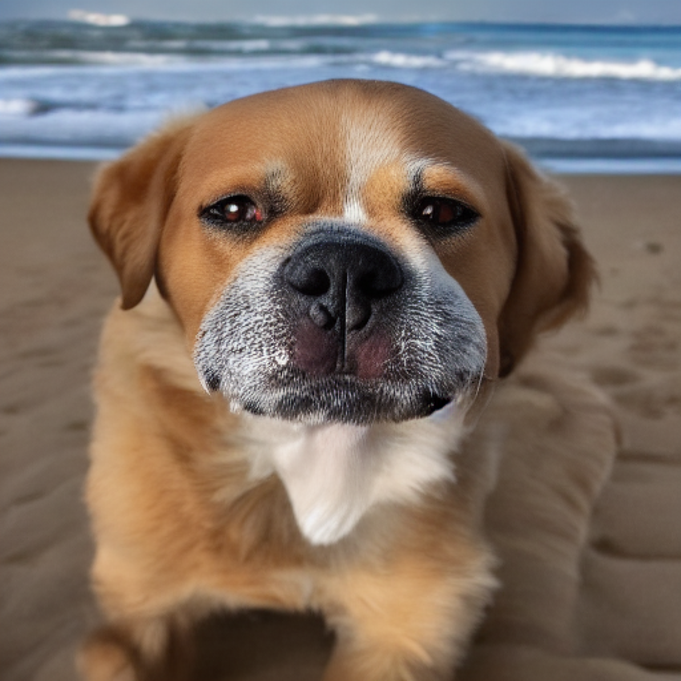}
    \end{subfigure}
      \begin{subfigure}[t]{.152\linewidth}
    \centering\includegraphics[width=\linewidth]{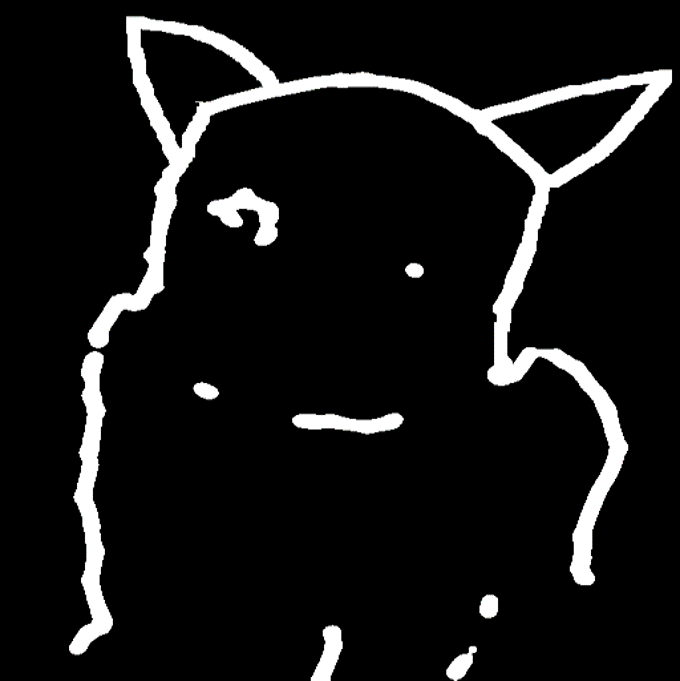}
    \end{subfigure}
  \begin{subfigure}[t]{.162\linewidth}
    \centering\includegraphics[width=\linewidth]{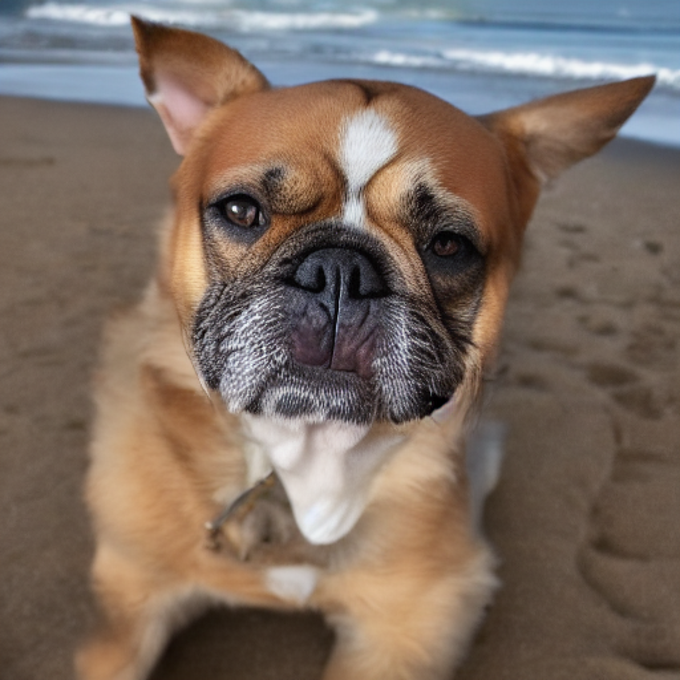}
  \end{subfigure}
  \begin{subfigure}[t]{.162\linewidth}
    \centering\includegraphics[width=\linewidth]{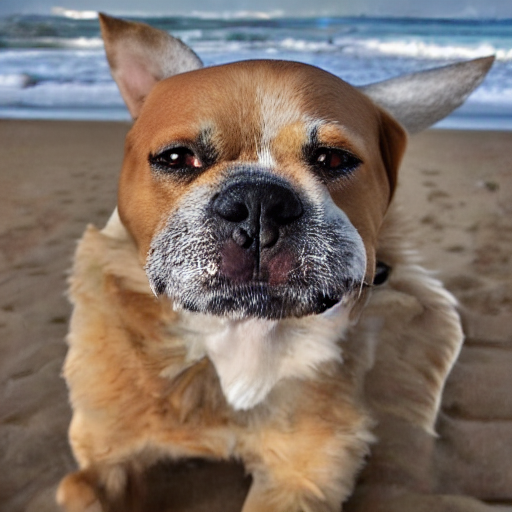}
  \end{subfigure}
  \begin{subfigure}[t]{.162\linewidth}
    \centering\includegraphics[width=\linewidth]{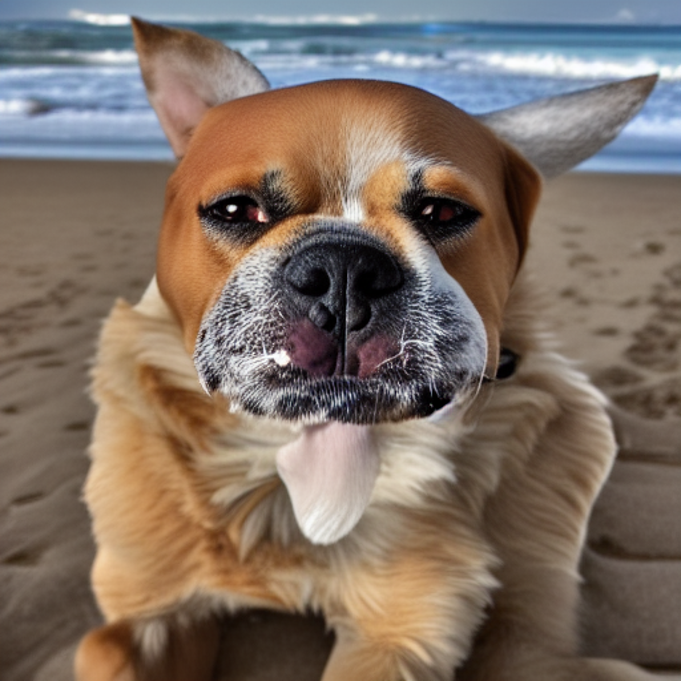}
  \end{subfigure}
  \begin{subfigure}[t]{.162\linewidth}
    \centering\includegraphics[width=\linewidth]{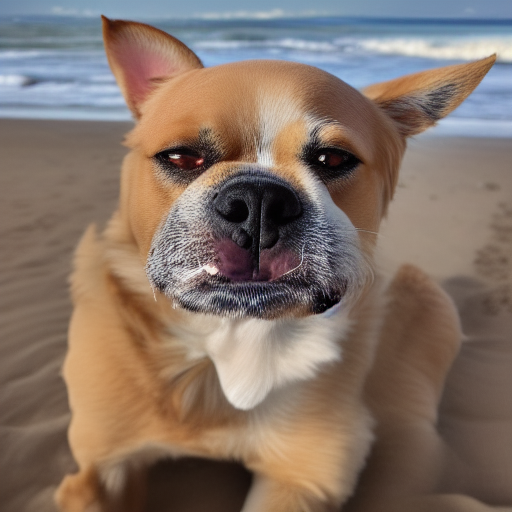}
  \end{subfigure}


  \begin{subfigure}[t]{.162\linewidth}
    \centering\includegraphics[width=\linewidth]{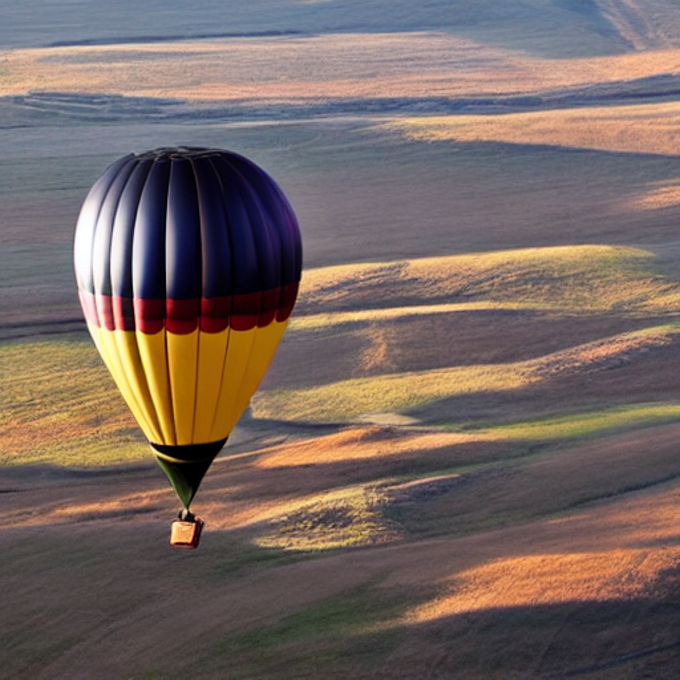}
    \caption{Original image}
    \end{subfigure}
      \begin{subfigure}[t]{.152\linewidth}
    \centering\includegraphics[width=\linewidth]{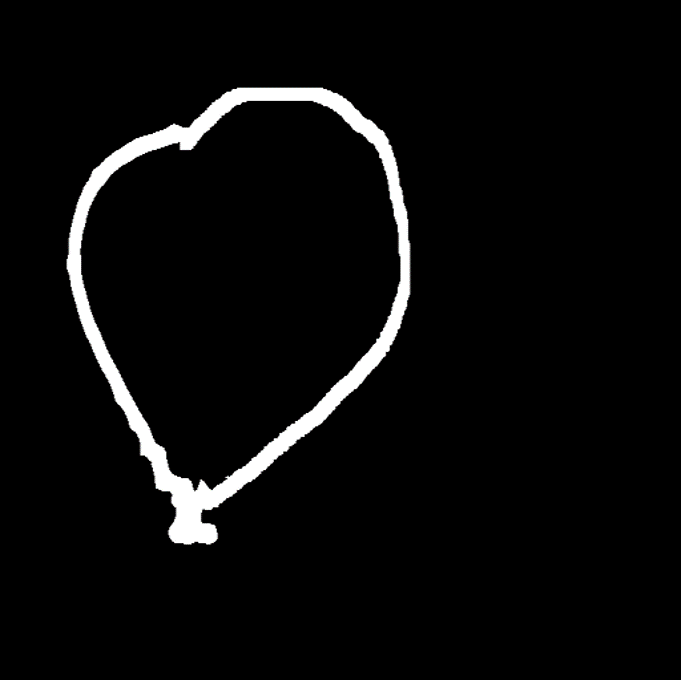}
    \caption{Target scribble}
    \end{subfigure}
  \begin{subfigure}[t]{.162\linewidth}
    \centering\includegraphics[width=\linewidth]{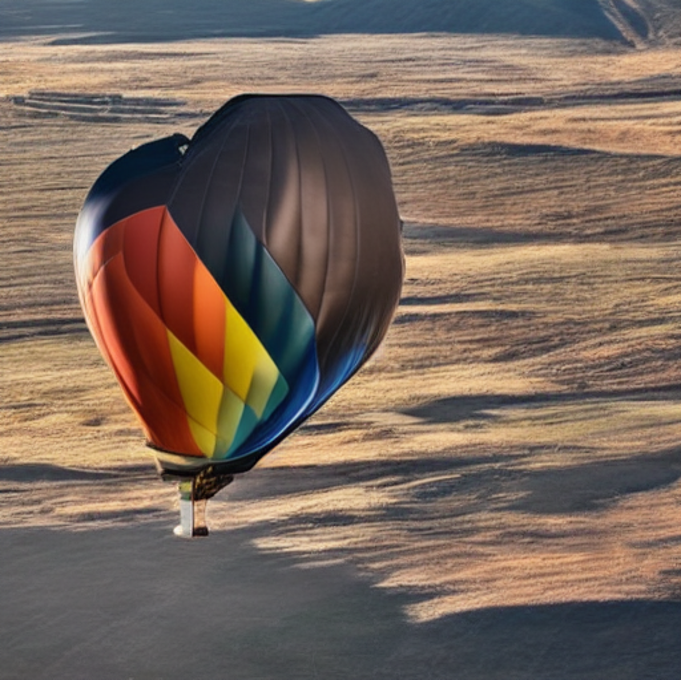}
    \caption{ControlNet}
  \end{subfigure}
  \begin{subfigure}[t]{.162\linewidth}
    \centering\includegraphics[width=\linewidth]{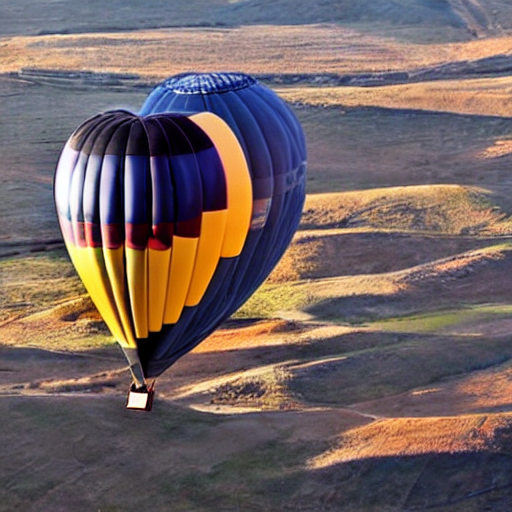}
    \caption{DiffEdit + our mask}
  \end{subfigure}
  \begin{subfigure}[t]{.162\linewidth}
    \centering\includegraphics[width=\linewidth]{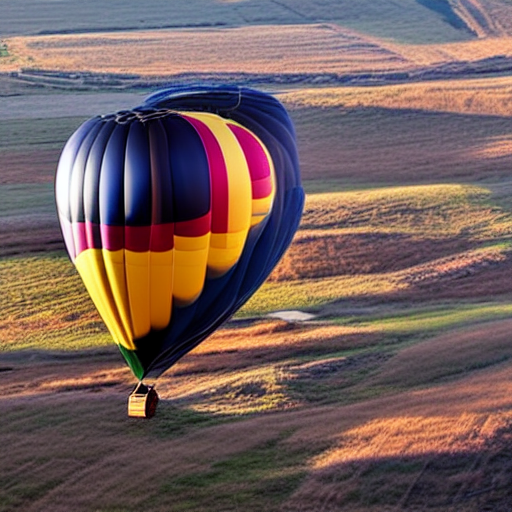}
    \caption{Ours ($\lambda=1$)}
  \end{subfigure}
  \begin{subfigure}[t]{.162\linewidth}
    \centering\includegraphics[width=\linewidth]{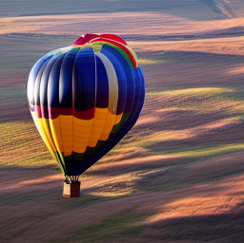}
    \caption{Ours}
  \end{subfigure}
\end{adjustbox}
\caption{ Visual results for scribble based editing.}
\label{fig:results:scribble}

\end{figure*}

\begin{figure*}[ht]
    \centering
    \begin{adjustbox}{minipage=\linewidth,scale=0.8}
    \begin{subfigure}[t]{.159\linewidth}
    \centering\includegraphics[width=\linewidth]{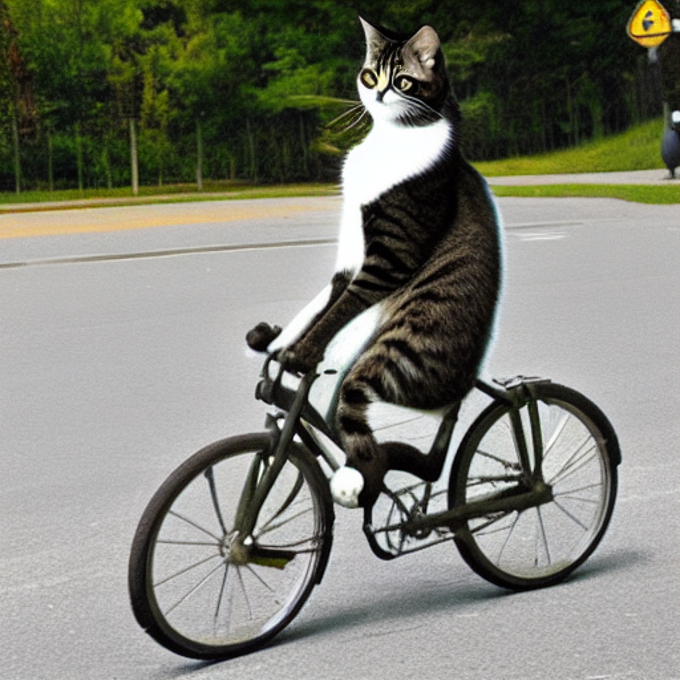}
    \end{subfigure}
  \begin{subfigure}[t]{.159\linewidth}
    \centering\includegraphics[width=\linewidth]{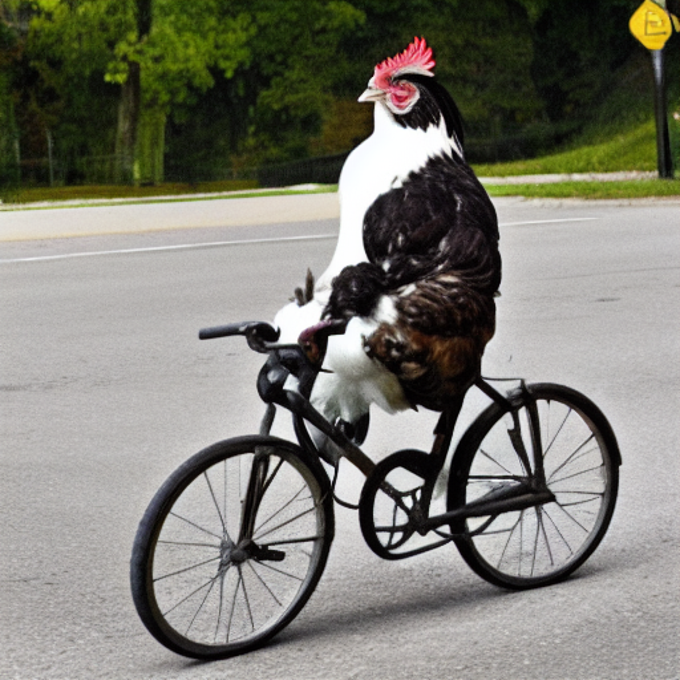}
  \end{subfigure}
  \begin{subfigure}[t]{.159\linewidth}
    \centering\includegraphics[width=\linewidth]{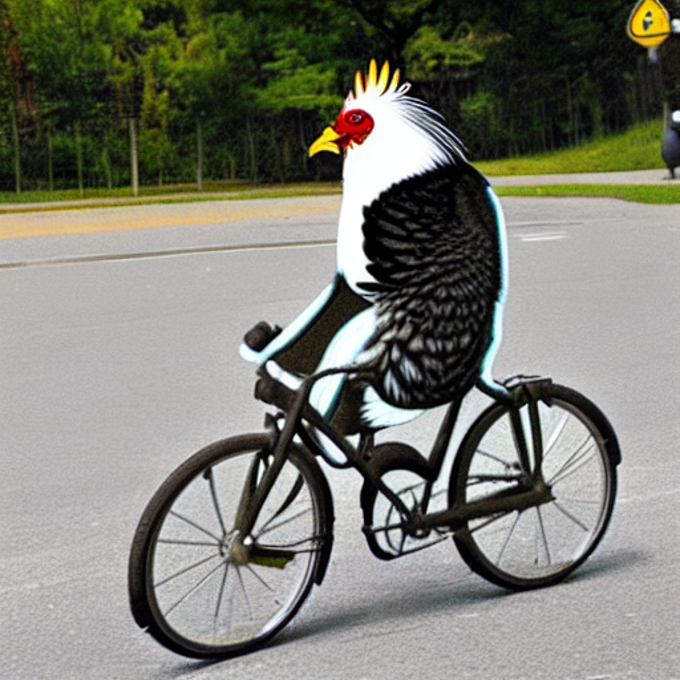}
  \end{subfigure}
    \begin{subfigure}[t]{.159\linewidth}
    \centering\includegraphics[width=\linewidth]{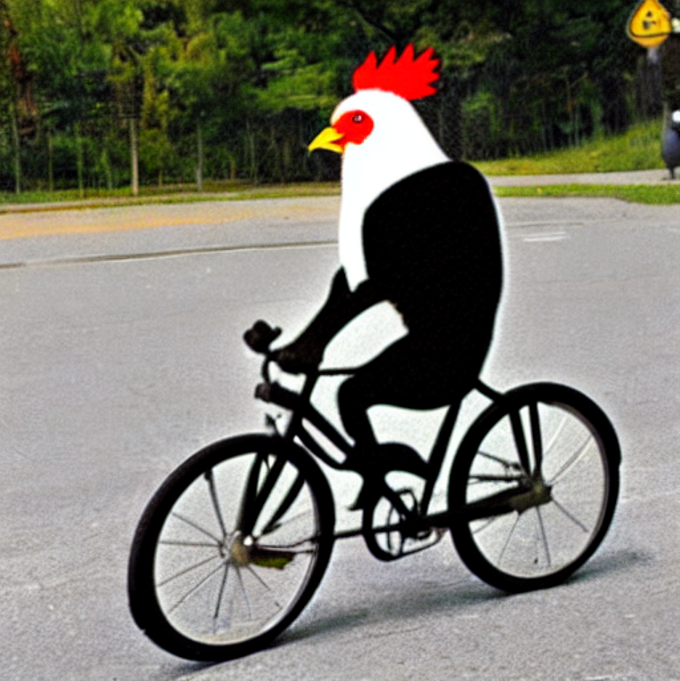}
  \end{subfigure}
  \begin{subfigure}[t]{.159\linewidth}
    \centering\includegraphics[width=\linewidth]{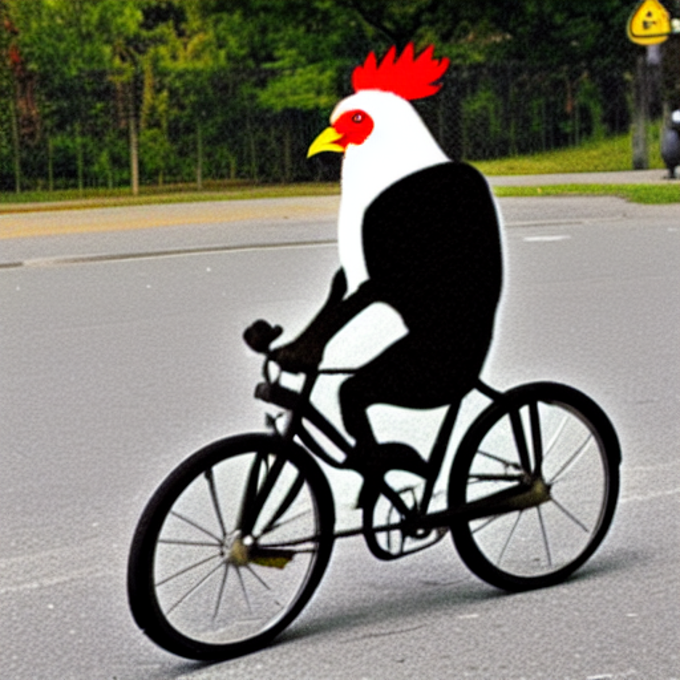}
  \end{subfigure}
  \begin{subfigure}[t]{.159\linewidth}
    \centering\includegraphics[width=\linewidth]{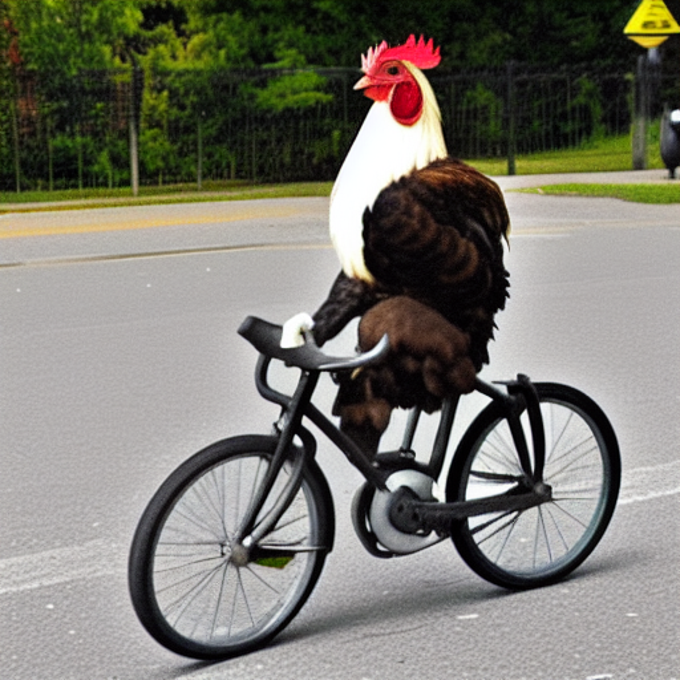}
  \end{subfigure}
  

\begin{subfigure}[t]{.159\linewidth}
    \centering\includegraphics[width=\linewidth]{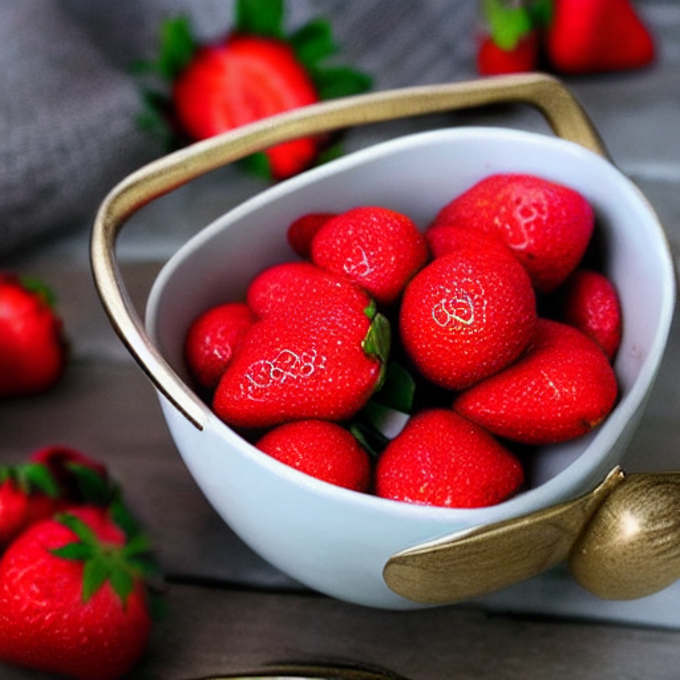}
    \end{subfigure}
  \begin{subfigure}[t]{.159\linewidth}
    \centering\includegraphics[width=\linewidth]{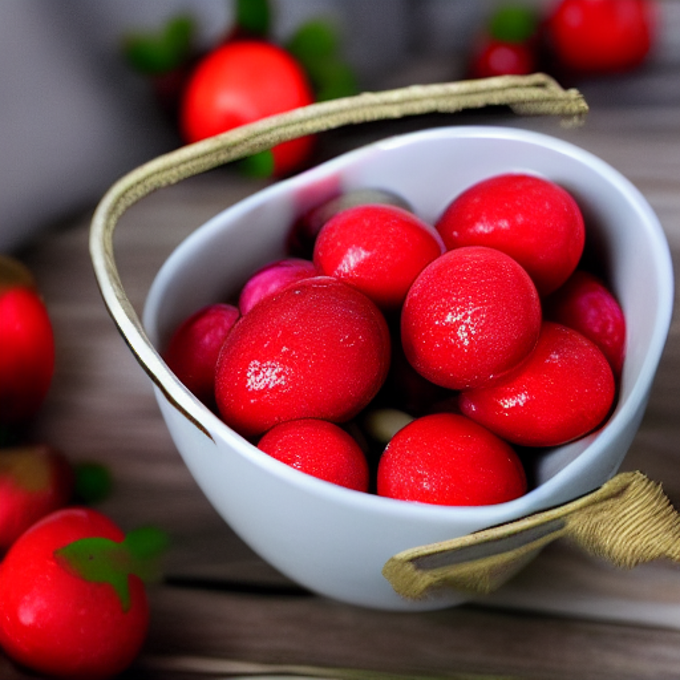}
  \end{subfigure}
  \begin{subfigure}[t]{.159\linewidth}
    \centering\includegraphics[width=\linewidth]{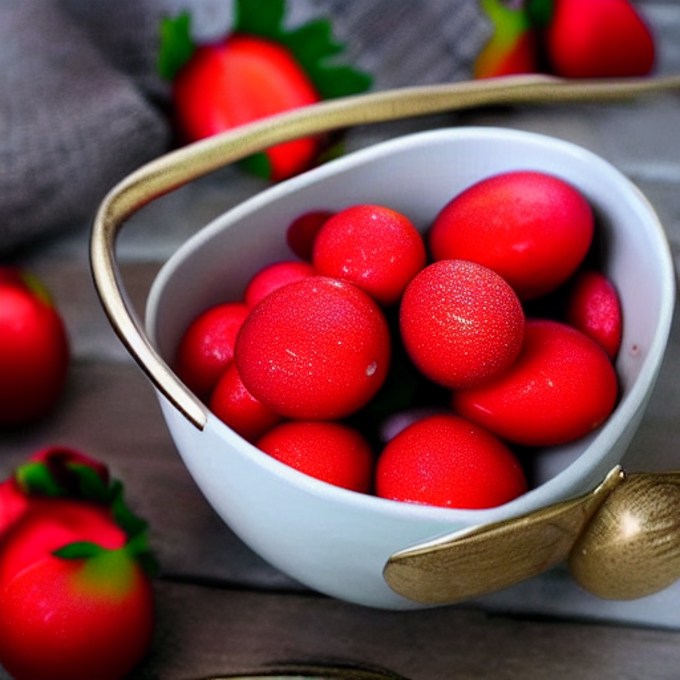}
  \end{subfigure}
    \begin{subfigure}[t]{.159\linewidth}
    \centering\includegraphics[width=\linewidth]{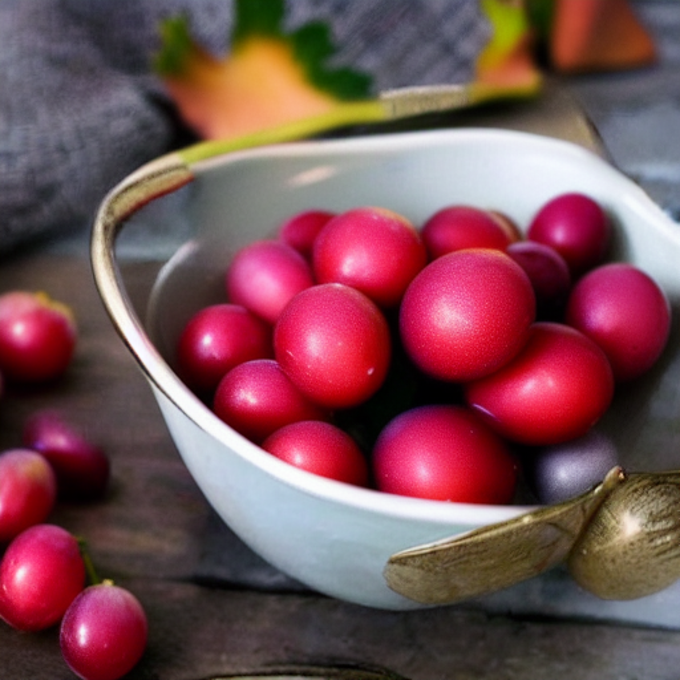}
  \end{subfigure}
  \begin{subfigure}[t]{.159\linewidth}
    \centering\includegraphics[width=\linewidth]{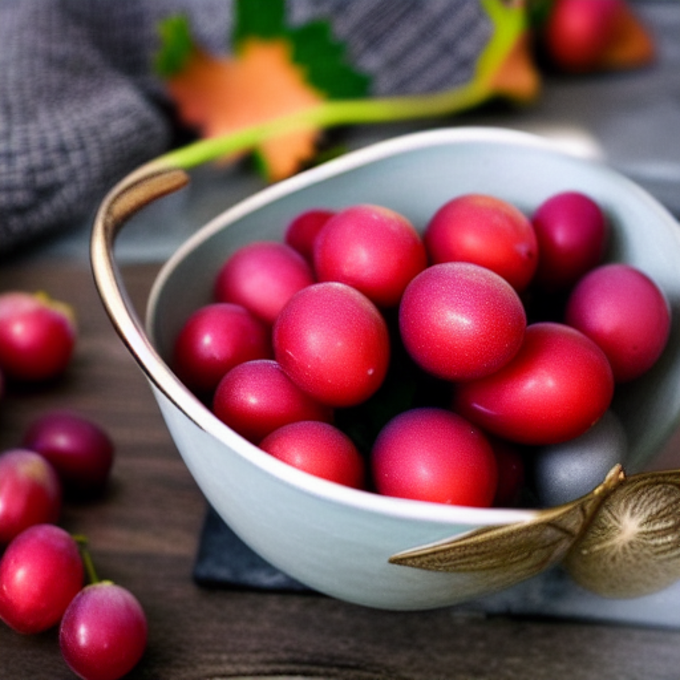}
  \end{subfigure}
  \begin{subfigure}[t]{.159\linewidth}
    \centering\includegraphics[width=\linewidth]{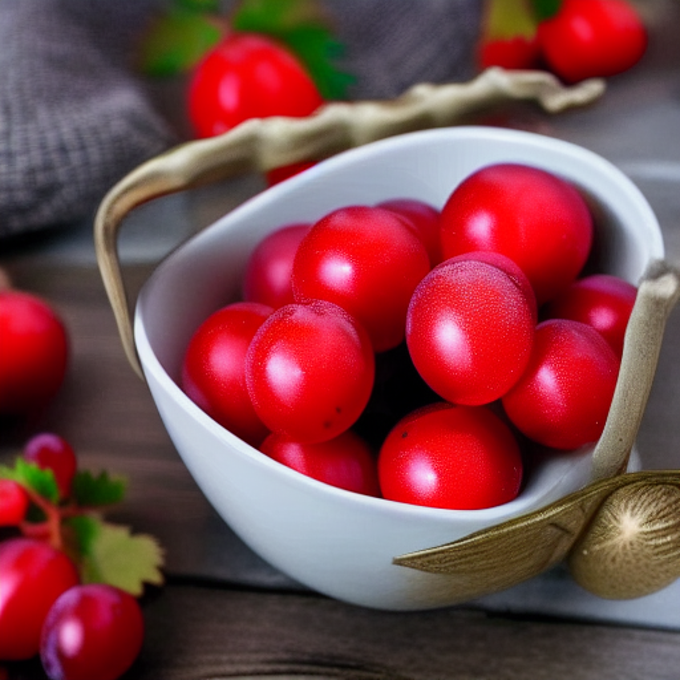}
  \end{subfigure}


\begin{subfigure}[t]{.159\linewidth}
    \centering\includegraphics[width=\linewidth]{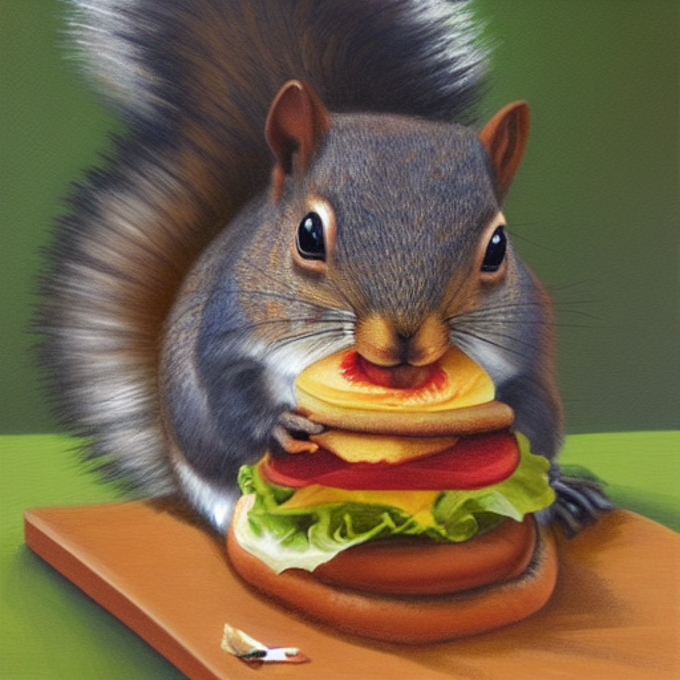}
    \end{subfigure}
  \begin{subfigure}[t]{.159\linewidth}
    \centering\includegraphics[width=\linewidth]{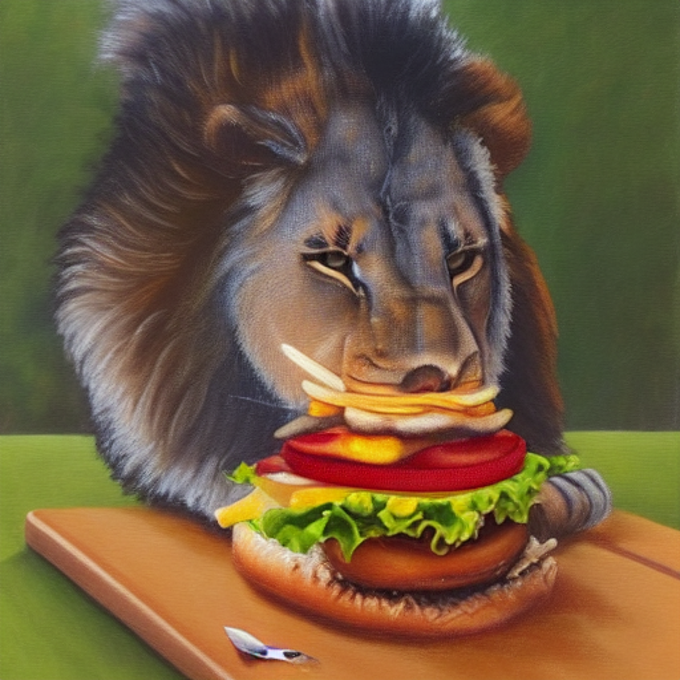}
  \end{subfigure}
  \begin{subfigure}[t]{.159\linewidth}
    \centering\includegraphics[width=\linewidth]{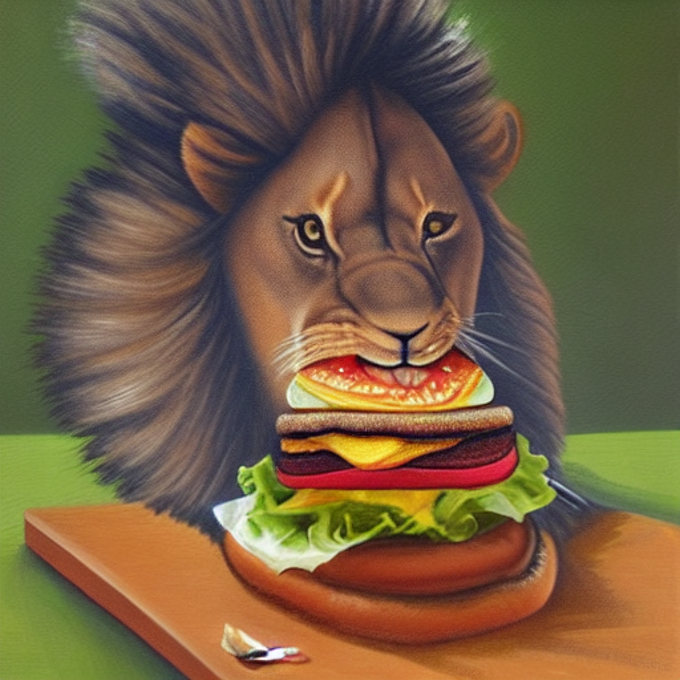}
  \end{subfigure}
    \begin{subfigure}[t]{.159\linewidth}
    \centering\includegraphics[width=\linewidth]{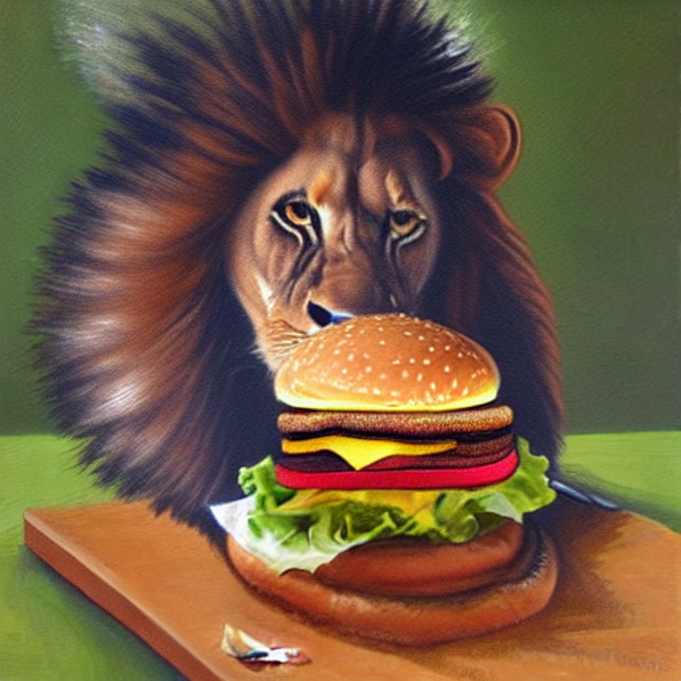}
  \end{subfigure}
  \begin{subfigure}[t]{.159\linewidth}
    \centering\includegraphics[width=\linewidth]{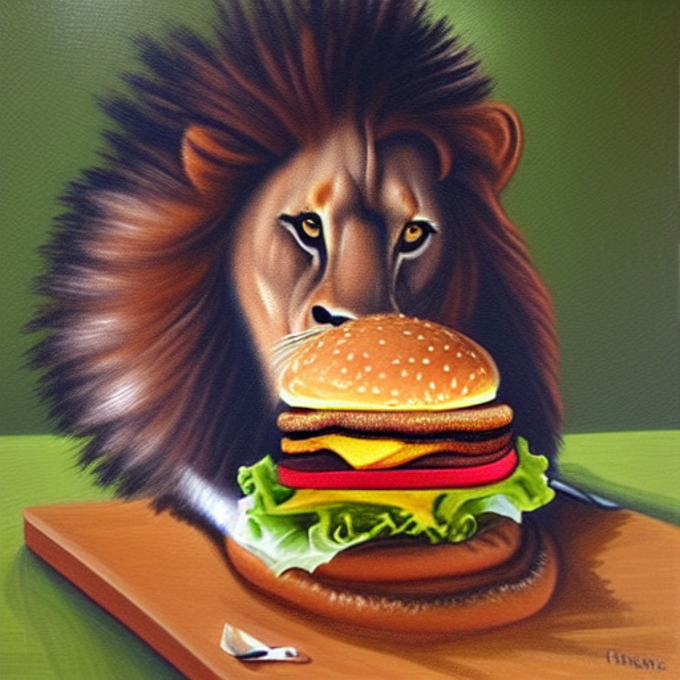}
  \end{subfigure}
  \begin{subfigure}[t]{.159\linewidth}
    \centering\includegraphics[width=\linewidth]{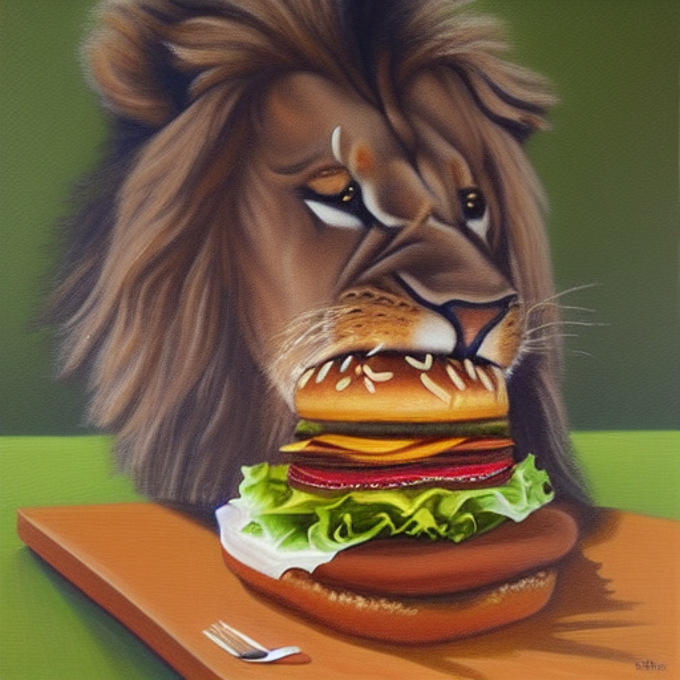}
  \end{subfigure}


  \begin{subfigure}[t]{.159\linewidth}
    \centering\includegraphics[width=\linewidth]{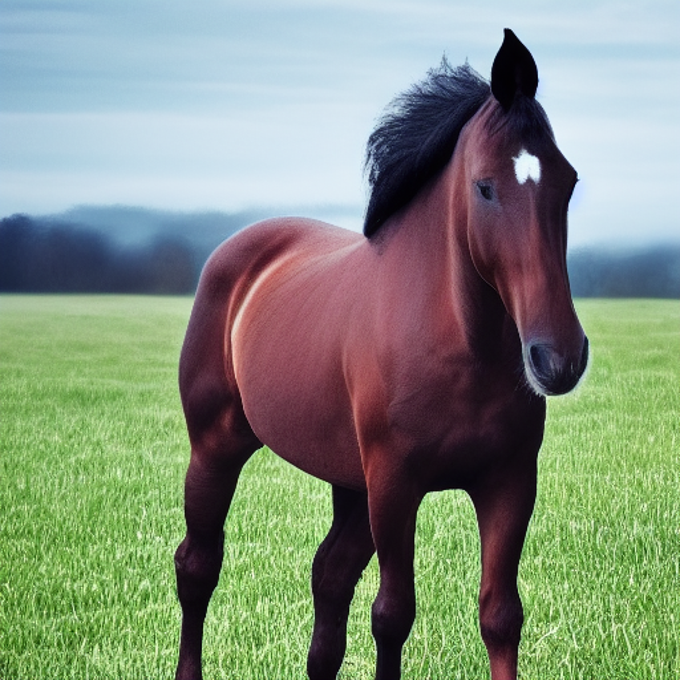}
    \caption{Original image}
    \end{subfigure}
  \begin{subfigure}[t]{.159\linewidth}
    \centering\includegraphics[width=\linewidth]{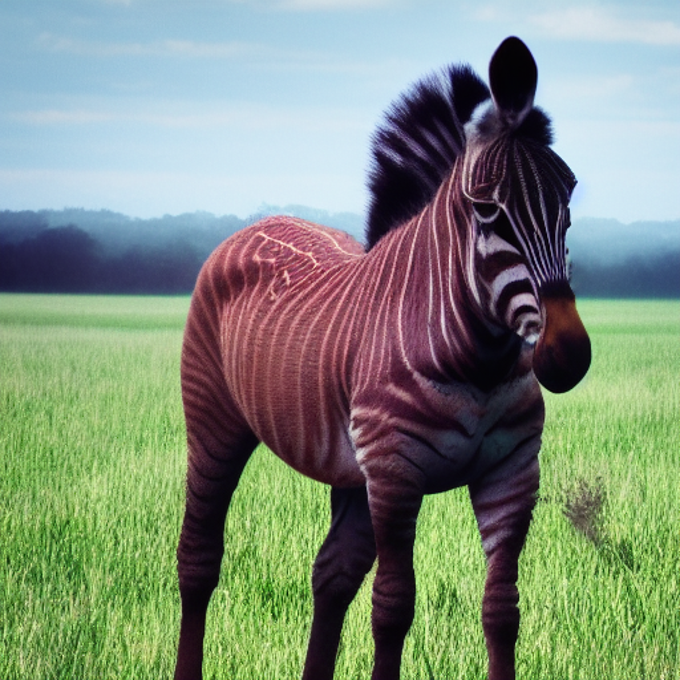}
    \caption{Prompt to prompt}
  \end{subfigure}
  \begin{subfigure}[t]{.159\linewidth}
    \centering\includegraphics[width=\linewidth]{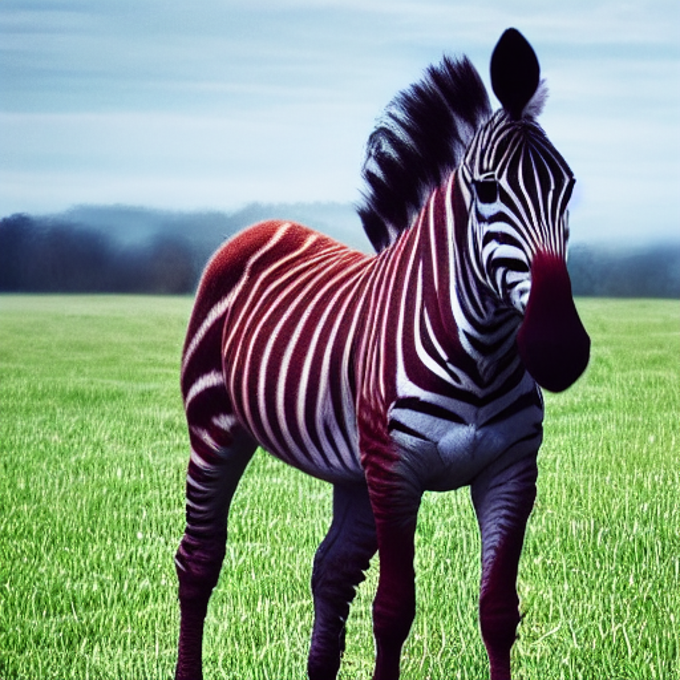}
    \caption{DiffEdit $ER=0.7$}
  \end{subfigure}
    \begin{subfigure}[t]{.159\linewidth}
    \centering\includegraphics[width=\linewidth]{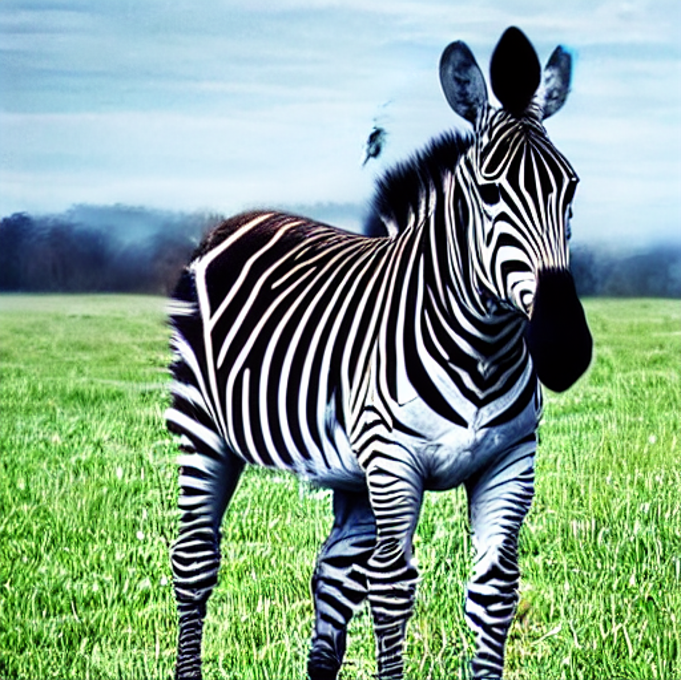}
    \caption{DiffEdit $ER=1.0$}
  \end{subfigure}
  \begin{subfigure}[t]{.159\linewidth}
    \centering\includegraphics[width=\linewidth]{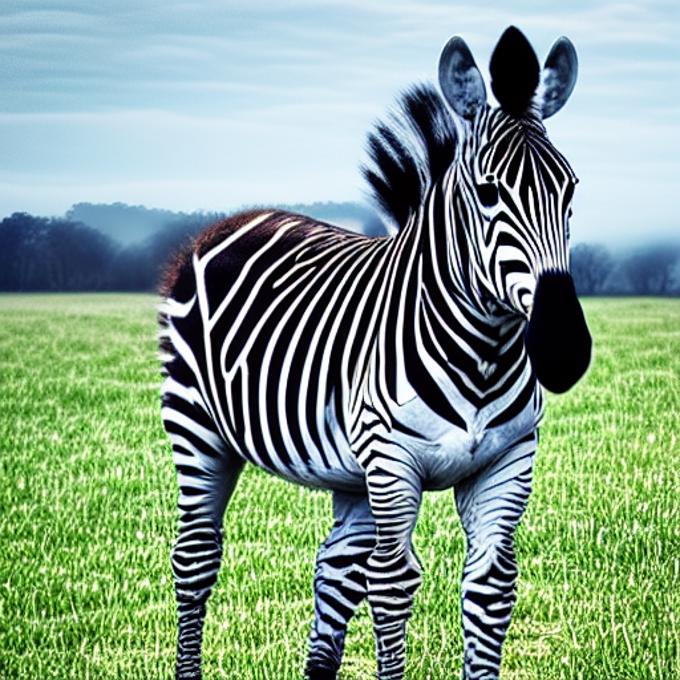}
    \caption{Ours $\lambda=1$}
  \end{subfigure}
  \begin{subfigure}[t]{.159\linewidth}
    \centering\includegraphics[width=\linewidth]{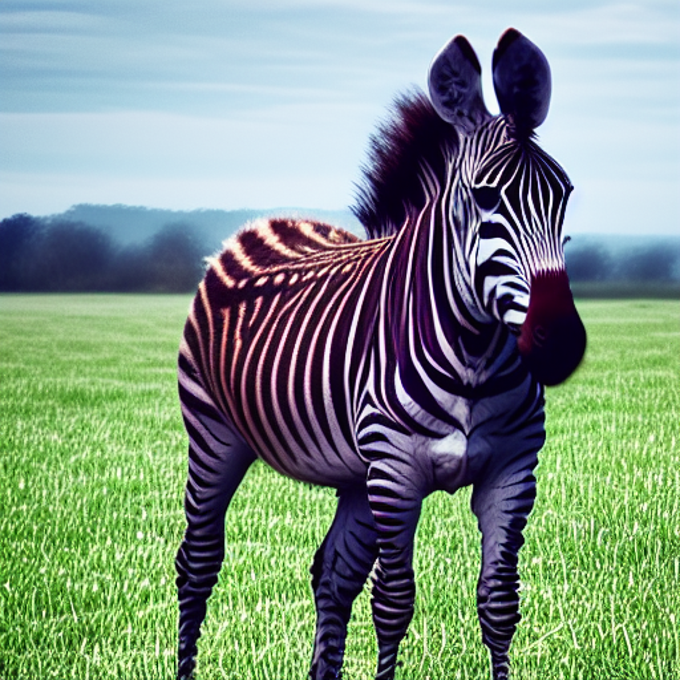}
    \caption{Ours}
  \end{subfigure}
\end{adjustbox}
\caption{ Visual results for text-conditioned image editing. From top to bottom, input/edit prompts are: a cat/chicken riding a bicycle, a bowl full of strawberries/grapes, a squirrel/lion eating a burger, a horse/zebra in a field. }
\label{fig:results:text}
\end{figure*}

\noindent \textbf{Implementation.} 
In our experiments we use ControlNet~\citep{zhang2023adding} to control a Stable Diffusion model v1.5 ~\citep{rombach2022high} with layout edit conditions. 
We use 25 sampling steps, and an encoding ratio of $1$ ($t_E=25$) unless specified otherwise. We highlight that this experimental setting is not optimised for image generation quality, but rather focuses on layout constrained editing ability. Balancing parameter $\lambda$ is set to $0.6$ unless specified otherwise. For mask generation, we follow the parameters of \citep{couairon2022diffedit}, but introduce Gaussian smoothing of the mask, and use a threshold of $0.1$. 
When using a mixture of guidance and preservation loss, we optimise features for the first $t_u=15$ timesteps for $k=1$ gradient step, for both guidance generation and final editing steps. 
We use the ADAM optimiser with a learning rate $\gamma=0.1$.  

\noindent\textbf{Layout-Conditioned Image Editing.}
As pre-existing image editing methods have not been developed for layout conditioning, we use ControlNet as our simplest baseline. Additionally, we adapt the Diffedit method~\citep{couairon2022diffedit} to leverage layout constraints through an integration of controlNet. In this setting, we set the encoding ratio for DiffEdit to 1.0, as we have observed better results, and use the binary mask estimated using our updated approach for layout conditions. 
Figs.~\ref{fig:results:pose} and~\ref{fig:results:scribble} show visual results for pose and scribble conditions, where we edit an image originally generated using a ControlNet pose or scribble constraint. We focus on synthetic images to allow comparisons with ControlNet, and provide experiments on real images in supp.~materials.
Fig.~\ref{fig:results:pose} shows visual results under different pose conditions. We can see that our approach is the only one to consistently achieve the correct pose modification while preserving image content. As expected, ControlNet successfully generates images with the correct pose, but fails to preserve image content. In contrast, while DiffEdit preserves image content thanks to its masking process, it often struggles to achieve the correct pose change, especially for more difficult queries. We highlight in particular how DiffEdit often suffers from local artefacts, often due to the strict mask blending constraint. Lastly, we point out the strong similarities between DiffEdit's performance and our full preservation special configuration ($\lambda=1$), with ours having fewer image artifacts (see the altered image quality for the dancer and bear images). This shows how our preservation loss provides a robust alternative to DiffEdit's mask merging, and the benefits of using our guidance image. 
Our scribble-based editing results, shown in Fig.~\ref{fig:results:scribble}, show a similar behaviour.

\noindent\textbf{Text-Driven Image Editing.}
Existing quantitative evaluation protocols and metrics, in addition to pre-existing methods, only consider text-guided editing. 
Towards direct comparison with recent work, we therefore exclusively considering text conditioning here. 
We follow the protocol of DiffEdit~\citep{couairon2022diffedit} on the ImageNet~\citep{deng2009imagenet} dataset. The task is to edit an image belonging to one class, such that it depicts an object belonging to another class. This is indicated by an edit text prompt comprising the name of the new class. 
Given the nature of the ImageNet dataset, edits require modifying the main object in the scene.

We compare our results to state of the art methods with default parameters: Prompt-to-prompt \cite{hertz2022prompt} and DiffEdit \cite{couairon2022diffedit} (encoding ratio=0.7).  We evaluate performance from four angles: 1) image preservation by computing the L1 distance between original and edited images, 2) image quality by computing the CSFID score~\citep{couairon2022flexit}, which is a class-conditional FID metric~\citep{heusel2017gans}, measuring both image realism and consistency with respect to the transformation prompt, 3) quality of the semantic modification, by measuring classification accuracy of the edited image using (a) the new category as ground truth, and (b) the original category as ground truth, and finally we measure 4) relative image quality and prompt faithfulness using the Pickscore \citep{kirstain2023pick}. As shown in Table~\ref{tab:results}, our approach outperforms competing works for all metrics, except the L1 consistency metric. This can be expected as edited regions often are the only subject of the image, such that successful edits achieve more substantial image modifications. Visual results of this experiment are additionally available in supp.~materials.

Additionally, Fig.~\ref{fig:results:text} shows visual results using text-based editing. Here, we edit images generated with a Stable Diffusion v1.5 model~\citep{rombach2022high}. We provide DiffEdit's results for an encoding ratio of 0.7 and 1.0, highlighting the similarity between our method when only using the preservation loss ($\lambda=1$). As evidenced with the zebra image, where DiffEdit shows local artifacts, we are less sensitive to subpar editing masks. We note how our method can achieve successful conversions, such as the bowl of grapes, where pre-existing works fail.  

\begin{figure*}
    \centering
    \includegraphics[width=0.83\linewidth]{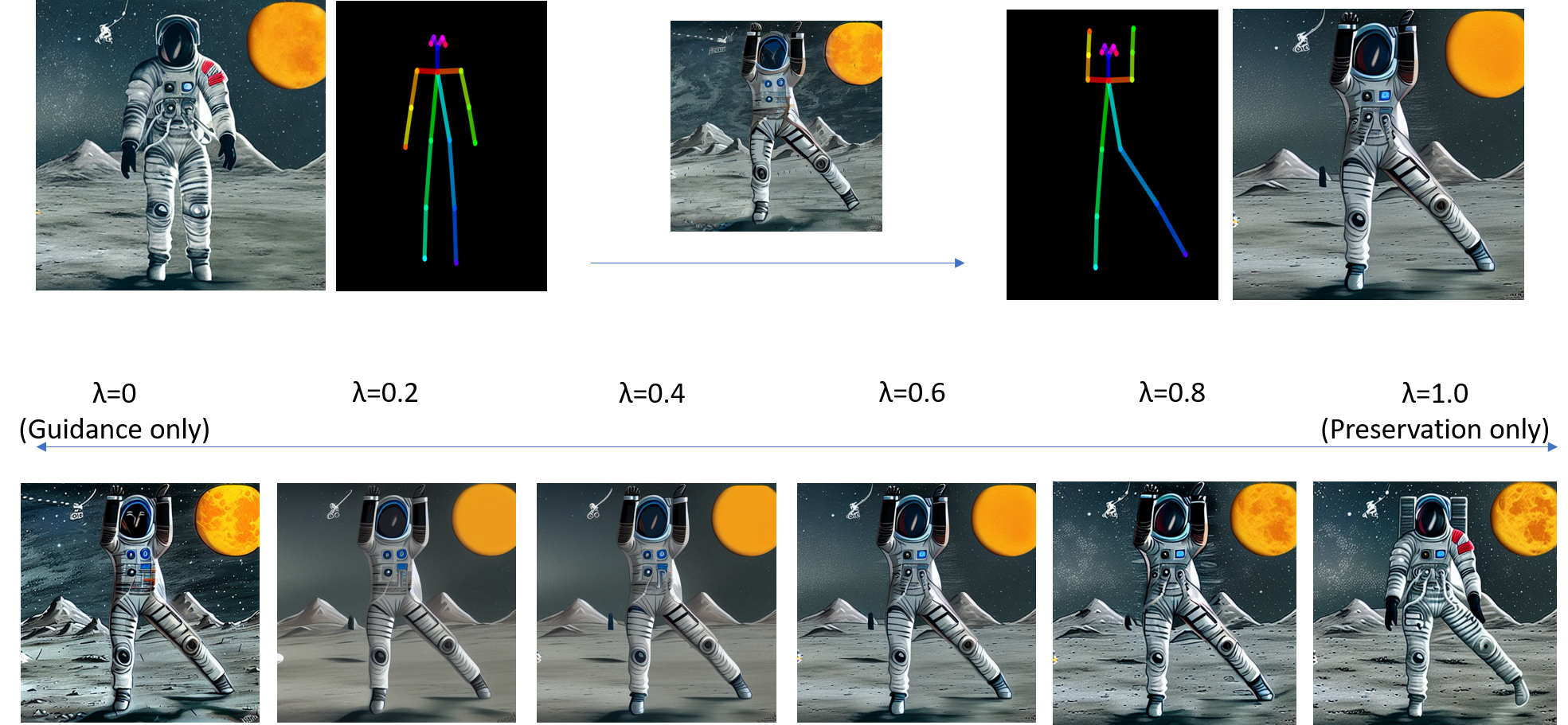}
    \caption{Influence of balancing parameter $\lambda$. Top left: input image and input pose, middle: guidance image, top right: output image and target pose.}
    \label{fig:lambda}
\end{figure*}

\noindent\textbf{Parameter Study.}
In Fig.~\ref{fig:lambda}, we illustrate the impact of adjusting balancing parameter $\lambda$ which controls the relative influence of preservation and guidance losses. We can see that lower values of $\lambda$ focus on foreground and accurate positioning. Larger $\lambda$ values increase background details at the cost of additional artifacts when $\lambda$ is too high. Extreme values yield an output very close to the guidance image ($\lambda=0$), or similar to the DiffEdit performance ($\lambda=1$); high background fidelity, but poorer pose edit quality. This behaviour highlights how, for simpler edit instructions (e.g. the bear or dancer in Fig. \ref{fig:results:pose}, where DiffEdit successfully achieves the pose modification), one can opt to discard the guidance image component and edit the image with a lone preservation loss. Extended results are available in the supp.~materials.

\section{Conclusion and Limitations}\label{sec:con}
We propose a novel method for diffusion model-driven image editing that is uniquely designed to leverage layout-based edit conditions, in addition to text instructions. By disentangling preservation and modification components and further accounting for multiple edit modalities, we clearly demonstrate that optimisation-based image editing provides a promising solution for complex image modification tasks where consistency is important. The work is currently limited to the editing of single 2D image instances however temporal extensions, robustly accounting for video sequences, provides a promising future direction. 
Future work will also explore alternative mask and guidance generation methods towards further improving edit qualities. 

\newpage
{
    \small
    \bibliographystyle{ieeenat_fullname}
    \bibliography{main}
}

\clearpage
\setcounter{page}{1}
\maketitlesupplementary

\setcounter{equation}{0}
\setcounter{figure}{0}
\setcounter{table}{0}
\setcounter{page}{1}
\renewcommand{\theequation}{S\arabic{equation}}
\renewcommand{\thefigure}{S\arabic{figure}}
\renewcommand{\thetable}{S\arabic{table}}

\section{Inference-time optimisation: detailed illustration}

\begin{figure*}[h]
    \centering
    \includegraphics[width=\linewidth]{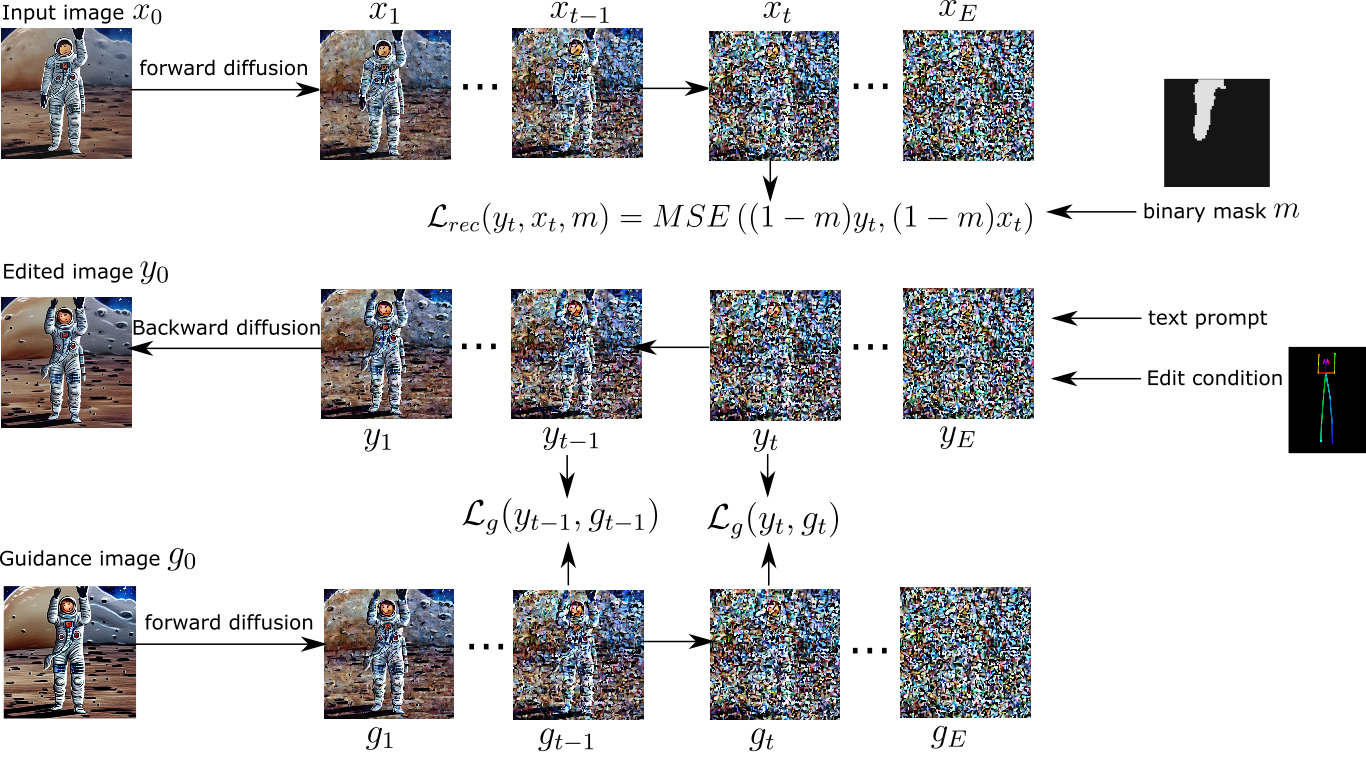}
    \caption{Overview of our inference time optimisation method with preservation and guidance losses.}
    \label{fig:supp:ITO}
\end{figure*}

In~\cref{fig:supp:ITO} we provide a detailed visualisation of our ITO process with guidance and preservation losses, to aid user comprehension. 

\section{Parameter study: update steps}

\begin{figure*}
\centering
 \begin{subfigure}[t]{.12\linewidth}
    \centering\includegraphics[width=\linewidth]{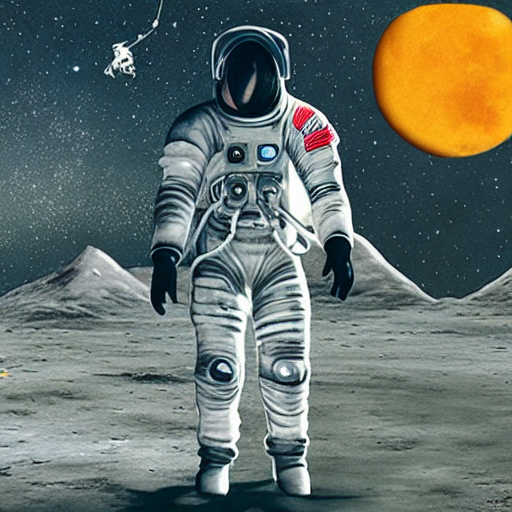}
    \caption{Original image}
    \end{subfigure}
 \begin{subfigure}[t]{.72\linewidth}
    \centering\includegraphics[width=\linewidth]{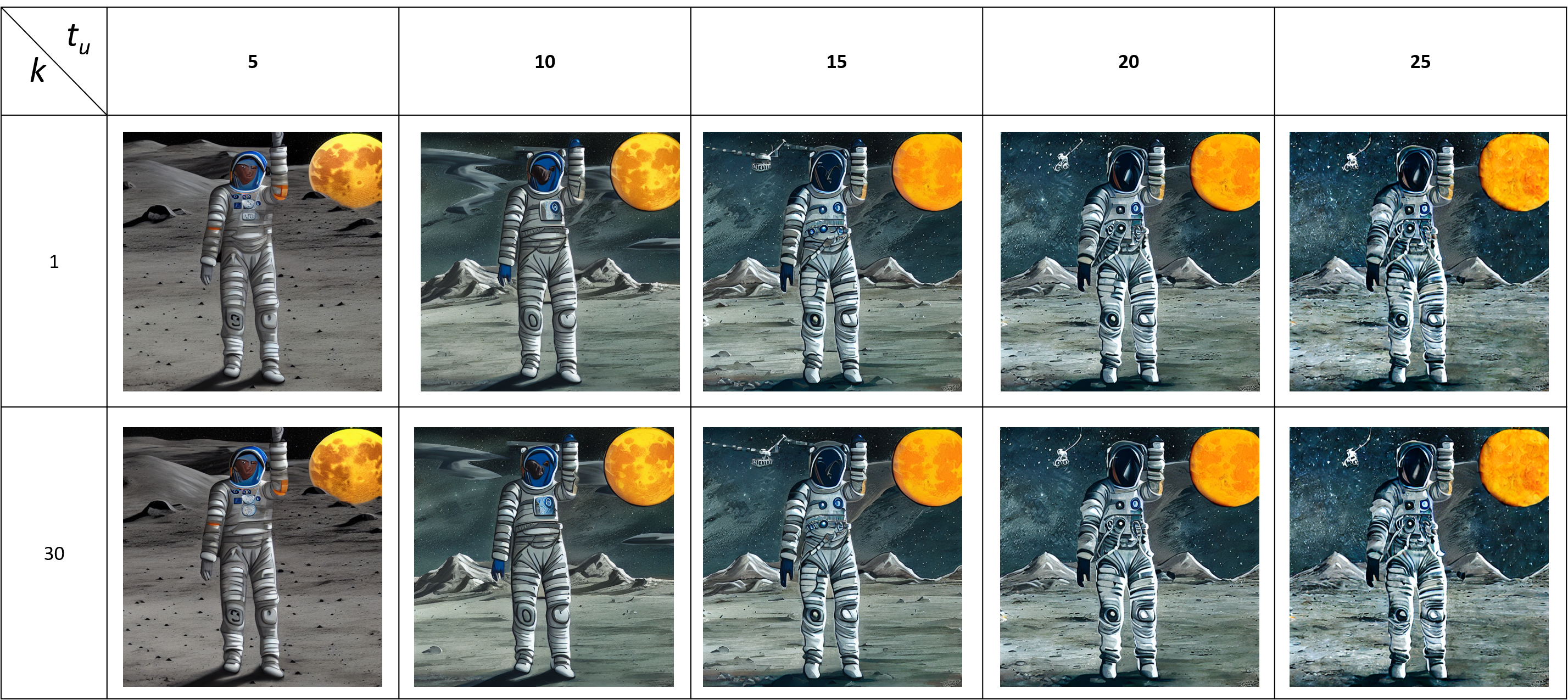}
    \caption{Guidance image}
    \label{fig:supp:param:ast_g}
    \end{subfigure}
      \begin{subfigure}[t]{.72\linewidth}
    \centering\includegraphics[width=\linewidth]{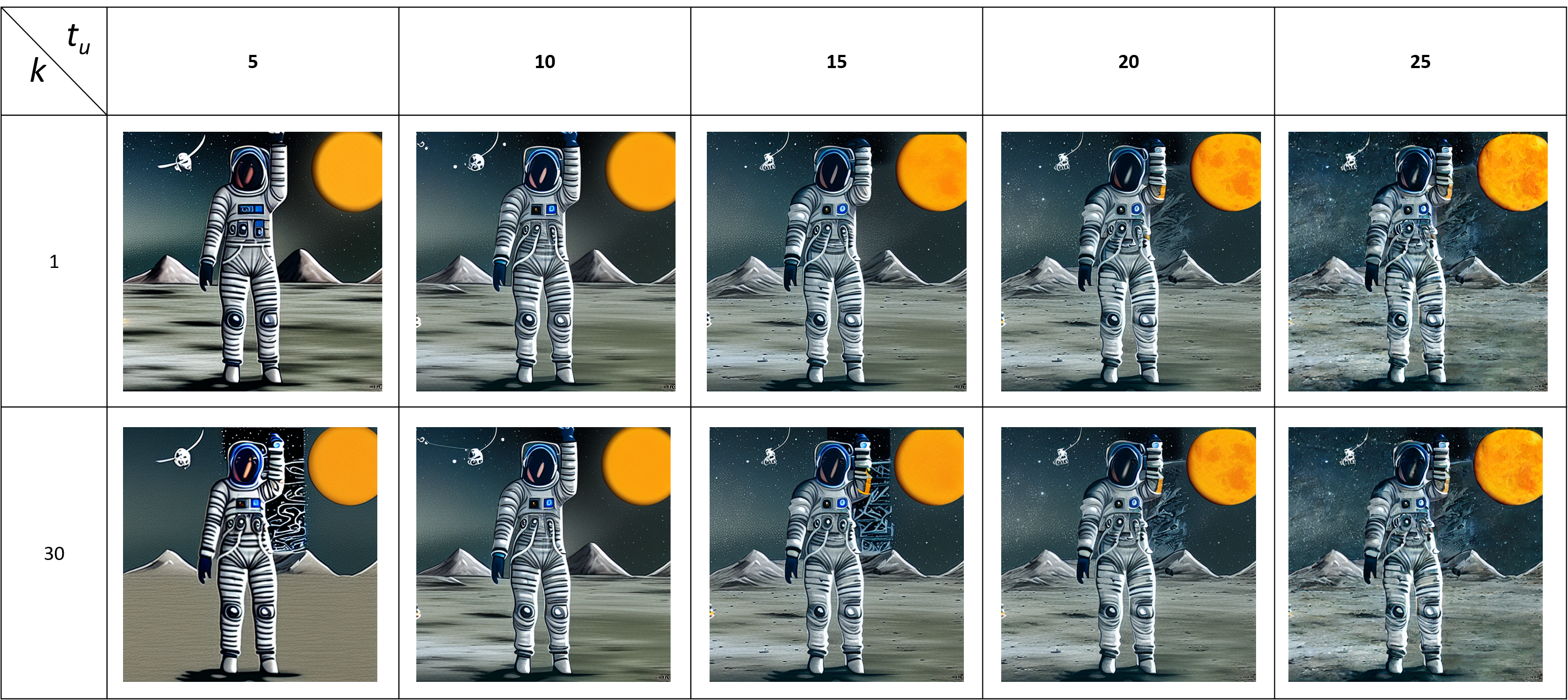}
    \caption{Final edit}
    \label{fig:supp:param:ast_fin}
    \end{subfigure} \\
     \begin{subfigure}[t]{.12\linewidth}
    \centering\includegraphics[width=\linewidth]{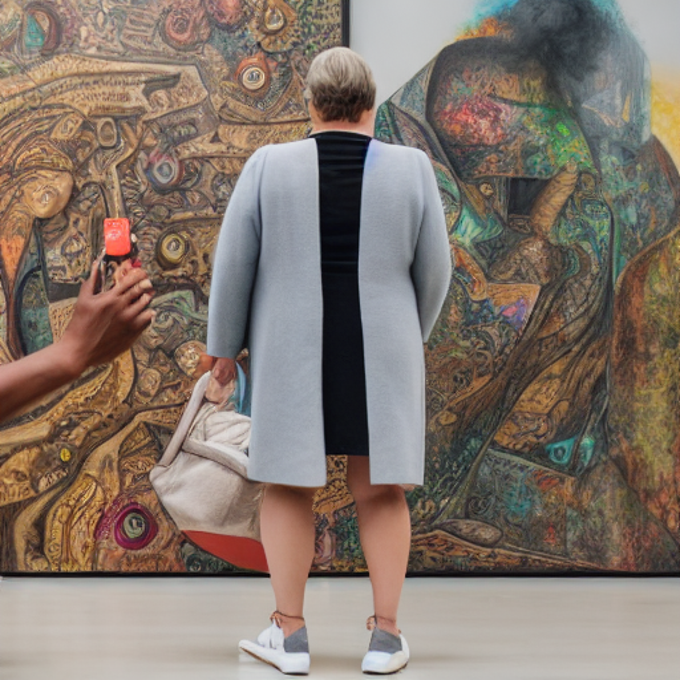}
    \caption{Original image}
    \end{subfigure}
    \begin{subfigure}[t]{.72\linewidth}
    \centering\includegraphics[width=\linewidth]{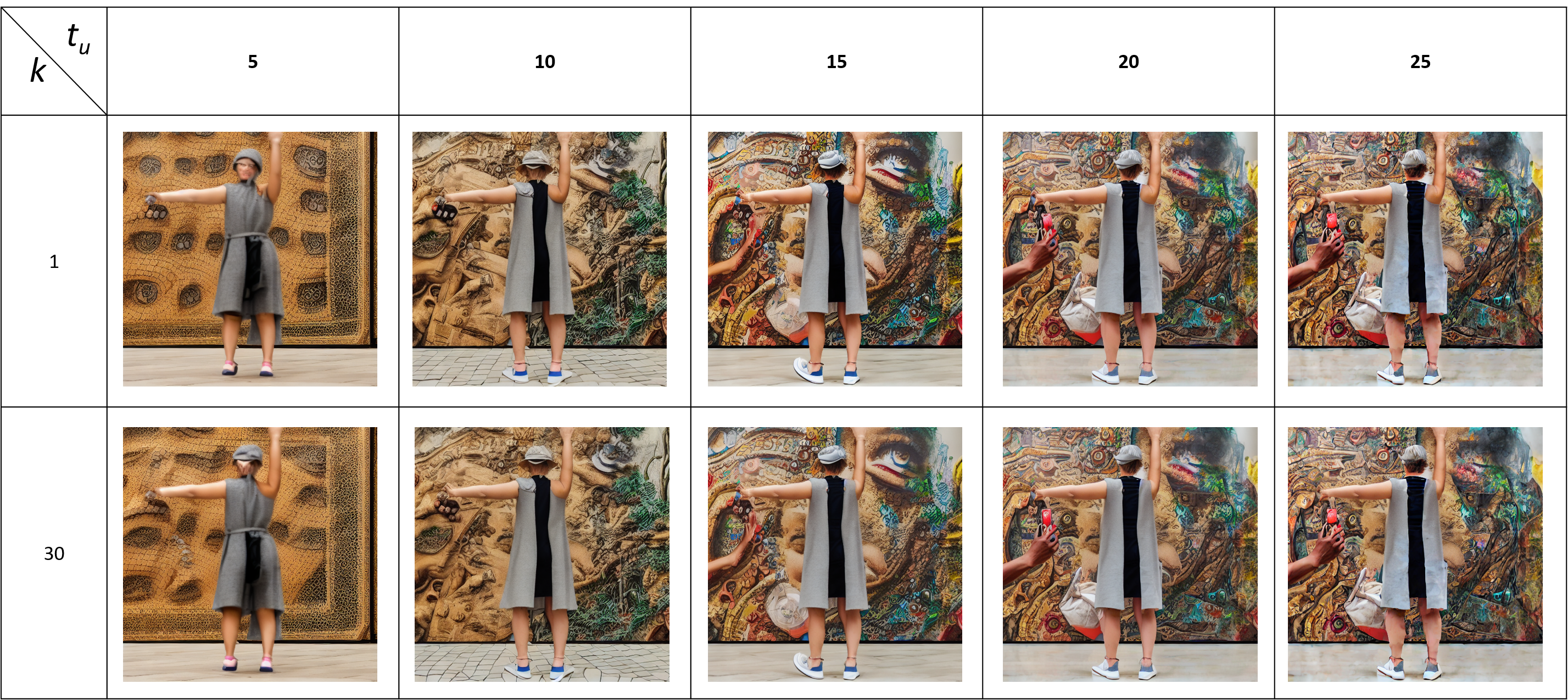}
    \caption{Guidance image}
    \label{fig:supp:param:tourist_g}
  \end{subfigure}
  \begin{subfigure}[t]{.72\linewidth}
    \centering\includegraphics[width=\linewidth]{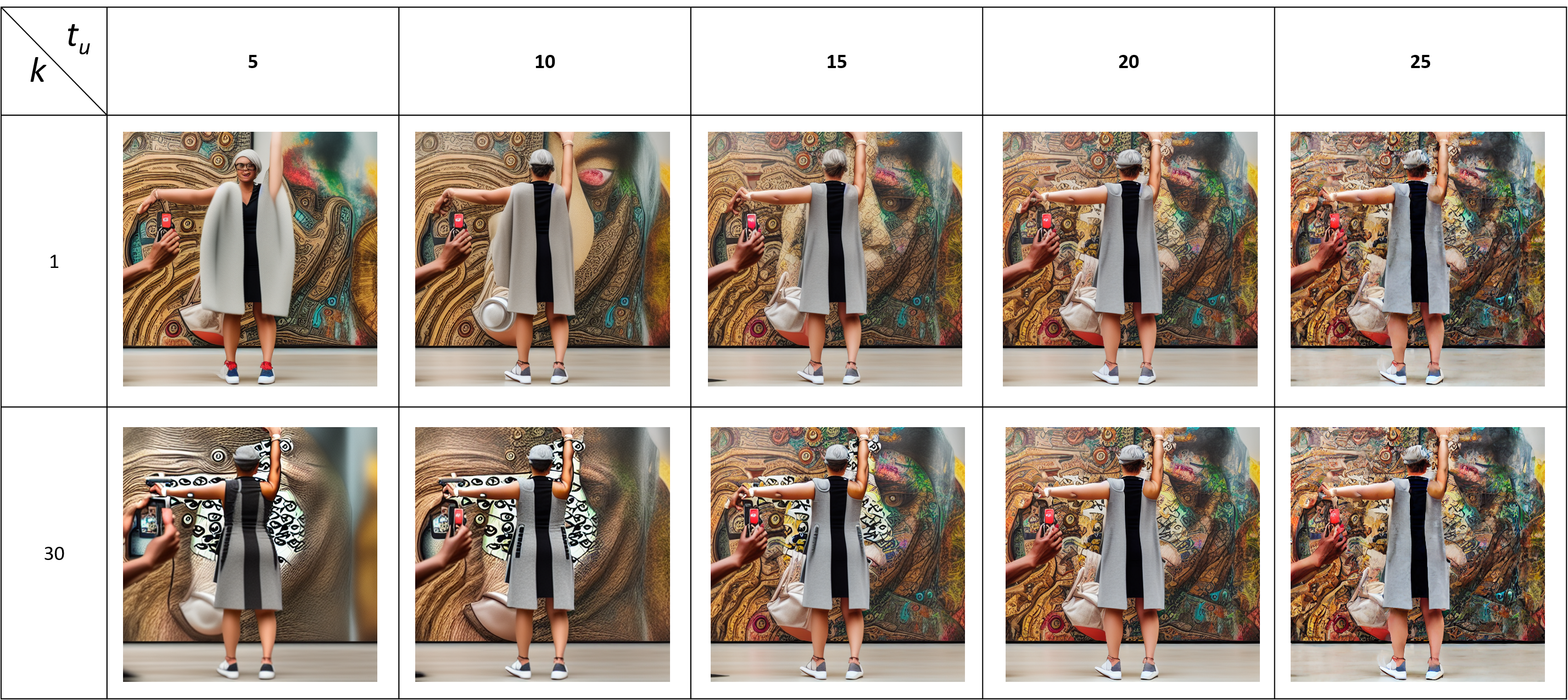}
    \caption{Final edit}
    \label{fig:supp:param:tourist_fin}
  \end{subfigure}
  \caption{ Impact of parameters $t_u$ and $k$ on guidance image and final edit generation for two example images. }
\label{fig:supp:param}
\end{figure*}

In \cref{fig:supp:param}, we visualise the influence of parameters $t_u$ (number of timesteps with optimisation) and $k$ (number of optimisation steps per timestep) over both guidance image generation and final edit. We provide visual results for two image edits using pose conditioning for $t_u \in \{5,10,15,20,25\}$ and $k=\{1,30\}$, choosing images with complex backgrounds to highlight the impact of these parameters more clearly. We can see that $t_u$ has a larger impact over the generated images than $k$, especially with regards to the guidance image. Compared to the input image, we can see that the best quality guidance images (in terms of content preservation, achieved modification, and introduced artefacts) is obtained for $t_u=20, k=30$, while $t_u=15,k=1$ yields a better final edit. For guidance images, better content preservation at $t_u=20$ is notably evidenced by the woman's hair in \cref{fig:supp:param:tourist_g}, and the symbol on the top left of the astronaut images in \cref{fig:supp:param:ast_g}. Increasing $k$ has a minor impact, but yields slightly improved results (see the sleeve on the woman's right shoulder in \cref{fig:supp:param:tourist_g}). For final edits (Figures \ref{fig:supp:param:ast_fin} and \ref{fig:supp:param:tourist_fin}), $t_u=15, k=1$ yields noticeably better results, with reduced artefacts (see astronaut image, both images for $t_u=25$ and all settings where $k=30$). In particular, we can see that noticeable artefacts are introduced within the masked area for lower values of $t_u$ and $k>1$.

For the sake of consistency and speed, we use $t_u=15,k=1$ in all settings in our experiments, but note here results could be further improved by adjusting guidance image parameters to $t_u=20, k=30$. We further highlight that the optimal value of $t_u$ is dependent on the number of sampling steps used to generate the image (here we use $25$ steps, as discussed in Section 4 of the main paper).  

\section{Visualisation of intermediate edit steps}

\begin{figure*}
\centering
 \begin{subfigure}[t]{.169\linewidth}
    \centering\includegraphics[width=\linewidth]{imgs/results/astronaut_pose0.png}
    \end{subfigure}
      \begin{subfigure}[t]{.123\linewidth}
    \centering\includegraphics[width=\linewidth]{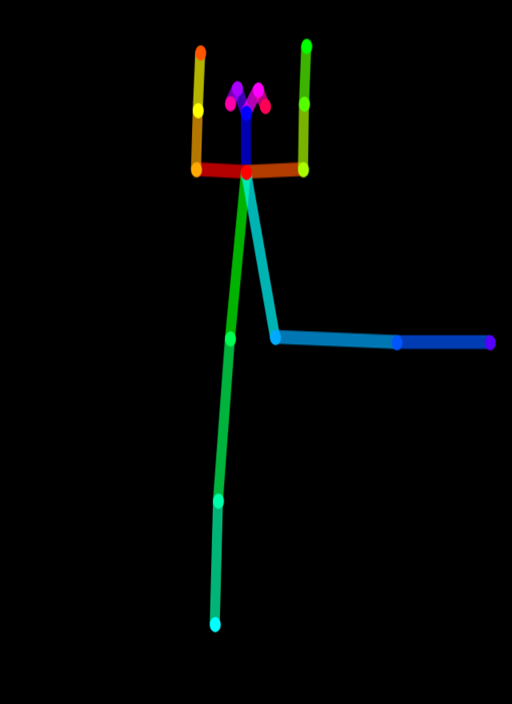}
    \end{subfigure}
  \begin{subfigure}[t]{.169\linewidth}
    \centering\includegraphics[width=\linewidth]{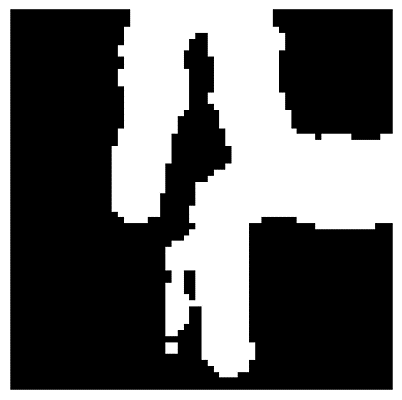}
  \end{subfigure}
  \begin{subfigure}[t]{.169\linewidth}
    \centering\includegraphics[width=\linewidth]{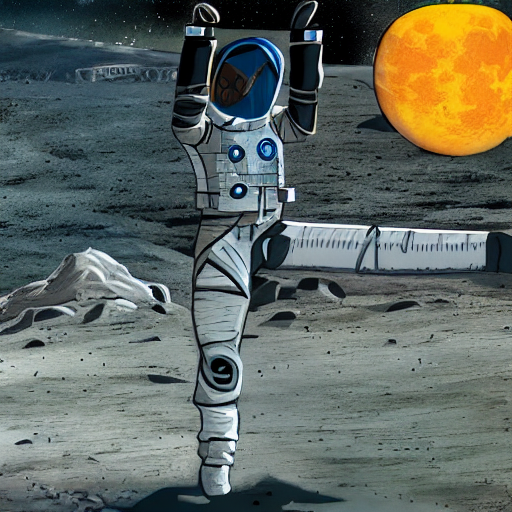}
  \end{subfigure}
  \begin{subfigure}[t]{.169\linewidth}
    \centering\includegraphics[width=\linewidth]{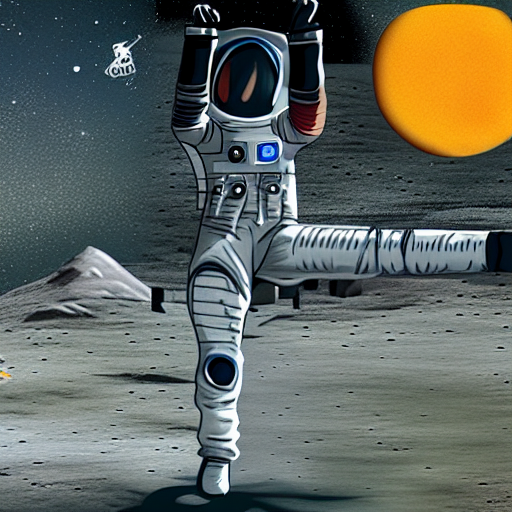}
  \end{subfigure}


 \begin{subfigure}[t]{.169\linewidth}
    \centering\includegraphics[width=\linewidth]{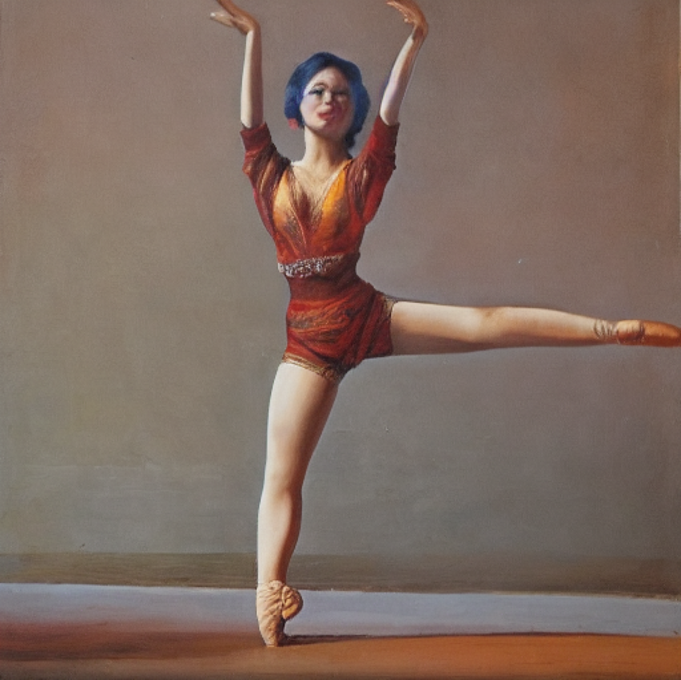}
    \end{subfigure}
      \begin{subfigure}[t]{.123\linewidth}
    \centering\includegraphics[width=\linewidth]{imgs/results/pose2.png}
    \end{subfigure}
  \begin{subfigure}[t]{.169\linewidth}
    \centering\includegraphics[width=\linewidth]{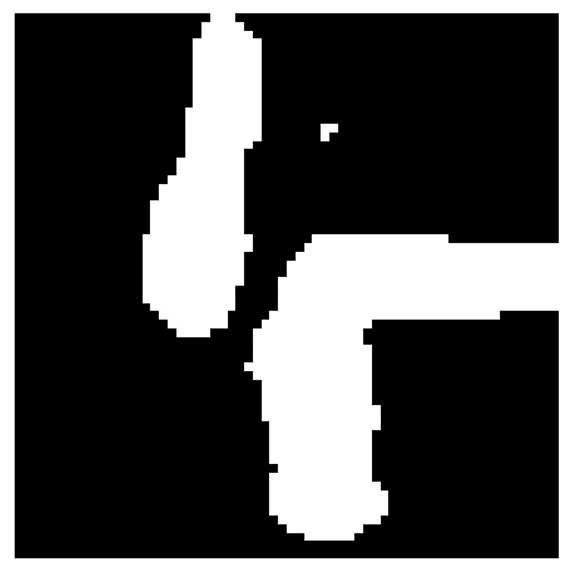}
  \end{subfigure}
  \begin{subfigure}[t]{.169\linewidth}
    \centering\includegraphics[width=\linewidth]{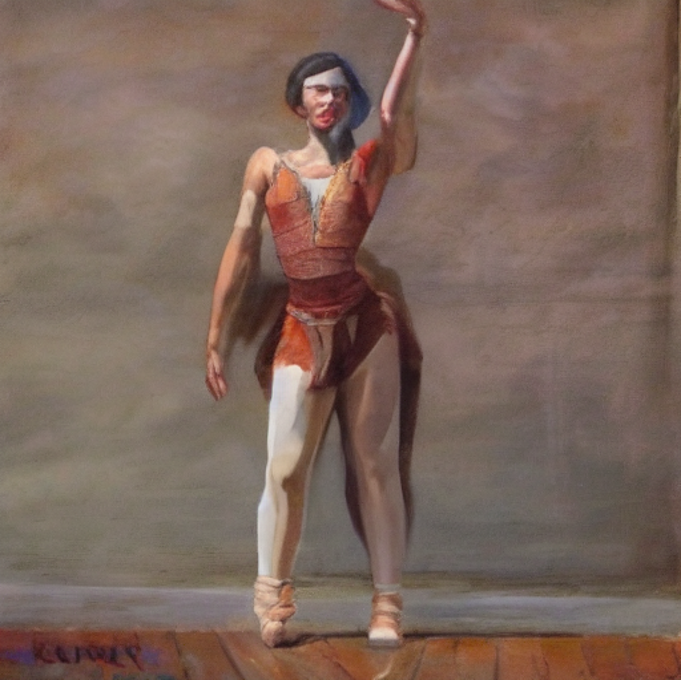}
  \end{subfigure}
  \begin{subfigure}[t]{.169\linewidth}
    \centering\includegraphics[width=\linewidth]{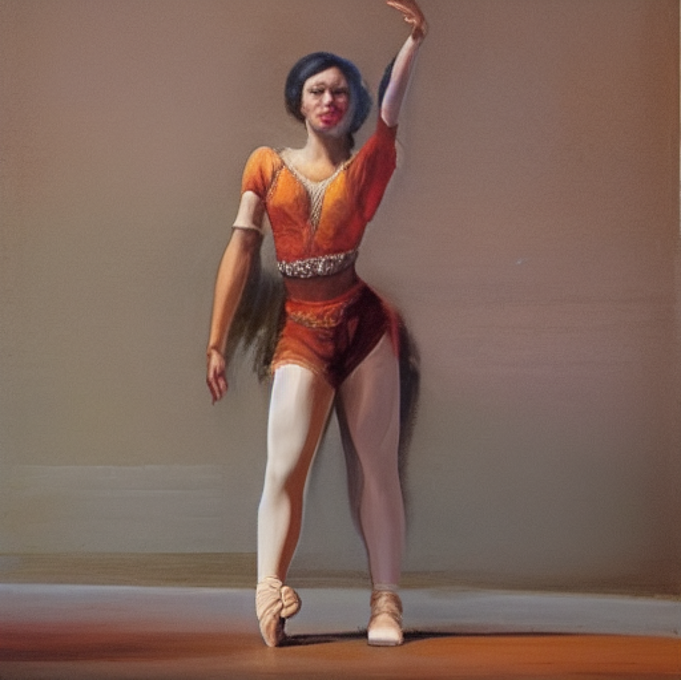}
  \end{subfigure}


 \begin{subfigure}[t]{.169\linewidth}
    \centering\includegraphics[width=\linewidth]{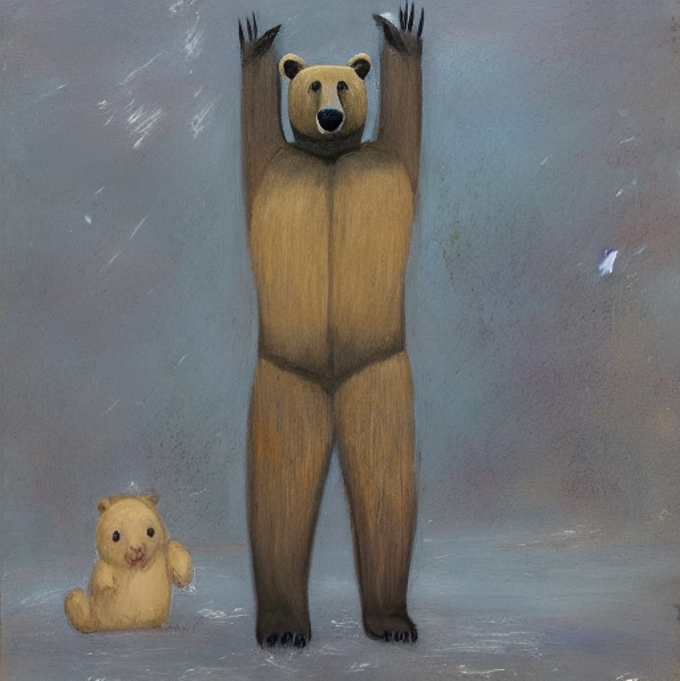}
    \end{subfigure}
      \begin{subfigure}[t]{.123\linewidth}
    \centering\includegraphics[width=\linewidth]{imgs/results/pose2.png}
    \end{subfigure}
  \begin{subfigure}[t]{.169\linewidth}
    \centering\includegraphics[width=\linewidth]{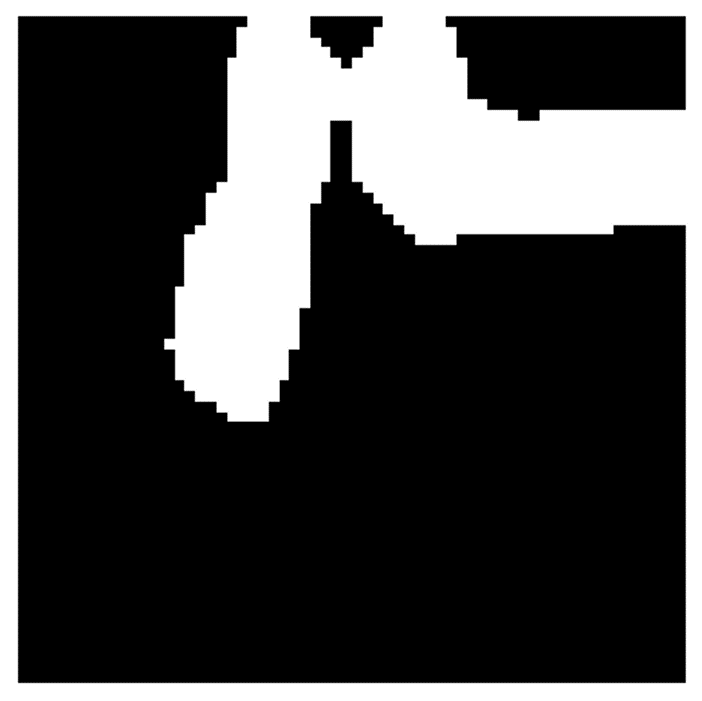}
  \end{subfigure}
  \begin{subfigure}[t]{.169\linewidth}
    \centering\includegraphics[width=\linewidth]{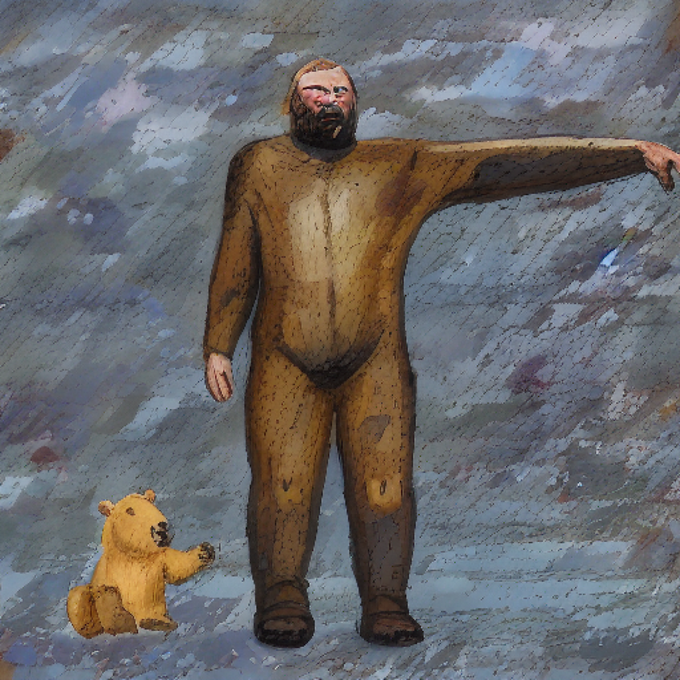}
  \end{subfigure}
  \begin{subfigure}[t]{.169\linewidth}
    \centering\includegraphics[width=\linewidth]{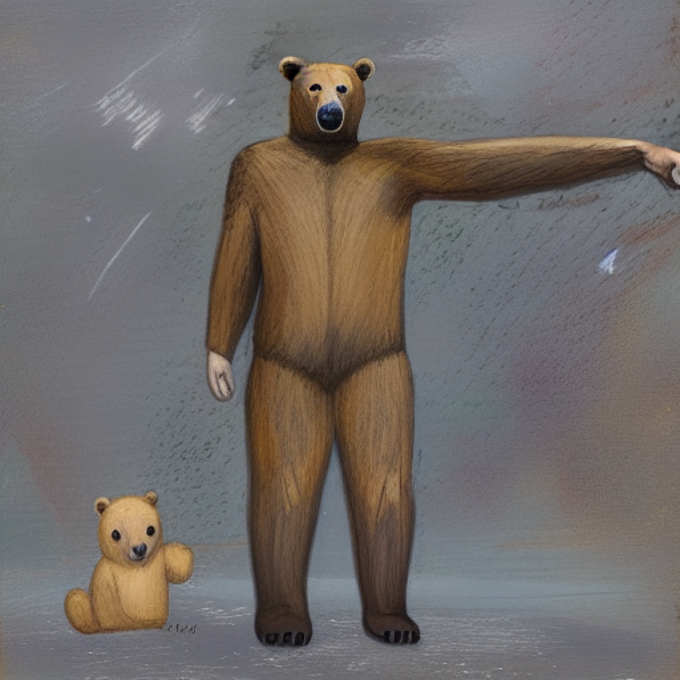}
  \end{subfigure}


\begin{subfigure}[t]{.169\linewidth}
    \centering\includegraphics[width=\linewidth]{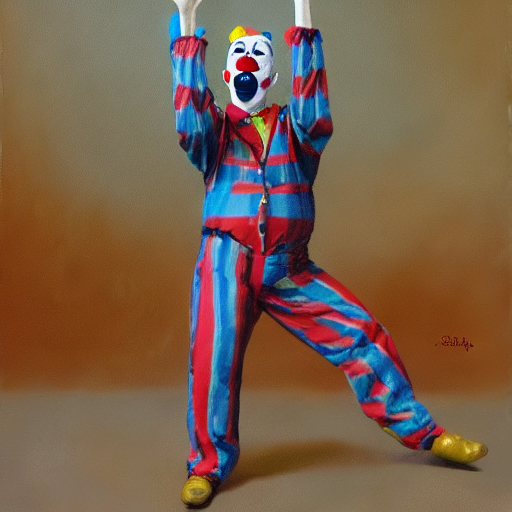}
    \caption{Original image}
    \end{subfigure}
      \begin{subfigure}[t]{.123\linewidth}
    \centering\includegraphics[width=\linewidth]{imgs/results/pose3.png}
    \caption{Target pose}
    \end{subfigure}
  \begin{subfigure}[t]{.169\linewidth}
    \centering\includegraphics[width=\linewidth]{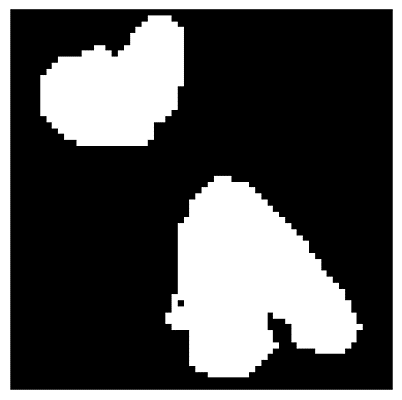}
    \caption{Edit mask}
  \end{subfigure}
  \begin{subfigure}[t]{.169\linewidth}
    \centering\includegraphics[width=\linewidth]{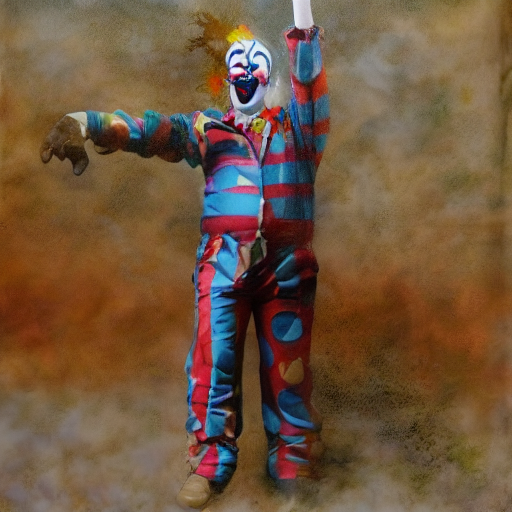}
    \caption{Guidance image}
  \end{subfigure}
  \begin{subfigure}[t]{.169\linewidth}
    \centering\includegraphics[width=\linewidth]{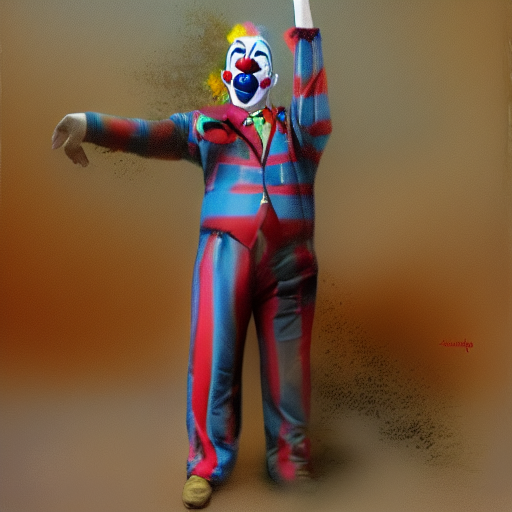}
    \caption{Final edit }
  \end{subfigure}


      \begin{subfigure}[t]{.162\linewidth}
    \centering\includegraphics[width=\linewidth]{imgs/results/superhero.png}
    \end{subfigure}
      \begin{subfigure}[t]{.152\linewidth}
    \centering\includegraphics[width=\linewidth]{imgs/results/superhero_s1.png}
    \end{subfigure}
  \begin{subfigure}[t]{.162\linewidth}
    \centering\includegraphics[width=\linewidth]{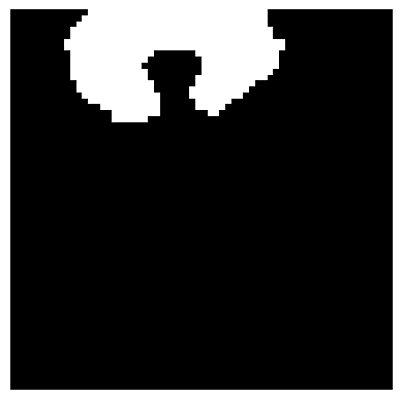}
  \end{subfigure}
  \begin{subfigure}[t]{.162\linewidth}
    \centering\includegraphics[width=\linewidth]{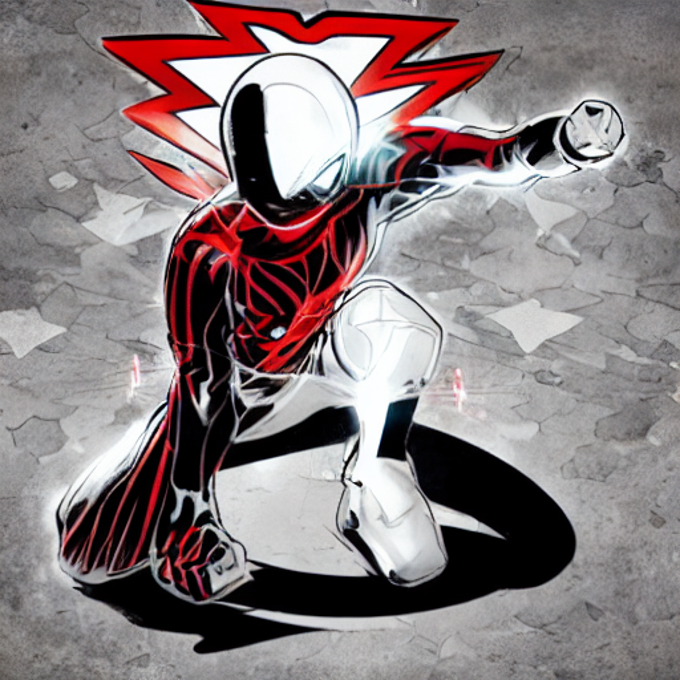}
  \end{subfigure}
  \begin{subfigure}[t]{.162\linewidth}
    \centering\includegraphics[width=\linewidth]{imgs/results/superhero_s1_final.png}
  \end{subfigure}


      \begin{subfigure}[t]{.162\linewidth}
    \centering\includegraphics[width=\linewidth]{imgs/results/cat.png}
     \caption{Original image}
    \end{subfigure}
      \begin{subfigure}[t]{.152\linewidth}
    \centering\includegraphics[width=\linewidth]{imgs/results/cat_s1.png}
     \caption{Target scribble}
    \end{subfigure}
  \begin{subfigure}[t]{.162\linewidth}
    \centering\includegraphics[width=\linewidth]{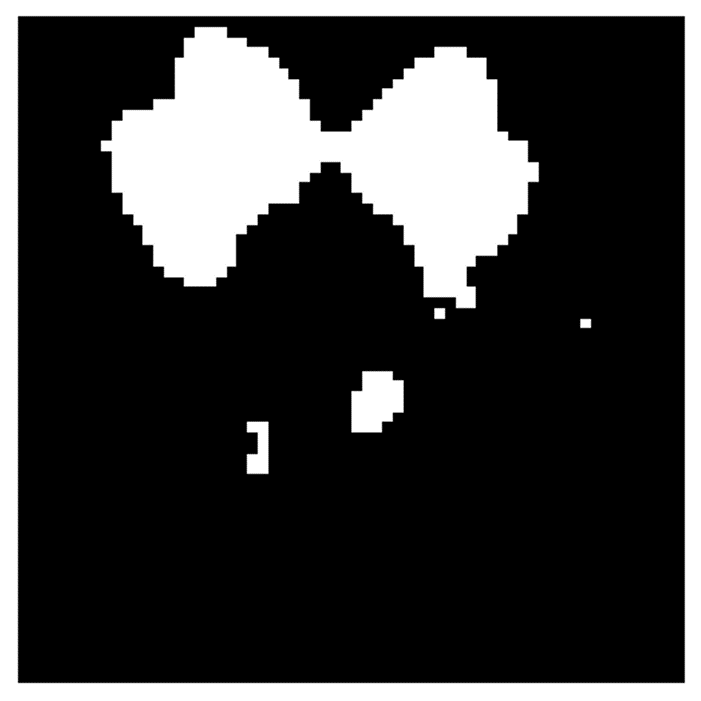}
    \caption{Edit mask}
  \end{subfigure}
  \begin{subfigure}[t]{.162\linewidth}
    \centering\includegraphics[width=\linewidth]{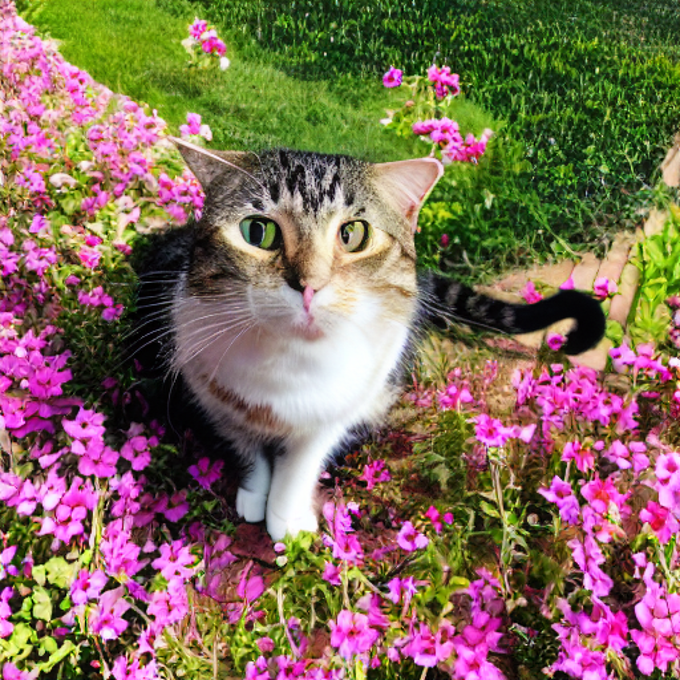}
    \caption{Guidance image}
  \end{subfigure}
  \begin{subfigure}[t]{.162\linewidth}
    \centering\includegraphics[width=\linewidth]{imgs/results/cat_final.png}
    \caption{Final edit}
  \end{subfigure}

\caption{ Visualisation of intermediate edit outputs on multiple pose and scribble conditionned image modifications. }
\label{fig:supp:intermediate}
\end{figure*}

In \cref{fig:supp:intermediate}, we provide detailed visualisations of our main paper results, specifically showing intermediate outputs of guidance images and estimated edit masks. 
It can be observed that the mask successfully identifies regions modified by the layout change. We highlight that the guidance image has reduced content preservation, and that larger changes are observed when the masked area is larger. This suggests that mask size may serve as a coarse proxy to task difficulty.

\section{ImageNet editing: visual examples}

\begin{figure*}[ht]
    \centering
\begin{subfigure}[t]{.2\linewidth}
    \centering\includegraphics[width=\linewidth]{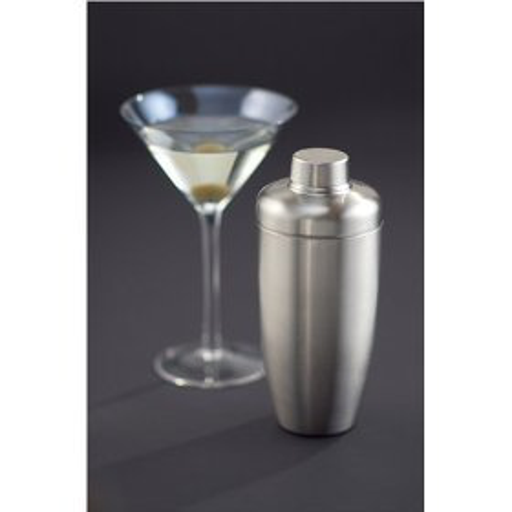}
    \caption{Cocktail shaker \textrightarrow coffee mug}
    \end{subfigure}
  \begin{subfigure}[t]{.2\linewidth}
    \centering\includegraphics[width=\linewidth]{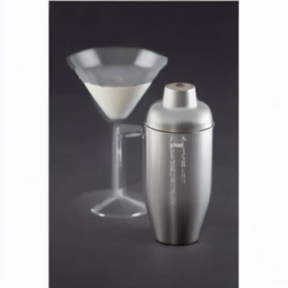}
  \end{subfigure}
  \begin{subfigure}[t]{.2\linewidth}
    \centering\includegraphics[width=\linewidth]{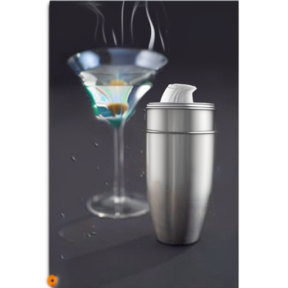}
  \end{subfigure}
    \begin{subfigure}[t]{.2\linewidth}
    \centering\includegraphics[width=\linewidth]{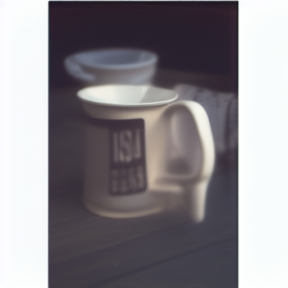}
  \end{subfigure}
  

\begin{subfigure}[t]{.2\linewidth}
    \centering\includegraphics[width=\linewidth]{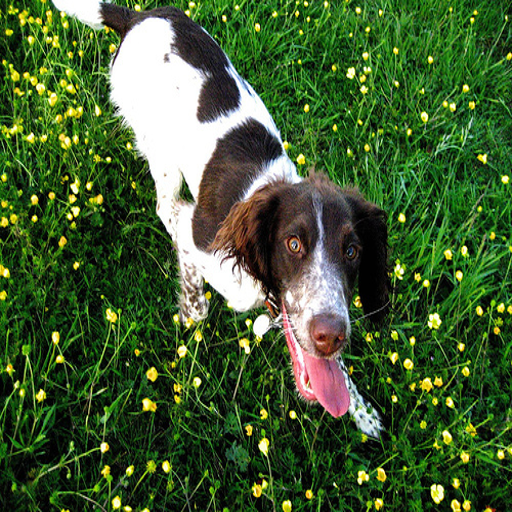}
    \caption{English springer \textrightarrow golden retriever}
    \end{subfigure}
  \begin{subfigure}[t]{.2\linewidth}
    \centering\includegraphics[width=\linewidth]{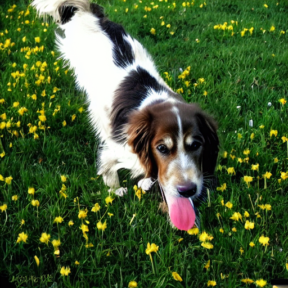}
  \end{subfigure}
  \begin{subfigure}[t]{.2\linewidth}
    \centering\includegraphics[width=\linewidth]{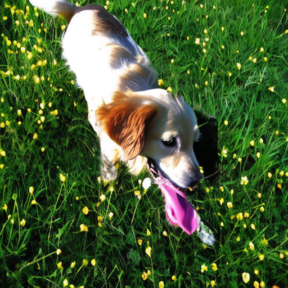}
  \end{subfigure}
    \begin{subfigure}[t]{.2\linewidth}
    \centering\includegraphics[width=\linewidth]{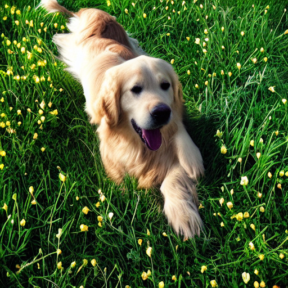}
  \end{subfigure}


\begin{subfigure}[t]{.2\linewidth}
    \centering\includegraphics[width=\linewidth]{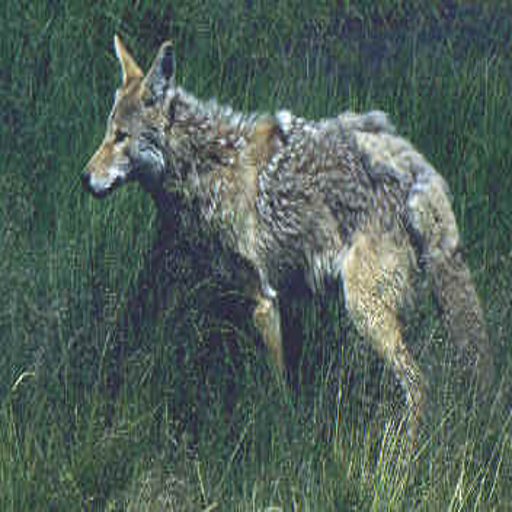}
    \caption{Coyote \textrightarrow red fox}
    \end{subfigure}
  \begin{subfigure}[t]{.2\linewidth}
    \centering\includegraphics[width=\linewidth]{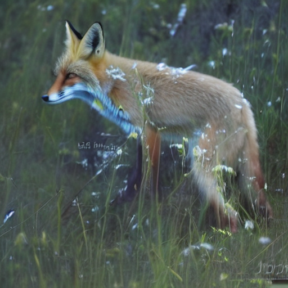}
  \end{subfigure}
  \begin{subfigure}[t]{.2\linewidth}
    \centering\includegraphics[width=\linewidth]{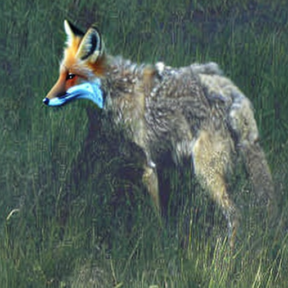}
  \end{subfigure}
    \begin{subfigure}[t]{.2\linewidth}
    \centering\includegraphics[width=\linewidth]{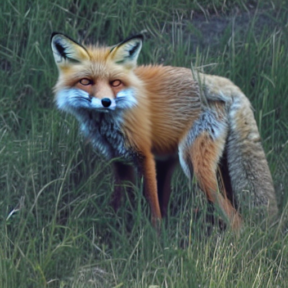}
  \end{subfigure}


\begin{subfigure}[t]{.2\linewidth}
    \centering\includegraphics[width=\linewidth]{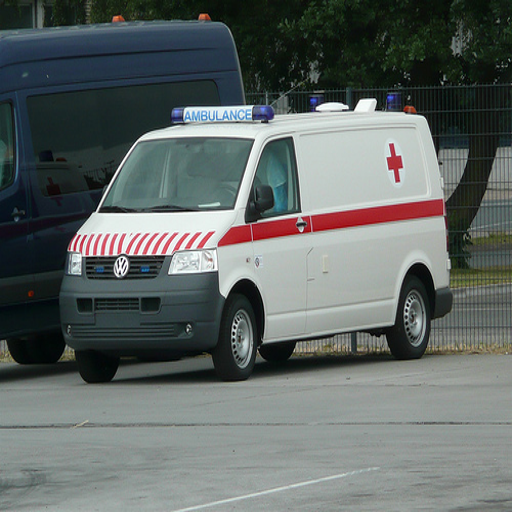}
    \caption{Ambulance \textrightarrow sports car}
    \end{subfigure}
  \begin{subfigure}[t]{.2\linewidth}
    \centering\includegraphics[width=\linewidth]{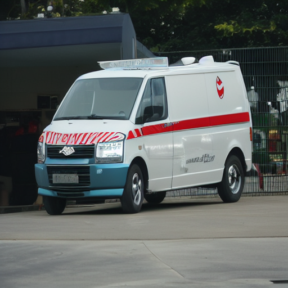}
  \end{subfigure}
  \begin{subfigure}[t]{.2\linewidth}
    \centering\includegraphics[width=\linewidth]{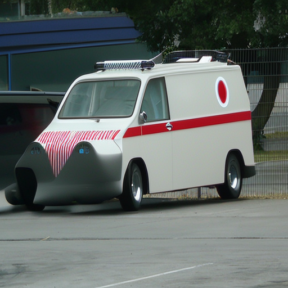}
  \end{subfigure}
    \begin{subfigure}[t]{.2\linewidth}
    \centering\includegraphics[width=\linewidth]{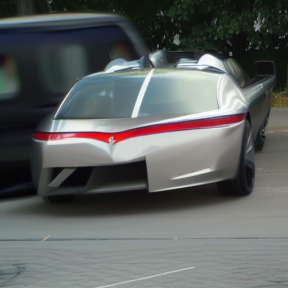}
  \end{subfigure}


  \begin{subfigure}[t]{.2\linewidth}
    \centering\includegraphics[width=\linewidth]{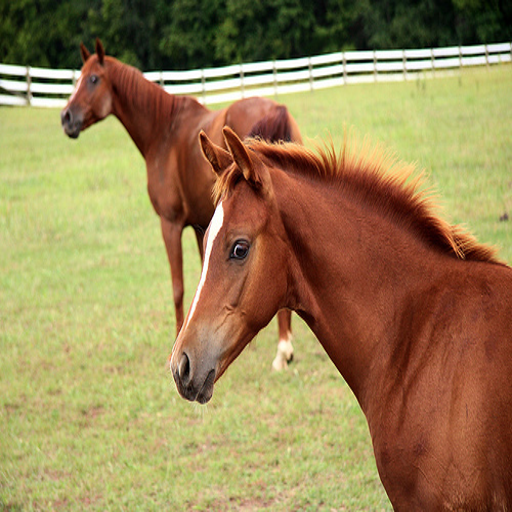}
    \caption{horse \textrightarrow zebra}
    \end{subfigure}
  \begin{subfigure}[t]{.2\linewidth}
    \centering\includegraphics[width=\linewidth]{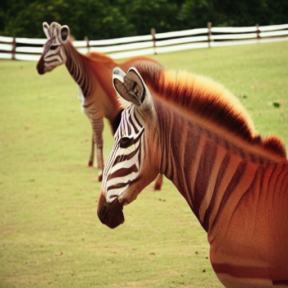}
    \caption{Prompt to prompt}
  \end{subfigure}
  \begin{subfigure}[t]{.2\linewidth}
    \centering\includegraphics[width=\linewidth]{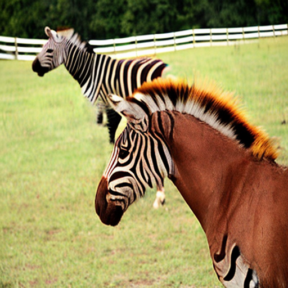}
    \caption{DiffEdit $ER=0.7$}
  \end{subfigure}
  \begin{subfigure}[t]{.2\linewidth}
    \centering\includegraphics[width=\linewidth]{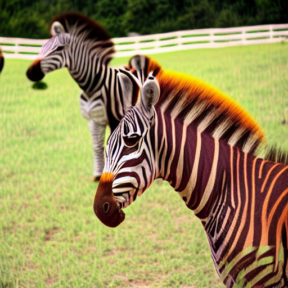}
    \caption{Ours}
  \end{subfigure}

\caption{ Visual results of the ImageNet editing task. }
\label{fig:supp:imagenet}
\end{figure*}

In Fig.~\ref{fig:supp:imagenet}, we provide a visual comparison of edits carried out on the ImageNet dataset, i.e.~our text-editing quantitative experiments reported in the main paper (see Sec. 4). We highlight how our method's flexibility achieves larger modifications (i.e.~shaker to coffee mug, english springer to golden retriever) when other methods fail, while preserving image structure. We further note how masks that are too restrictive can negatively impact DiffEdit based editing (see e.g.~coyote to red fox where the mask is limited to the animal's head). While we leverage the same mask as DiffEdit, its integration into our optimisation procedure affords more flexibility.

\section{Additional editing settings}
We provide additional results depicting our model's ability to perform editing with additional types of edit instructions. We demonstrate image editing using Hough line map conditions, as well as multi-condition editing. 

\subsection{Editing real image with pose conditioning}
\begin{figure*}[ht]
    \centering
    \begin{subfigure}[t]{.194\linewidth}
    \centering\includegraphics[width=\linewidth]{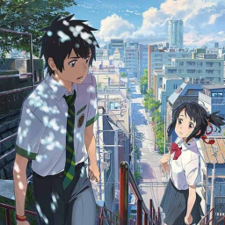}
    \caption{Original image}
    \end{subfigure}
      \begin{subfigure}[t]{.144\linewidth}
    \centering\includegraphics[width=\linewidth]{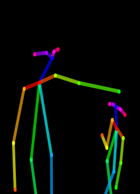}
    \caption{Target pose}
    \end{subfigure}
  \begin{subfigure}[t]{.194\linewidth}
    \centering\includegraphics[width=\linewidth]{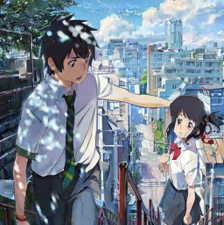}
    \caption{Diffedit}
  \end{subfigure}
  \begin{subfigure}[t]{.194\linewidth}
    \centering\includegraphics[width=\linewidth]{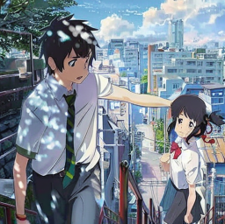}
    \caption{Ours ($\lambda=1$)}
  \end{subfigure}
  \begin{subfigure}[t]{.194\linewidth}
    \centering\includegraphics[width=\linewidth]{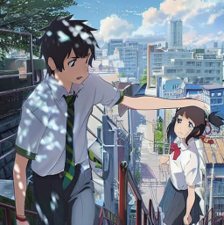}
    \caption{Ours}
  \end{subfigure}
  

      \centering
    \begin{subfigure}[t]{.194\linewidth}
    \centering\includegraphics[width=\linewidth]{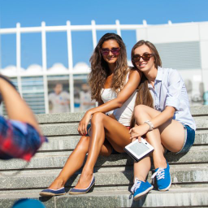}
    \caption{Original image}
    \end{subfigure}
      \begin{subfigure}[t]{.144\linewidth}
    \centering\includegraphics[width=\linewidth]{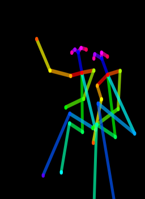}
    \caption{Target pose}
    \end{subfigure}
  \begin{subfigure}[t]{.194\linewidth}
    \centering\includegraphics[width=\linewidth]{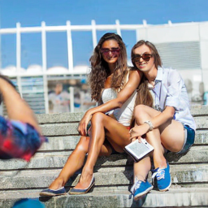}
    \caption{Diffedit}
  \end{subfigure}
  \begin{subfigure}[t]{.194\linewidth}
    \centering\includegraphics[width=\linewidth]{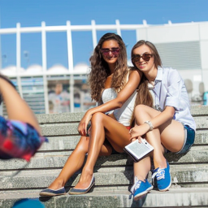}
    \caption{Ours ($\lambda=1$)}
  \end{subfigure}
  \begin{subfigure}[t]{.194\linewidth}
    \centering\includegraphics[width=\linewidth]{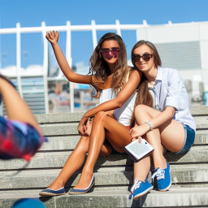}
    \caption{Ours}
  \end{subfigure}

      \centering
    \begin{subfigure}[t]{.194\linewidth}
    \centering\includegraphics[width=\linewidth]{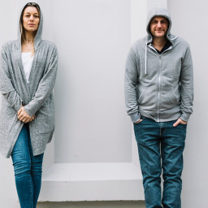}
    \caption{Original image}
    \end{subfigure}
      \begin{subfigure}[t]{.144\linewidth}
    \centering\includegraphics[width=\linewidth]{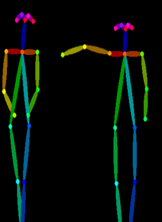}
    \caption{Target pose}
    \end{subfigure}
  \begin{subfigure}[t]{.194\linewidth}
    \centering\includegraphics[width=\linewidth]{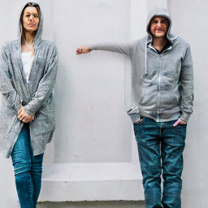}
    \caption{Diffedit}
  \end{subfigure}
  \begin{subfigure}[t]{.194\linewidth}
    \centering\includegraphics[width=\linewidth]{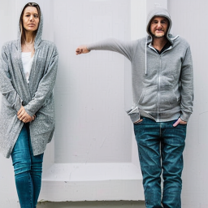}
    \caption{Ours ($\lambda=1$)}
  \end{subfigure}
  \begin{subfigure}[t]{.194\linewidth}
    \centering\includegraphics[width=\linewidth]{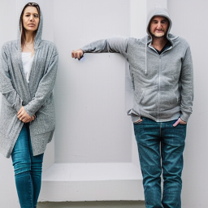}
    \caption{Ours}
  \end{subfigure}

\caption{ Visual results for editing of real images with pose-based conditioning.}
\label{fig:supp:real}
\end{figure*}

In Fig.~\ref{fig:supp:real} we show editing examples of real images comprising multiple persons, compared to DiffEdit. We adapt our parameter set to the different type of input images, setting $t_u=20$ and $k=20$, and the controlNet conditionning scale to 1.0. We additionally increase our mask threshold to $0.2$ to improve content preservation. Once again, our method achieves very close performance to DiffEdit with $\lambda=1$, while we are able to achieve more modifications, with a more realistic look.

\subsection{Hough line conditioning}
\begin{figure*}[ht]
    \centering
    \begin{subfigure}[t]{.194\linewidth}
    \centering\includegraphics[width=\linewidth]{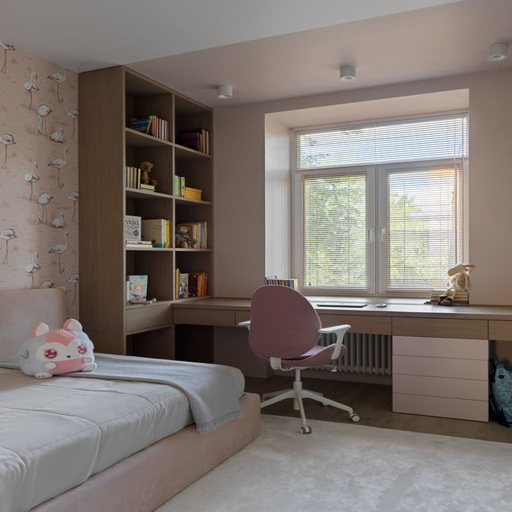}
    \caption{Original image}
    \end{subfigure}
      \begin{subfigure}[t]{.194\linewidth}
    \centering\includegraphics[width=\linewidth]{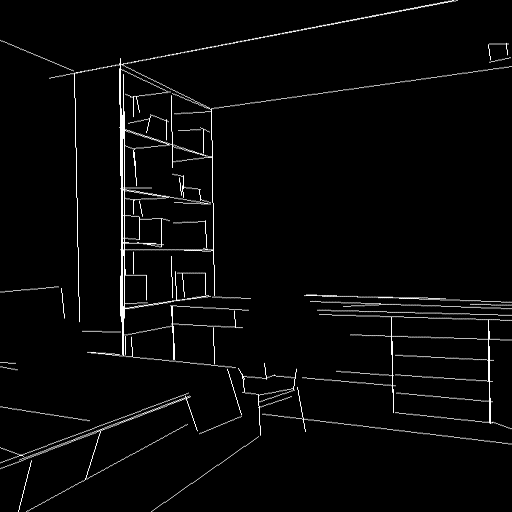}
    \caption{Target Hough map}
    \end{subfigure}
  \begin{subfigure}[t]{.194\linewidth}
    \centering\includegraphics[width=\linewidth]{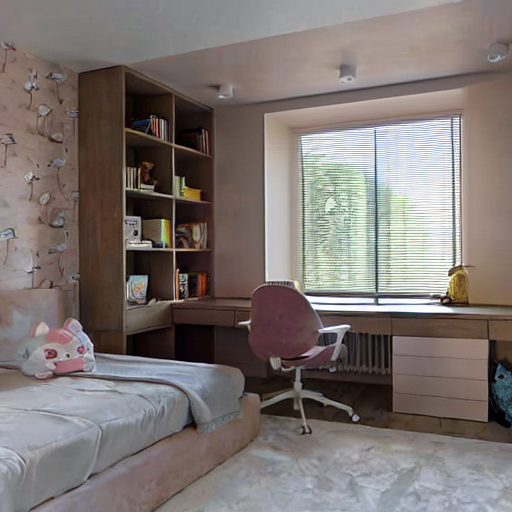}
    \caption{DiffEdit + our mask}
  \end{subfigure}
  \begin{subfigure}[t]{.194\linewidth}
    \centering\includegraphics[width=\linewidth]{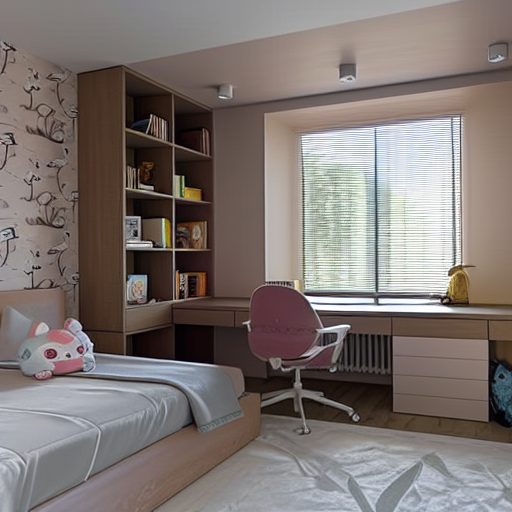}
    \caption{Ours $\lambda=1$}
  \end{subfigure}
  \begin{subfigure}[t]{.194\linewidth}
    \centering\includegraphics[width=\linewidth]{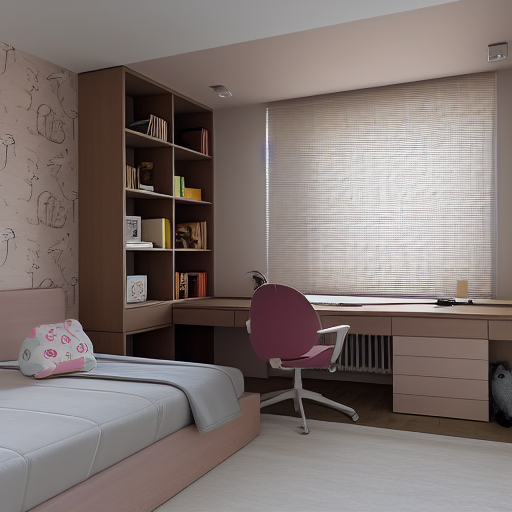}
    \caption{Ours}
  \end{subfigure}
\caption{ Visual results of editing with Hough line map conditioning.}
\label{fig:supp:hough}
\end{figure*}

Fig.~\ref{fig:supp:hough} shows an example edit using Hough line maps as conditioning, compared to DiffEdit. As with other forms of conditioning, we can see the DiffEdit output is very similar to our $\lambda=1$ output, while our $\lambda=0.6$ output allows larger modifications, more aligned with the new conditioning. 

\subsection{Multiple conditions}
\begin{figure*}[ht]
    \centering
    \begin{subfigure}[t]{.194\linewidth}
    \centering\includegraphics[width=\linewidth]{imgs/results/dog.png}
    \caption{Original image: a dog at the beach \textrightarrow a cat at the beach}
    \end{subfigure}
      \begin{subfigure}[t]{.194\linewidth}
    \centering\includegraphics[width=\linewidth]{imgs/results/dog_s1.png}
    \caption{Target scribble}
    \end{subfigure}
  \begin{subfigure}[t]{.194\linewidth}
    \centering\includegraphics[width=\linewidth]{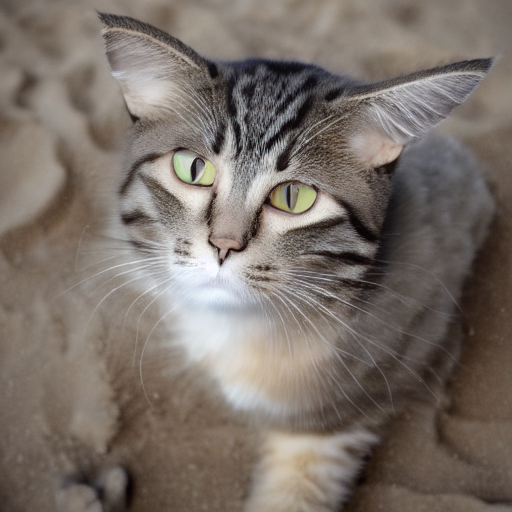}
    \caption{ControlNet}
  \end{subfigure}
  \begin{subfigure}[t]{.194\linewidth}
    \centering\includegraphics[width=\linewidth]{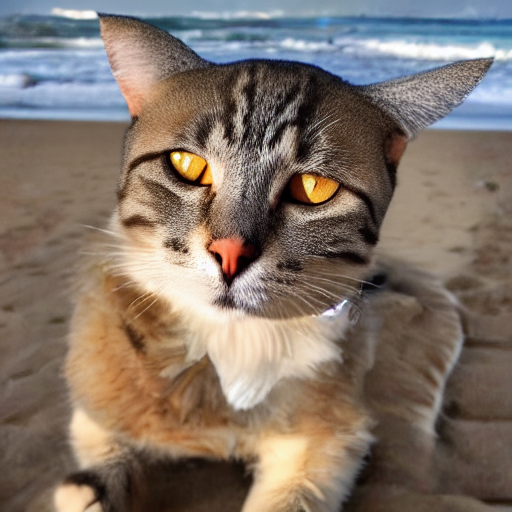}
    \caption{DiffEdit + our mask}
  \end{subfigure}
  \begin{subfigure}[t]{.194\linewidth}
    \centering\includegraphics[width=\linewidth]{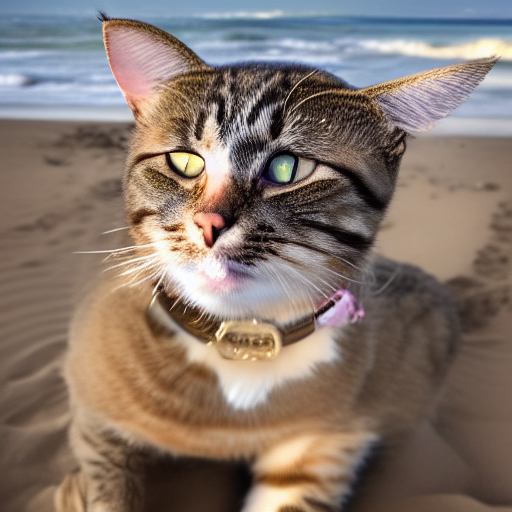}
    \caption{Ours}
  \end{subfigure}
  

  \begin{subfigure}[t]{.204\linewidth}
    \centering\includegraphics[width=\linewidth]{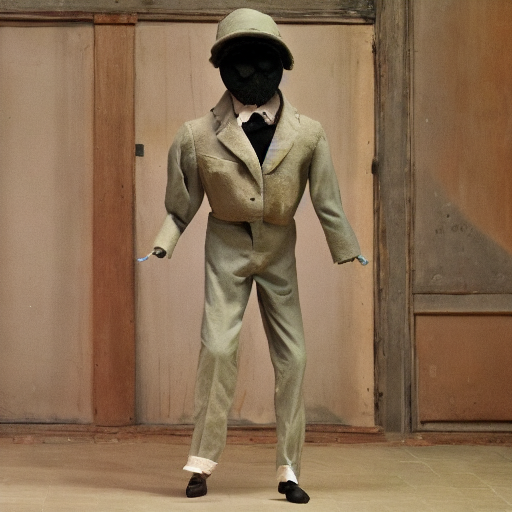}
    \caption{Original image: a puppet dancing \textrightarrow a pirate dancing}
    \end{subfigure}
      \begin{subfigure}[t]{.148\linewidth}
    \centering\includegraphics[width=\linewidth]{imgs/results/pose1.png}
    \caption{Target pose}
    \end{subfigure}
  \begin{subfigure}[t]{.204\linewidth}
    \centering\includegraphics[width=\linewidth]{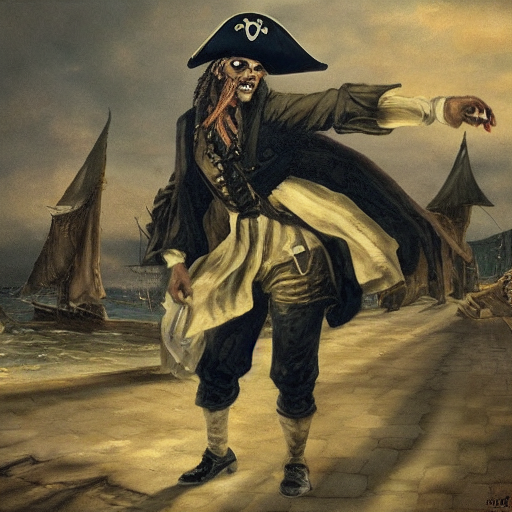}
    \caption{ControlNet}
  \end{subfigure}
  \begin{subfigure}[t]{.204\linewidth}
    \centering\includegraphics[width=\linewidth]{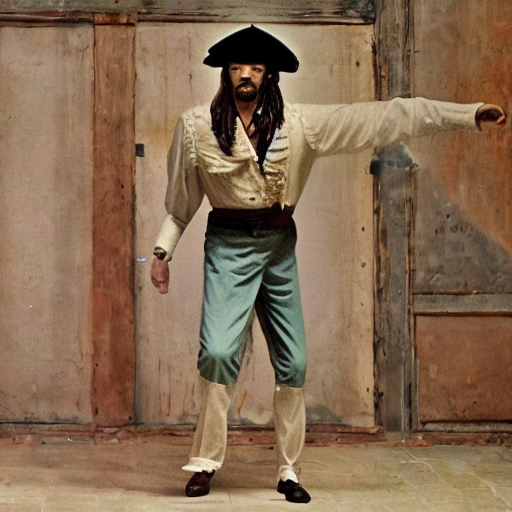}
    \caption{DiffEdit + our mask}
  \end{subfigure}
  \begin{subfigure}[t]{.204\linewidth}
    \centering\includegraphics[width=\linewidth]{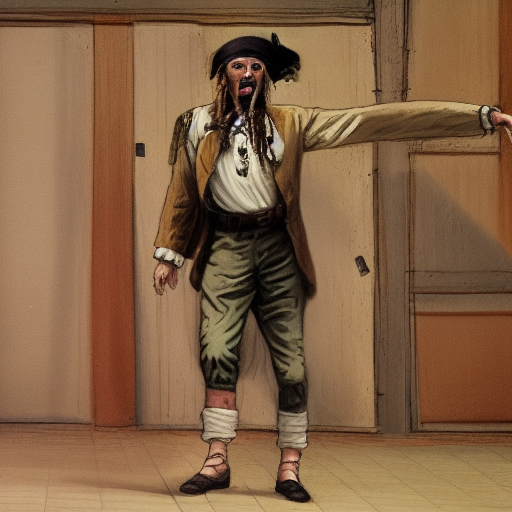}
    \caption{Ours}
  \end{subfigure}

\caption{ Visual results for multi-condition editing, combining pose/scribble with text edit instructions.}
\label{fig:supp:multi}
\end{figure*}

Fig.~\ref{fig:supp:multi} shows examples combining multiple types of editing conditions, namely $\text{text}{+}\text{scribble}$ and $\text{text}{+}\text{pose}$. We compare our results to ControlNet and DiffEdit. We observe that both our approach and DiffEdit are able to integrate two types of edit conditions, while ControlNet fails to maintain visual consistency, as one would expect due to the lack of preservation mechanisms, respectively. 

\section{Addressing failure modes: guidance image refinement}
For certain images with complex background and large editing constraints, it can be difficult to generate a guidance image that successfully preserves all image and background details. This can lead to the introduction of local image artefacts. To address this limitation, we investigate whether it is possible to refine our guidance image $G$ using information from our original input image $I$, effectively re-injecting information from the source image. To do this, we leverage our inference time optimisation with $\lambda=0$, leveraging only a guidance loss. In this setting, the input image to be updated is $G$, and the reference image is $I$. We carry out this update using $t_u=6$ and $k=1$. We note that this approach can alter the quality of the guidance image, as it is encoded/decoded multiple times, and therefore recommend integrating this step only if editing results are unsatisfactory with our original architecture. 

\begin{figure*}
\centering
 \begin{subfigure}[t]{.169\linewidth}
    \centering\includegraphics[width=\linewidth]{imgs/results/astronaut_pose0.png}
    \end{subfigure}
      \begin{subfigure}[t]{.123\linewidth}
    \centering\includegraphics[width=\linewidth]{imgs/results/pose3.png}
    \end{subfigure}
  \begin{subfigure}[t]{.169\linewidth}
    \centering\includegraphics[width=\linewidth]{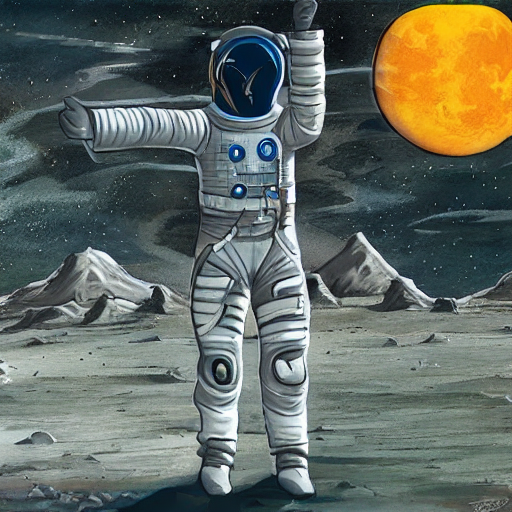}
  \end{subfigure}
  \begin{subfigure}[t]{.169\linewidth}
    \centering\includegraphics[width=\linewidth]{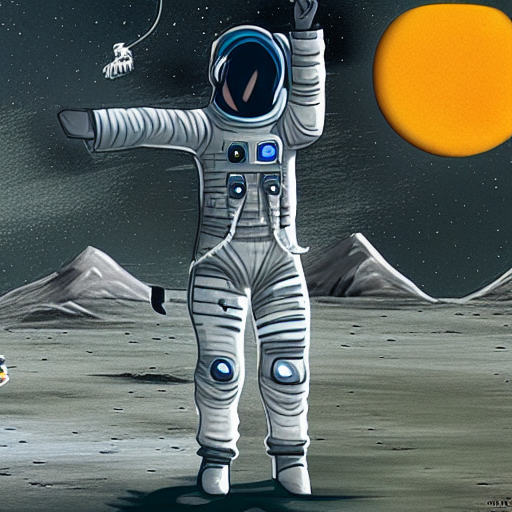}
  \end{subfigure}
  \begin{subfigure}[t]{.169\linewidth}
    \centering\includegraphics[width=\linewidth]{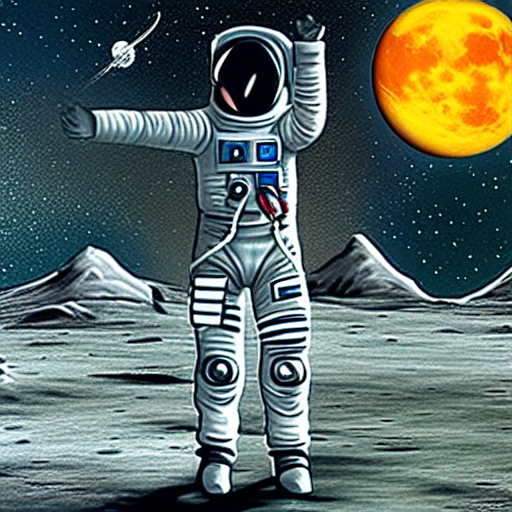}
  \end{subfigure}
  \begin{subfigure}[t]{.169\linewidth}
    \centering\includegraphics[width=\linewidth]{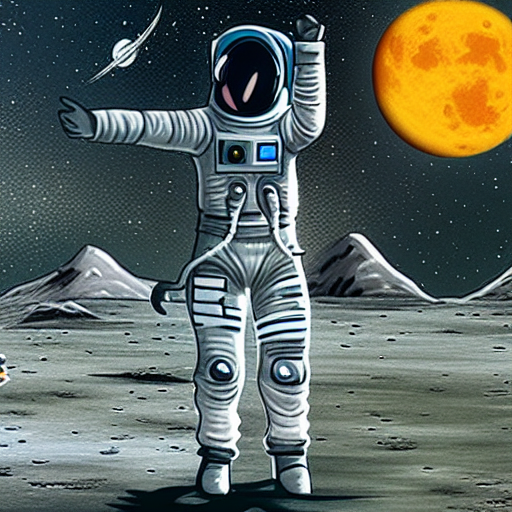}
  \end{subfigure}


\begin{subfigure}[t]{.169\linewidth}
    \centering\includegraphics[width=\linewidth]{imgs/results/astronaut_pose0.png}
    \caption{Original image}
    \end{subfigure}
      \begin{subfigure}[t]{.123\linewidth}
    \centering\includegraphics[width=\linewidth]{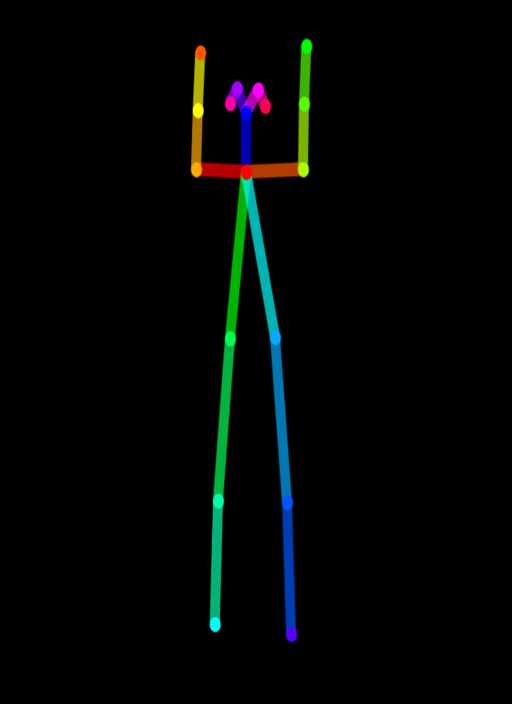}
    \caption{Target pose}
    \end{subfigure}
  \begin{subfigure}[t]{.169\linewidth}
    \centering\includegraphics[width=\linewidth]{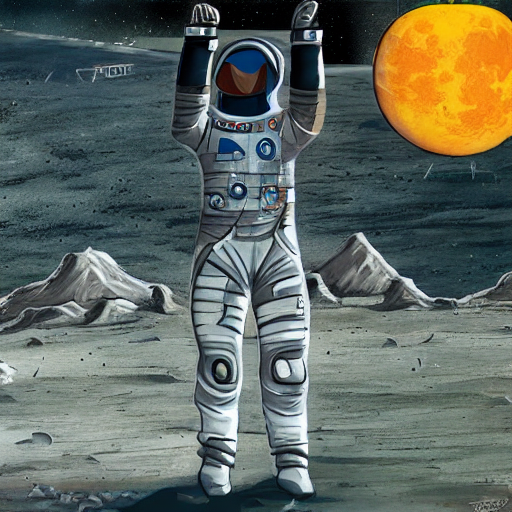}
    \caption{Guidance image}
  \end{subfigure}
  \begin{subfigure}[t]{.169\linewidth}
    \centering\includegraphics[width=\linewidth]{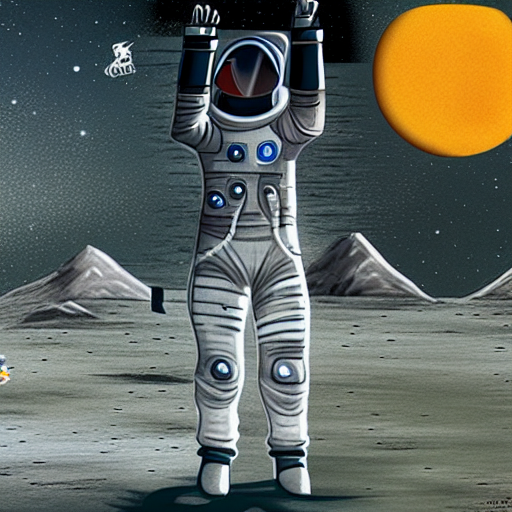}
    \caption{Final edit with guidance image}
  \end{subfigure}
  \begin{subfigure}[t]{.169\linewidth}
    \centering\includegraphics[width=\linewidth]{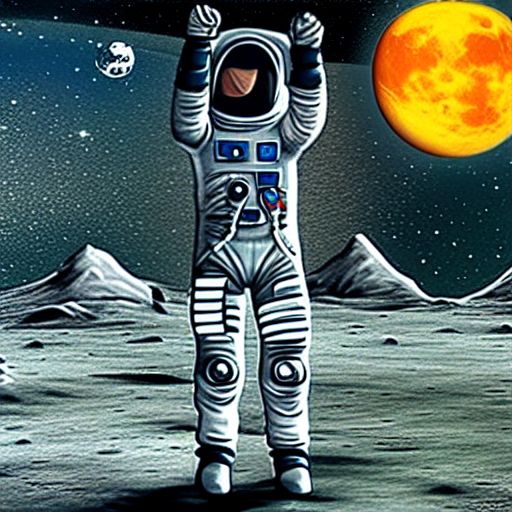}
    \caption{Refined guidance image}
  \end{subfigure}
  \begin{subfigure}[t]{.169\linewidth}
    \centering\includegraphics[width=\linewidth]{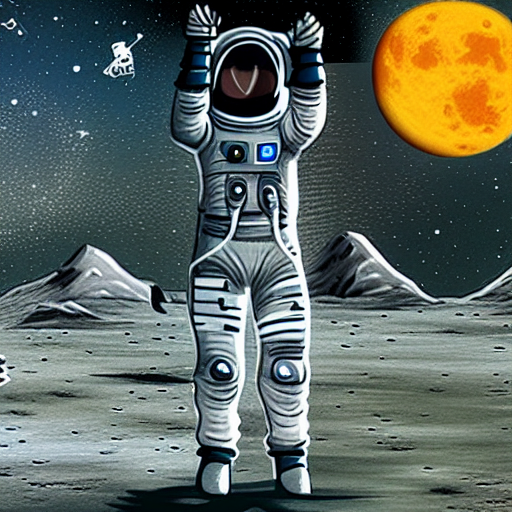}
    \caption{Final edit with refined guidance image}
  \end{subfigure}

\caption{ Illustration of the impact of our additional guidance image refinement process, using a pose editing condition. }
\label{fig:supp:refinement}
\end{figure*}

We illustrate the failure mode and proposed guidance image refinement solution in Fig~~\ref{fig:supp:refinement}, showing how this additional step can successfully eliminate artefacts introduced by our original guidance image. 

\end{document}